\documentclass[preprints,article,accept,moreauthors,pdftex]{Definitions/mdpi}
\usepackage{algorithmic}
\usepackage[ruled,vlined]{algorithm2e}

\firstpage{1} 
\pubvolume{xx}
\issuenum{1}
\articlenumber{5}
\pubyear{2019}
\copyrightyear{2019}
\history{Received: date; Accepted: date; Published: date}
 \usepackage{xcolor}

\Title{High-dimensional separability for one- and few-shot learning}

\Author{Alexander N. Gorban $^{1,2,}$*\orcidA{}, Bogdan Grechuk $^{1}$\orcidE{},  Evgeny M. Mirkes $^{1,2}$\orcidB{}, Sergey V. Stasenko$^{2}$\orcidC{}	 and Ivan Y. Tyukin $^{1,2,3}$\orcidD{}}

\AuthorNames{Alexander N. Gorban, Bogdan Grechuk, Evgeny M. Mirkes, Sergey V. Stasenko,   Ivan Y. Tyukin}

\address{%
$^{1}$ \quad Department of Mathematics, University of Leicester, Leicester,   UK\\
$^{2}$ \quad Lobachevsky University, Nizhni Novgorod, Russia\\
$^{3}$ \quad Department of Geoscience and Petroleum, Norwegian University of Science and Technology}

\corres{Correspondence: a.n.gorban@le.ac.uk}

\abstract{This work is driven by a practical question: corrections of Artificial Intelligence (AI) errors. These corrections should be quick and non-iterative. To solve this problem without modification of a legacy AI system, we propose special `external' devices, correctors. Elementary  correctors consist of two parts, a classifier that separates the situations with high risk of error from the situations in which the legacy AI system works well and a new decision that should be recommended for situations with potential errors. Input signals for the correctors can be the inputs of the legacy AI system, its internal signals, and outputs. If the intrinsic dimensionality of data is high enough then the classifiers for correction of small number of errors can be very simple. According to  the blessing of dimensionality effects, even simple and robust Fisher's discriminants can be used for one-shot learning of AI correctors. Stochastic separation theorems provide the mathematical basis for this one-short learning. However, as the number of correctors needed grows, the cluster structure of data becomes important and a new family of stochastic separation theorems is required.
We refuse the classical hypothesis of the regularity of the data distribution and assume that the data can have a rich fine-grained structure with many clusters and corresponding peaks in the probability density. 
 New stochastic separation theorems for data with fine-grained structure are formulated and proved. On the basis of these theorems, the multi-correctors for granular data are proposed. The advantages of the multi-corrector technology  were demonstrated by examples of correcting errors and learning new classes of objects by a deep convolutional neural network on the CIFAR-10 dataset. The key problems of the non-classical high-dimensional data analysis are reviewed together with the basic preprocessing steps including the correlation transformation, supervised Principal Component Analysis (PCA), semi-supervised PCA, transfer component analysis, and new domain adaptation PCA.}
\keyword{Artificial Intelligence; blessing of dimensionality; clusters; errors; separability; discriminant; dimensionality reduction}
 
\begin{document}

\section{Introduction}

 \subsection{AI Errors and Correctors}

The main driver of our research is the problem of Artificial Intelligence (AI) errors and their correction: {\em {all AI systems sometimes make errors and will make errors in the future.} 
} These errors must be detected and corrected immediately and locally in the networks of AI systems. If we do not solve this problem, then a new AI winter will come.
Recall that the previous AI winters came after the hype peaks of inflated expectations and bold advertising: 
 the general overconfidence of experts was a typical symptom of inflated expectations before the winter came \cite{Armstrong2014}. ``It was recognised that AI advocates were called to account for making promises that they could not fulfill. There was disillusionment'' \cite{Perez Cruz2021} and  ``significant investments were made, but real breakthroughs were very rare and both time and patience ran out...'' \cite{Lloyd1995}. A richer picture of the AI winter, including the dynamics of government funding, the motivation of AI researchers, the transfer of AI to industry, and hardware development, was sketched in \cite{Hendler2008}. The winter may come back and we better be ready \cite{Floridi2020}.
 For the detailed discussion of AI trust, limitations, conflation, and hype we refer to the analytic  review of  Bowman and Grindrod \cite{Bowman2019}.

Gartner's Hype Cycle is a  convenient tool to represent of R\&D trends. According to Gartner \cite{Gartner2019}, the data-driven Artificial Intelligence (AI)  has already left the Peak of Inflated Expectation and is  descending  into the Trough of Disillusionment. If we look at  Gartner's Hype Cycle in more detail, we will see that Machine Learning and Deep Learning are going down. Explainable AI joined them in 2020,  but Responsible AI, Generative AI, and Self-Supervising Learning are still climbing up the peak. \cite{Gartner2020}.  
  
\begin{figure}[H]
\includegraphics[width=0.8\textwidth]{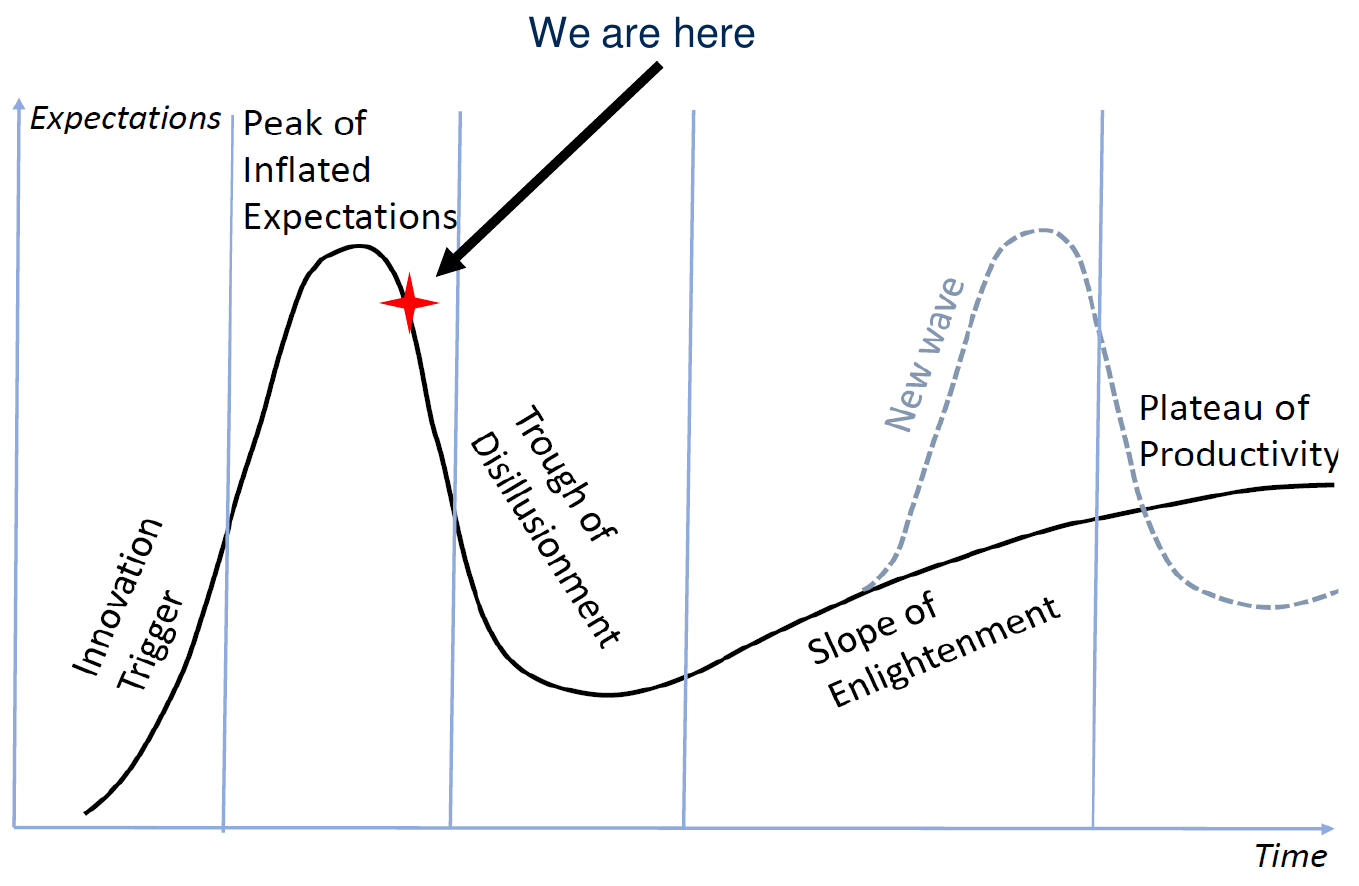}
\caption{\label{Gartner} Gartner Hype Cycle and its phases. Position of the data-driven AI on the hype cycle is marked by a four-pointed star. A possible new hype peak (new wave) is represented by the dashed line.} 
\end{figure}

According to Gartner's  Hype cycle model, the   Trough of Disillusionment will turn into the Slope of Enlightenment that leads to the Plateau of Productivity. The modern Peak and Trough are not the first in the history of AI. Surprisingly, previous troughs (AI winters)  did not turn into the performance plateaus. Instead they went through new peaks of hype and inflated                      expectations (Figure~\ref{Gartner}) \cite{GorbanGrechukTykin2018}.

 
What pushes the AI downhill now? Is it the same problem that pushed the AI down previous slopes decades ago? Data driven systems ``will inevitably and unavoidably generate errors'', and this is of great  concern \cite{Yeung2019}.  The main problem for the widespread use of AI around the world is unexpected errors in real-life applications:
\begin{itemize}
\item The mistakes can be dangerous;
\item Usually, it remains unclear who is responsible for them;
\item The types of errors are numerous and often unpredictable;  
\item The real world is not a good i.i.d. (independent identically distributed) sample;
\item We cannot rely on a statistical estimate of the probability of errors in real life.
\end{itemize}

The hypothesis of i.i.d. data samples is very popular in machine learning theory.   It means that there exists a probability measure on the data space and the data points are drawn from the space according to this measure independently \cite{Cucker2002}. It is worth mentioning that the data point for supervising learning includes both the input and the desired output and the probability is defined on the input $\times$ output space. Existence and stationarity of the probability distribution in real life is a very strong hypothesis. To weaken this assumption, many auxiliary concepts have been developed, such as concept drift. Nevertheless, i.i.d samples remain a central assumption of statistical learning theory: the dataset is presumed to be an i.i.d. random sample drawn from a probability distribution \cite{FriedmanHastieTib2009}. 
 
Fundamental origins of AI errors could be different. Of course, they include software errors, unexpected human behaviour, and non-intended use as well as many other possible reasons. Nevertheless, the universal cause  of errors is uncertainty in training data and in training process. The real world possibilities are not covered by the dataset.

The mistakes should be corrected. The systematic retraining of a big AI system seems to be rarely possible (here and below, AI skill means the ability to correctly solve a group of similar tasks):
\begin{itemize}
\item To preserve existing skills we must use the full set of training data;
\item This approach requires significant computational resources for each error;
\item However, new errors may appear in the course of retraining;
\item The preservation of existing skills is not guaranteed;
\item The probability of  damage  to skills is a priori unknown.
\end{itemize}

Therefore, quick non-iterative methods which are free from the disadvantages listed above are required. This is the main challenge for the one- and few-shot learning methods. 
 
To provide fast error correction, we must consider developing {\em correctors 
}, external devices that complement legacy Artificial Intelligence systems, diagnose the risk of error, and correct errors. The original AI system remains a part of the extended 'system + corrector' complex. Therefore, the correction is reversible, and the original system can always be extracted from the augmented  AI complex. Correctors have two different functions: (1)~they should recognise potential errors and (2) provide corrected outputs for situations with potential errors. The idealised scheme of a legacy AI system augmented with an elementary corrector is presented in Figure~\ref{SingleCorrector}. {Here, the legacy AI system is represented as a transformation that maps the {\em input } signals into {\em internal} signals and then into {\em output} signals: $inputs \to internal \to outputs$. The elementary corrector takes all these signals as inputs and makes a decision about correction (see Figure~\ref{SingleCorrector}). 

\begin{figure}[H]
\includegraphics[width=0.65\textwidth]{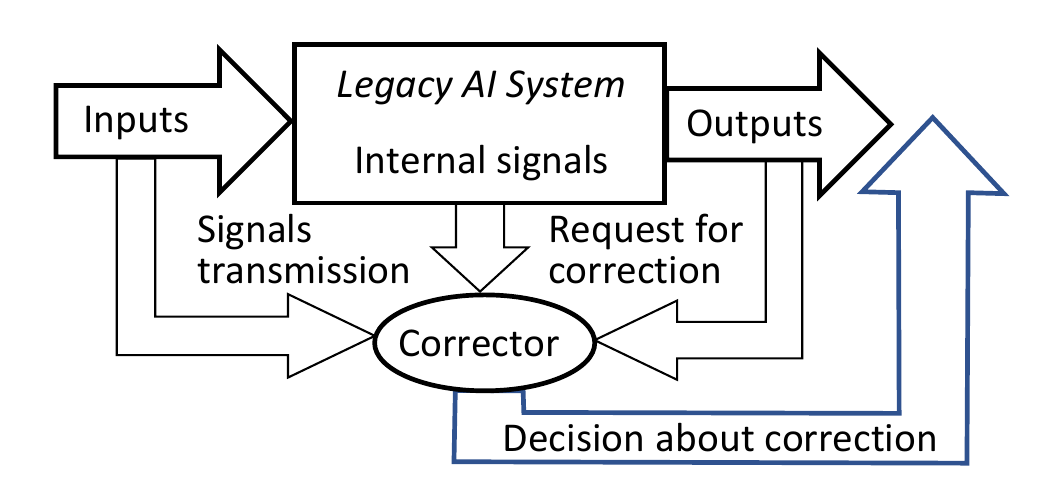}
\caption{A scheme of the operation of an elementary corrector of legacy AI systems.
 The elementary corrector receives the  input signals of legacy AI system, the internal signals generated by this system in the decision-making process, and its output signals. The corrector then assesses the need for correction. The elementary corrector includes a binary classifier that separates situations with a high risk of error from normal functioning.  If correction is required, the corrector sends a warning signal and a modified output for further use. \label{SingleCorrector}}
\end{figure} 

The universal part of the AI corrector is a classifier that should separate  situations  with erroneous behaviour from normal operation. It is a binary classifier for all types of AI. The generalisation ability of this classifier is its ability to recognise errors that it had never seen before. The training set for corrector consists of a collection of situations with normal operation of the legacy AI system (the `normal' class) and a set of labelled errors. The detection and labelling of errors for training correctors can be performed by various methods, which include human inspection, decisions of other AI systems of their committees, signals of success or failure from the outer world, and other possibilities that are outside the scope of our work.

We can usually expect that a normal class of error-free operations includes many more examples than a set of labelled errors. Moreover, even the situation with one newly labelled error is of considerable interest. All the stochastic separation theorems were invented to develop the one- of few-shot learning rules for the binary error/normal operation classifiers.

A specific component of the AI corrector is the modified decision rule (the `correction' itself). Of course, the general theory and algorithms are focused on the universal part of the correctors. For many classical families of data distributions, it is proved that the well-known Fisher discriminant is surprisingly a powerful tool for constructing correctors if the dimension of the data space is sufficiently high (most results of this type are collected in \cite{Grechuk2021}). This is proven for a wide class of distributions, including log-concave distributions, their convex combinations, and product distributions with bounded support. 

In this article, we refuse the classical hypothesis of the regularity of the data distribution and assume that the data can have a rich fine-grained structure with many clusters and corresponding peaks in the probability density. Moreover, the very notion of probability distribution in high dimensions may sometimes create more questions than answers.
Therefore, after developing new stochastic separation theorems for data with fine-grained clusters, we present a possibility to substitute the probabilistic approach to foundations of the theory by more robust methods of functional analysis with  the limit transition to infinite dimension. 

The idea of the presence of fine-grained structures in data seems to be very natural and universal: the observable world consists of things. The data points represent situations. The qualitative  difference between situations is in existence/absence of notable things there.

Many approaches to machine learning are based on the correction of errors. A well-known example is the backpropagation of errors, from the classical perceptron  algorithms~\cite{Rosenblatt1962} to modern deep learning \cite{Goodfellow2016}. The need for correction of AI errors has been discussed in the reinforcement learning literature. In the area of model-based reinforcement learning, the motivation stems from inevitable discrepancies between the models of environments used for training an agent and the reality this agent operates in. In order to address the problem, a meta-learning approach, Hallucinated Replay, was suggested in~\cite{Talvitie2014}. In this approach, the agent is trained to predict correct states of the real environment from states generated by the model \cite{Venkatraman2015}. Formal justifications and performance bounds for Hallucinated Replay were established in \cite{Talvitie2017}. Notwithstanding these successful developments, we note that the settings to which such strategies apply are largely Markov Decision Processes. Their practical relevance is therefore constrained by dimensionality of the system's state. In high dimension, the costs of exploring all states grows exponentially with dimension and, as a result, alternative approaches are needed.  
Most error correction algorithms use large training sets to prevent new errors from being created in situations where the system was operating normally. These algorithms are iterative in nature. On the contrary, the corrector technology in high dimension aims at non-iterative one- or few--shot error corrections.

\subsection{One- and Few-Shot Learning}

A set of labelled errors is needed for creation of AI corrector. If we have such a set, then the main problem is the fast non-iterative training of classifiers that separate situations with a high risk of error from situations in which the legacy AI system works well.   Thus, the corrector problem includes the one- or few-shot learning problem, and one class is presented by a relatively small sample of errors.

Learning new concepts rapidly from small low-sample data is a key challenge in machine learning \cite{VinyalsOneshot2016}. 
 Despite the widespread perception of neural networks as monstrous giant systems, whose iterative training requires a lot of time and resources, mounting empirical evidence points to numerous successful examples of learning from modestly-sized datasets \cite{Zhang2021}. 
Moreover,  training with one or several shots is possible. By definition, which has already become classic, ``one-shot learning'', consists of learning a class from a single labelled example \cite{VinyalsOneshot2016}. In  ``few-shot learning''  a classifier must generalise to new classes not seen in the training set, given only a small number of examples of each new class \cite{Snell2017}.
 
 Several modern approaches to enabling this type of learning require preliminary training tasks that are similar but not fully identical to new tasks to be learned. After such preliminary training the system acquires new meta-skills:  it can now learn new tasks, which are not crucially different from the previous ones, without the need for large training sets and training time. This heuristic is utilised in various constructions of one- and few-shot learning algorithms \cite{Sachin2017,Wang2020}.  Similar meta-skills and learnability can also be gained through previous experience of solving various relevant problems or an appropriately organised meta-learning \cite{Snell2017,Sung2018}.  

In general, a large body of one- and few-shot learning algorithms is based  on combinations of a reasonable preparatory learning  that aims to increase learnability and create meta-skills and simple learning routines facilitating learning from small number of examples after this propaedeutics. These simple methods create appropriate latent feature spaces for the trained models which are preconditioned for the task of learning from few or single examples. Typically, a copy of the same  pretrained system is used for different   one- and few-shot learning tasks. Nevertheless,  plenty of approaches are applicable to few-shot minor modifications of the features using new tasks.
 
Despite a large number of different algorithms implementing one- and few-shot learning schemes have been proposed to date, effectiveness of one- and few-shot simple methods  is based on either significant {\em {dimensionality reductions}} or the {\em {blessing of dimensionality}}  effects  \cite{GorTyukPhil2018,TyukinIJCNN2021}. 

A significant reduction in dimensionality means that several features have been extracted that are already sufficient for the purposes of learning.  Thereafter, a well-elaborated library of efficient lower-dimensional statistical learning methods can be applied to solve  new problems using the same features. 

The blessing of dimensionality is a relatively new idea \cite{Kainen1997, Donoho2000, AndersonEtAl2014, GorbanTyuRom2016}. It means that  simple classical techniques like linear Fisher's discriminants become unexpectedly powerful in high dimensions under some mild assumptions about regularity of probability distributions~\cite{GorbanTyukinNN2017,Gorbetal2018,GorbMakTyuk2019}. These assumptions typically require absence of extremely dense lumps of data, which can be defined as  areas with relatively low volume but unexpectedly high probability (for more detail we refer to \cite{Grechuk2021}). These lumps correspond to  narrow but high peaks of probability density. 

If a dataset consists of $k$ such lumps then, for moderate values of $k$,  this can be considered as a special case of dimensionality reduction. The centres of clusters are considered as `principal points' to stress the analogy with principal components \cite{Flury1990, GorbanZin2010}. Such a clustered structure in system's latent space may emerge in the course of preparatory learning:  images of data points in the latent space   {\em 'attract similar and repulse dissimilar'} data points. 
  
 The  one- and few-shot learning can be organised in all three situations described above: 
 \begin{enumerate}
 \item If the feature space is effectively reduced, then the challenge of large dataset can be mitigated and we can rely on  classical linear or non-linear methods of statistical learning.
 \item In the situation of `blessing of dimensionality', with sufficiently regular probability distribution in high dimensions, the simple linear (or kernel \cite{TyukinKernel2019})  one- and few-shot methods become effective \cite{GorbMakTyuk2019,Grechuk2021,TyukinIJCNN2021}.
 \item  If the data points in the latent space form dense clusters, then position of new data with respect to  these clusters can be utilised for solving new tasks. We can also expect that new data may introduce new clusters, but persistence of the cluster structure seems to be important. The clusters themselves can be distributed in a multidimensional feature space. This is the novel and more general setting we are going to focus on below in Section~\ref{StochFineGrain}.
 \end{enumerate}
 
 There is a rich set of tools for dimensionality reduction. It includes the classical prototype, principal component analysis (PCA) (see, \cite{Joliffe2011, GorbanZin2010} and Appendix~\ref{PCAy}), and many generalisations, from principal manifolds \cite{GorbanKegl2008} and kernel PCA \cite{Scholkopf1998} to principal graphs~\cite{GorbanZin2010,GorbanZin2010b} and autoencoders \cite{Kramer1991,HintonSalakh2006}. We briefly describe some of these elementary tools in the context of data preprocessing (Appendix~\ref{PrePostclassical}), but the detailed analysis of dimensionality reduction is out of the main scope of the paper.

In a series of previous works, we focused on the second item \cite{GorbanTyuRom2016,GorbanTyukinNN2017,Gorbetal2018,GorbMakTyuk2019,Grechuk2021,
GorTyukPhil2018,GorbMakTyuk2020}. The blessing of dimensionality effects that make the one- and few-shot learning possible for regular distributions of data are based on the stochastic separation theorems. All these theorems have a similar structure: for large dimensions, even in an exponentially large (relatively to the dimension) set of points, each point is separable from the rest by a linear functional, which is given by a simple explicit formula. These blessings of dimensionality phenomena are closely connected to the concentration of measure \cite{GianMilman2000, Gromov2003, Ledoux2005, Vershynin2018, Wainwright2019} and to the various versions of the central limit theorem in probability theory \cite{Kreinovich2021}. Of course, there remain open questions about sharp estimates for some distribution classes, but the general  picture seems to be clear now. 
 
In this work, we focus mainly on the third point and explore the blessings of dimensionality and related methods of one- and few-shot learning  for multidimensional data with rich cluster structure.  Such datasets cannot be described by regular probability densities with a priori bounded Lipschitz constants. Even more general assumptions  about absence of sets with relatively small volume but relatively high probability fail. We believe that this option is especially important for applications.

\subsection{Bibliographic Comments}

All references presented in the paper matter. However, a separate quick guide to the   bibliographic references about the main ideas may be helpful:
\begin{itemize} 
\item{\em {Blessing of dimensionality.} } In data analysis, the idea of blessing of dimensionality was formulated by Kainen \cite{Kainen1997}. Donoho  considered the effects of the dimensionality blessing to be the main direction of the development of modern data science \cite{Donoho2000}. The mathematical backgrounds of blessing of dimensionality are in the measure concentration phenomena. The same phenomena form the background of statistical physics (Gibbs, Einstein, Khinchin---see the review \cite{GorTyukPhil2018}). Two modern books include most of the classical results and many new achievements of concentration of measure phenomena needed in data science \cite{Vershynin2018,Wainwright2019} (but they do not include new stochastic separation theorems). Links between the blessing of dimensionality and the classical central limit theorems are  recently discussed in \cite{Kreinovich2021}.
\item{\em One-shot and few-shot learning.} This is a new direction in machine learning. Two papers give a nice introduction in this area \cite{VinyalsOneshot2016,Zhang2021}. Stochastic separation theorems explained ubiquity of one- and few-shot learning \cite{TyukinIJCNN2021}.
\item{\em AI errors.} The problem of AI errors is widely recognised. This is becoming the most important issue of serious concern when trying to use AI in real life. The Council of Europe Study report \cite{Yeung2019} demonstrates that the inevitability of errors of data-driven AI is now a big problem for society. Many discouraging examples of such errors are published \cite{Foxx2018, Strickland2019}, collected in reviews \cite{Banerjee2020},  and accumulated in a special database, Artificial Intelligence Incident Database (AIID) \cite{AIID,  PartnershipOnAI}. The research interest to this problem increases as an answer of the scientific community to the request of AI users. There are several fundamental origins of AI errors including uncertainty in training data, uncertainty in training process, and uncertainty of real world---reality can deviate significantly from the fitted model. The systematic manifestations of these deviations are known as concept drift or model degradation phenomena \cite{Tsymbal2004}.  
\item{\em AI correctors.} The idea of elementary  corrector together with statistical foundations was proposed in \cite{GorbanTyuRom2016}. First stochastic separation theorems were proved for several simple data distributions (uniform distributions in a ball and product distributions with bounded support)  \cite{GorbanTyukinNN2017}. The collection of results for many practically important classes of distributions, including convex combinations of log-concave distributions is presented in \cite{Grechuk2021}. Kernel version of stochastic separation theorem was proved \cite{TyukinKernel2019}. The stochastic separation theorems were used for development of correctors tested on various data and problems, from the straightforward  correction of errors \cite{Gorbetal2018} to knowledge transfer between AI systems \cite{TyukinAtAlKnowledge2018}.
\item{\em Data compactness.}~This is an old and celebrated idea proposed by Braverman in early 1960s \cite{Braverman1967}. Several methods of measurement compactness of data clouds were invented  \cite{Duin1999}. The possibility to replace data points by compacta in training of neural networks was discussed \cite{Kainen2004compacta}. Besides theoretical backgrounds of  AI and data mining,  data compactness was used for unsupervised outlier detection in high dimensions~\cite{Rehman2021} and other practical needs.      
\end{itemize}

\subsection{The Structure of the Paper}

 In Section~\ref{Postclassical} we briefly discuss the phenomenon of post-classical data. We begin with Donoho's definition of post-classical data analysis problems, where the number of attributes is greater than the number of data points \cite{Donoho2000}. Then we discuss alternative definitions and end with a real case study  that started with a dataset in the dimension $ 5 \times 10^5 $ and ended with five features that give an effective solution to the initial classification problem.

Section~\ref{StochFineGrain} includes the main theoretical results of the paper, the stochastic separation theorems for the data distributions with fine-grained structure.
For these theorems, we model clusters by geometric bodies (balls or ellipsoids) and work with distributions of ellipsoids in high dimensions. The hierarchical structure of data universe is introduced where each data cluster has a granular internal structure, etc. Separation theorems in infinite-dimensional limits are proven under assumptions of compact embedding of patterns into data space.

In Section~\ref{Multicluster}, the algorithms (multi-correctors)  for corrections of AI errors that work for  multiple clusters of error are developed and tested.   For such datasets, several elementary correctors and a dispatcher are required, which distributes situations for analysis to the most appropriate elementary corrector.   In multi-corrector, each elementary corrector separates its own area of high-risk error situations and contains an alternative rule for making decisions in situations from this area. The input signals of the correctors are the input, internal, and output signals of the AI system to be corrected as well as any other available attributes of the situation. The system of correctors is controlled by a dispatcher, which is formed on the basis of a cluster analysis of errors and distributes the situations specified by the signal vectors between elementary correctors for evaluation and, if necessary, correction.
  
Multi-correctors are tested on the CIFAR-10 dataset. 
In this case study, we will illustrate how 'clustered' or 'granular universes' can arise in real data and show how a granular representation based multi-correctors structure can be used in challenging machine learning and Artificial Intelligence problems.
 These problems include learning new classes of data in legacy deep learning AI models and predicting AI errors. We present simple algorithms and workflows which can be used to solve these challenging tasks circumventing the needs for computationally expensive retraining.  We also illustrate potential technical pitfalls and dichotomies requiring additional attention from the algorithms' users and designers.

In conclusion, we briefly review the results (Section~\ref{Conclusion}). Discussion (Section~\ref{Discussion}) aims at explaining the main message: the success or failure of many machine learning algorithms, the possibility of meta-learning, and opportunities to learn continuously from relatively small data samples depend on the world structure. 
The capability of representing a real world situation as a collection of things with some features (properties) and relationships between these entities is the fundamental basis of knowledge of both humans and AI. 

Appendices  include auxiliary mathematical results and relevant technical information.
In particular, in Appendix~\ref{PrePostclassical} we discuss the following preprocessing operations that may move the dataset from the postclassical area:
\begin{itemize}
\item Correlation transformation that maps the dataspace into cross-correlation space between data samples:
\item PCA;
\item Supervised PCA;
\item Semi-supervised PCA;
\item Transfer Component Analysis (TCA);
\item The novel expectation-maximization Domain Adaptation PCA (`DAPCA').
\end{itemize}

\section{Postclassical Data \label{Postclassical}}

High-dimensional post-classical world was defined in \cite{Donoho2000} by the inequality
\begin{equation}\label{DonohoPost}
\mbox{The number of attributes }  d \gg  \mbox{ The number of examples } N.
\end{equation}

This post-classical world is different from the `classical world', where we can consider infinite growth of the sample size for the given number of attributes. 
The classical statistical methodology was developed for the classical world based on the assumption of $$d < N \mbox{ and } N \to \infty.$$ 
 
Thus, the classical statistical learning  theory is mostly  useless in the multidimensional post-classical world.
These results all fail if $d > N$. The $d > N$ case is not anomalous for the modern big data problems. It is the generic case: both the sample size and the number of attributes grow, but in many important cases the number of attributes grows faster than the number of labelled examples \cite{Donoho2000}.

High-dimensional effects of the curse and blessing of dimensionality appear in a much wider area than specified by the inequality (\ref{DonohoPost}). A typical example gives the penomenon of quasiorthogonal dimension \cite{Kurkova1993,KainenKurkova2020,GorbTyuProSof2016}: for a given $\varepsilon >0$ and $\vartheta >0$ (assumed small) a random set of $N$ vectors  $\boldsymbol{x}_i$ on a high-dimensional  unit $d$-dimensional sphere satisfies the inequality $$|(\boldsymbol{x}_i,\boldsymbol{x}_j)|<\varepsilon$$ for all $i \neq j$ with probability $p>1-\vartheta$ when $N<a\exp(bd)$ and $a$ and $b$ depend  on $\varepsilon$ and $\vartheta$ only. This means that the quasiorthogonal dimension of an Euclidean space grows exponentially with dimension $d$.
Such effects are important in machine learning \cite{GorbTyuProSof2016}. Therefore, the Donoho boundary should be modified: the postclassical effects appear in high dimension when
\begin{equation}\label{logarithmicPost}
d\gg log N.
\end{equation}

\textls[-15]{The two different definitions  of postclassical area, (\ref{DonohoPost}) and  (\ref{logarithmicPost}),  are illustrated in \mbox{Figure~\ref{Fig:Post}a}.}

The definition of the postclassical data world needs one more comment. The inequalities (\ref{DonohoPost}) and (\ref{logarithmicPost}) used the number of attributes as the equivalent of the dimension of the data space. Behind this approach is the hypothesis that there is no strong dependency between attributes. In the real situations, the data dimensionality can be much less that the number of attributes, for example, in the case of the strong multicollinearity. If, say,  the data are located along a straight line then for most approaches  the dimension of the  dataset  is 1 and the value of $d$ does not matter. Therefore, the definition  (\ref{logarithmicPost}) of the postclassical world needs to be modified further with the dimension of the dataset, $\dim(DataSet)$ instead of $d$:
\begin{equation}\label{PostClassDim}
\dim(DataSet) \gg \log N.
\end{equation}

There are many various definitions of data dimensionality, see a brief review in \cite{Camastra2003,Bac2020}. For all of them, we can assume that  $\dim(DataSet)<N$ and $\dim(DataSet)\leq d$ (see \mbox{Figure~\ref{Fig:Post}b}). It may happen that the intrinsic dimensionality of the datasets is surprisingly low and  variables have hidden interdependencies.  The structure of multidimensional data point clouds can have globally complicated organisation which is sometimes difficult to represent with regular mathematical objects (such as manifolds) \cite{Bac2020, Albergante2020}.  

\begin{figure} [H]
\includegraphics[width=0.4\textwidth]{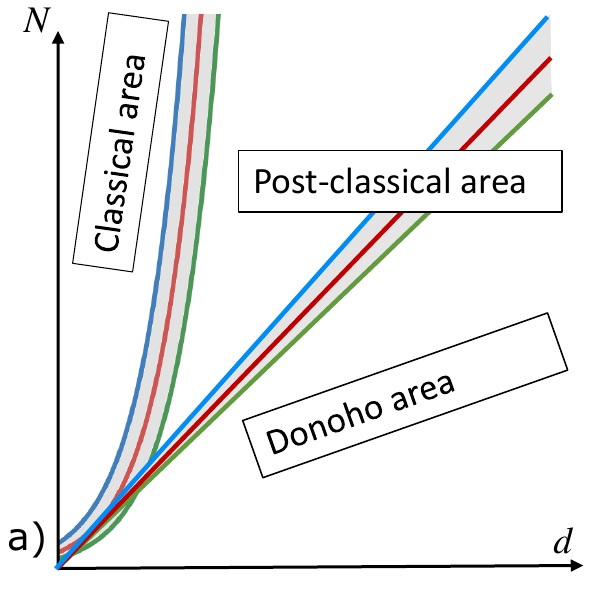}\;\;\;\;
\includegraphics[width=0.4\textwidth]{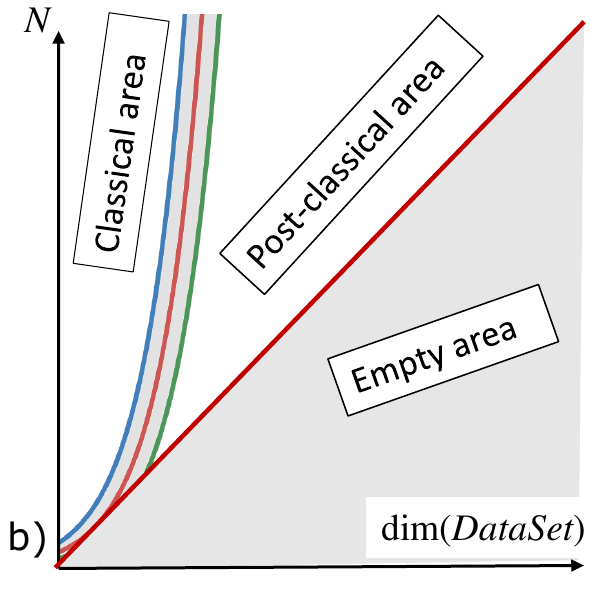}
\caption{\label{Fig:Post}Different zones of data world: (\textbf{a}) Separation of Donoho's postclassical data world, where $d>N$ (below the bisector), the classical world, where $d\ll log N$ and the `postclassical' area below the exponent, $d\gg log N$; (\textbf{b})  Classical and postclassical data worlds according to the definition (\ref{PostClassDim})  (the area below the bisector is empty).  The gray areas around the borders between the different areas symbolise the fuzziness of the borders. Here, $d$ is the number of attributes, $N$ is the number of samples, and  $\dim(DataSet)$ is the intrinsic dimensionality of the dataset, $d\geq \dim(DataSet)$ and $N> \dim(DataSet)$.} 
\end{figure} 

The  postclassical world effects include the blessing and curse of dimensionality.  The blessing and curse are based on the concentration of measure phenomena \cite{GianMilman2000, Gromov2003, Ledoux2005, Vershynin2018} and are, in that sense, two sides of the same coin \cite{GorbMakTyuk2019,GorbMakTyuk2020}.
  
It may be possible to resolve the difficulties with the data analysis in Donoho area by adequate preprocessing described in Appendix~\ref{PrePostclassical}. Consider an  example of successful descent from data dimension $5\times 10^5$ to five-dimensional decision space \cite{Moczko2016}.
The problem was to develop an `optical tongue' that recognises toxicity of various substances. The optical assay included a mixture of sensitive fluorescent dyes and human skin cells. They generate fluorescence spectra distinctive for particular conditions. 
The system produced characteristic response to toxic chemicals.

Two fluorescence images were received for each chemical: with growing cells and without them (control). The images were $511 \times 511$ arrays of fluorescence intensities as functions of emission and excitation. The dataset included 34 irritating and 28 non-irritating (Non-IRR) compounds (62 chemicals in total).  The input data vector for each compound had dimension 522,242. This dataset belonged to the Donoho area.

After selection of a training set, each fluorescence image was represented by the vector of its correlation coefficients with the images from the training set. The size of the training set was 43 examples (with several randomised training set/test set splittings) or 61 example (for leave one out cross-validation). After that, the data matrix was $43 \times 43$ or $61 \times 61 $ symmetric matrix.  Then the classical PCA was applied with the standard selection of the number of components by Kaiser rule that returned five components.  Finally, in the reduced space the classical classification algorithms were applied (kNN, decision tree, linear discriminant, and other). Both sensitivity and specificity of the 3NN classifiers with adaptive distance and of decision tree exceeded 90\% in leave one out cross-validation. 

This case study demonstrates that simple preprocessing can sometimes return postclassical data to the classical domain. However, in truly multidimensional datasets, this approach can fail due to the quasiorthogonality effect \cite{Kurkova1993,KainenKurkova2020,GorbTyuProSof2016}: centralised random vectors in large dimensions are nearly orthogonal under very broad assumptions, and the matrix of empirical correlation coefficients with high probability is often close to the identity matrix even for exponentially large data samples \cite{GorbTyuProSof2016}.

\section{Stochastic Separation for Fine-Grained Distributions \label{StochFineGrain}}

\subsection{Fisher Separability}

Recall that the classical Fisher discriminant between two classes with means $\boldsymbol{\mu}_1$ and $\boldsymbol{\mu}_2$ is separation of the classes by a hyperplane orthogonal to $\boldsymbol{\mu}_1-\boldsymbol{\mu}_2$ in the inner product 
\begin{equation}\label{Fisher1}
 \langle \boldsymbol{ x},\boldsymbol{ y}\rangle=( \boldsymbol{ x}, \boldsymbol{S}^{-1} \boldsymbol{ y}),
 \end{equation}
 where $( \cdot, \cdot )$ is the standard inner product and $ \boldsymbol{S}$ is the average (or the weighted average) of the sample covariance matrix of these two classes. 

Let the dataset be preprocessed. In particular, we assume that it is {\em centralised, normalised, and approximately whitened}. In this case, we use in the definition of Fisher's discriminant the standard inner product instead of $\langle \cdot , \cdot \rangle$.

\begin{Definition}\label{def:Fisher0}
A point $\boldsymbol{ x}$ is \emph{Fisher separable} from a set $\mathbf{Y} \subset {\mathbb R}^n$ with threshold  $\alpha \in (0, 1]$, or $\alpha$-Fisher separable in short, if inequality
\begin{equation}\label{eq:Fisher}
\alpha(\boldsymbol{ x},\boldsymbol{ x}) \geq (\boldsymbol{ x},\boldsymbol{ y}), 
\end{equation}
holds for all $\boldsymbol{ y}\in \mathbf{Y}$.  
\end{Definition}
  
\begin{Definition}\label{def:Fisher}
A finite set $\mathbf{Y} \subset {\mathbb R}^n$ is \emph{Fisher separable} with threshold $\alpha \in (0, 1]$, or $\alpha$-Fisher separable in short, if inequality (\ref{eq:Fisher}) 
holds for all $\boldsymbol{ x}, \boldsymbol{ y} \in \mathbf{Y} $ such that $\boldsymbol{ x}\neq  \boldsymbol{ y}$. 
\end{Definition}

Separation of points by simple and explicit inner products (\ref{eq:Fisher}) is, from the practical point of view, more convenient than general linear separability that can be provided by support vector machines, for example. Of course, linear separability is more general than Fisher separability. This is obvious from the everyday low-dimensional experience,  but in high dimensions Fisher separability becomes a generic phenomenon \cite{GorbanTyuRom2016,GorbanTyukinNN2017}.

Theorem~\ref{Th:prototype} below is a prototype of most stochastic separation theorems. 
Two heuristic conditions for the probability distribution of data points are used in the stochastic separation theorems:
\begin{itemize}
\item   The probability distribution has no heavy tails;
\item The sets of relatively small volume should not have large probability.
\end{itemize}

These conditions are not necessary and could be relaxed \cite{Grechuk2021}.

In the following Theorem~\ref{Th:prototype} \cite{Gorbetal2018} the absence of heavy tails is formalised as the tail cut: the support of the distribution is a subset of the $n$-dimensional unit ball $\mathbb{B}_n$.
The absence of the sets of small volume but large probability is formalised in this theorem by the inequality:
\begin{equation}\label{bounded}
\rho(\boldsymbol{x})<\frac{C}{r^n V_n(\mathbb{B}_n)},
 \end{equation}
where $\rho$ is the distribution density, $C>0$ is an arbitrary constant, $V_n(\mathbb{B}_n)$ is the volume of the ball $\mathbb{B}_n$ and   $1>r>1/(2\alpha)$. This inequality guarantees that the probability measure of each ball with the radius  $R\leq 1/(2\alpha)$  decays for $n\to \infty$ in a geometric progression with denominator $R/r$.  Condition   $1>r>1/(2\alpha)$ is possible only if $\alpha > 0.5$, hence, in \mbox{Theorem~\ref{Th:prototype}} we assume $\alpha \in (0.5,1]$.

\begin{Theorem}[\cite{Gorbetal2018}]\label{Th:prototype}
Let $1\geq \alpha >1/2$, $1>r>1/(2\alpha)$,  $1>\delta>0$, $Y\subset \mathbb{B}_n$ be a finite set, $|Y|<\delta (2r\alpha)^n/C$  and $\boldsymbol{x}$ be a randomly chosen point from a distribution in the unit ball with the bounded probability density $\rho(\boldsymbol{x})$. Assume that $\rho(\boldsymbol{x})$ satisfies inequality (\ref{bounded}). Then with probability $p>1-\delta$ point $\boldsymbol{x}$ is Fisher-separable from $Y$ with threshold $\alpha$ (\ref{eq:Fisher}).
\end{Theorem}
\begin{proof}
For a given $\boldsymbol{ y}$, the  set of such  $\boldsymbol{ x}$ that $\boldsymbol{ x}$ is not  $\alpha$-Fisher separable from $\boldsymbol{ y}$ by inequality~\mbox{(\ref{eq:Fisher})}  is a ball given by  inequality (\ref{eq:Fisher})
\begin{equation}\label{excludedvolume}
\left\{\boldsymbol{z} \ \left| \ \left\|\boldsymbol{z}-\frac{\boldsymbol{y}}{2\alpha }\right\|< \frac{\|\boldsymbol{y}\|}{2\alpha} \right.  \right\}.
\end{equation}

This is the ball of excluded volume. The volume of the ball (\ref{excludedvolume}) does not exceed $V=\left(\frac{1}{2\alpha}\right)^nV_n(\mathbb{B}_n)$ for each $\boldsymbol{y}\in Y$. The probability that point  $\boldsymbol{x}$ belongs to such a ball does not exceed
$$V \sup_{z\in \mathbb{B}_n }\rho(z)\leq C \left(\frac{1}{2r \alpha}\right)^n.$$

The probability that $\boldsymbol{x}$   belongs to the union of $|Y|$ such balls does not exceed $|Y| C\left(\frac{1}{2r\alpha}\right)^n$. For $|Y|<\delta (2r\alpha)^n/C$ this probability is smaller than $\delta$ and $p>1-\delta$.
\end{proof}

Note that:
\begin{itemize}
\item The finite set $Y$ in Theorem~\ref{Th:prototype} is  just a finite subset of the ball $ \mathbb{B}_n$ without any assumption of its randomness.  We only used the assumption about distribution of  $\boldsymbol{x}$.
\item The distribution of  $\boldsymbol{x}$ may deviate significantly from the uniform distribution in the ball $ \mathbb{B}_n$. Moreover, this deviation may grow with dimension $n$ as a geometric progression:
$$\rho(\boldsymbol{x})/\rho_{\rm uniform}\leq {C}/{r^n},$$
where $\rho_{\rm uniform}=1/ V_n(\mathbb{B}_n)$ is the density of uniform distribution and $1/(2\alpha)<r<1$ under assumption that  $1/2<\alpha  \leq 1$.
\end{itemize}
 
Let, for example, $\alpha = 0.8$, $r=0.9$, $C=1$, $\delta=0.01$.   Table~\ref{Table1} shows the upper bounds on  $|Y|$ given by Theorem~\ref{Th:prototype} in various dimensions $n$ that guarantees $\alpha$-Fisher separability of a random point $\boldsymbol{x}$ from $Y$  with probability $\geq 0.99$ if the ratio $\rho(\boldsymbol{x})/\rho_{\rm uniform}$ is bounded by the geometric progression $1/r^n$.   
 \begin{table}[h]  
\begin{center}
\caption{\label{Table1} The upper bound on $|Y|$ that guarantees separation of $\boldsymbol{x}$ from $Y$  by Fisher's discriminant with probability 0.99 according to Theorem~\ref{Th:prototype} for  $\alpha = 0.8$, $r=0.9$, $C=1$ in various dimensions.}
\begin{tabular}{ rllllll } 
 \hline
 {$n$} &  {10} & {25} & {50} & {100} & {150} & {200} \\
 \midrule
$|Y|\leq$& 0.38& $91$ & $8.28 \times 10^5$ & $6.85 \times 10^{13}$ 
& $5.68 \times 10^{21}$ & $4.70 \times 10^{29}$ \\
 $\rho(\boldsymbol{x})/\rho_{\rm uniform}\leq$ &
 2.86 & 13.9 & 194 & $3.76 \times 10^4$ & $7.30 \times 10^6$ & $1.41 \times 10^9$\\
 \hline
\end{tabular}
\end{center}
\end{table}

For example, for $n=100$, we see that for any set with $|Y|<6.85 \times 10^{13}$ points in the unit ball and any distribution whose density $\rho$ deviates from the uniform one by a factor at most $3.76 \times 10^4$, a random point from this distribution is Fisher-separable (\ref{def:Fisher}) with $\alpha=0.8$ from all points in $Y$ with $99\%$ probability.

If we consider $Y$ as a random set in $\mathbb{B}_n$ that satisfies (\ref{bounded}) for each point then with high probability  $Y$ is  $\alpha$-Fisher separable (each point from the rest of $Y$) under some constraints of $|Y|$ from above.
From Theorem~\ref{Th:prototype} we get the following corollary.
\begin{Corollary}\label{Cor:1}
If  $Y \subset \mathbb{B}_n$ is a random set $Y= \{\boldsymbol{y}_1, \ldots , \boldsymbol{y}_{|Y|}\}$ and for each $j$ the conditional distributions of vector $\boldsymbol{y}_j$ for any given positions of the other  $\boldsymbol{y}_k$ in $\mathbb{B}_n$    satisfy the same conditions as the distribution of $\boldsymbol{x}$ in  Theorem~\ref{Th:prototype}, then the probability of the random set $Y$ to be $\alpha$-Fisher separable can be easily estimated: 
$$p\geq 1- |Y|^2 C\left(\frac{1}{2r\alpha}\right)^n.$$
\end{Corollary}
Thus, let us take, for example, $p>0.99$ if $|Y|<(1/10)\,C^{-1/2}(2r\alpha)^{n/2}$ (Table~\ref{Table2}). 
\begin{table}[h]  
\begin{center}
\caption{\label{Table2} The upper bound on $|Y|$ that guarantees $\alpha$-Fisher's separability of $Y$ with probability $\geq 0.99$ according to Corollary~\ref{Th:prototype} for  $\alpha = 0.8$, $r=0.9$, $C=1$ in various dimensions.}
\begin{tabular}{ r|llllll } 
 \hline
 {$n$} &  {10} & {25} & {50} & {100} & {150} & {200} \\
 \midrule
$|Y|\leq$& 0.61& $9.5$ & $910$ & $8.28 \times 10^{6}$ 
& $7.53 \times 10^{10}$ & $6.85 \times 10^{14}$ \\
 \hline
\end{tabular}
\end{center}
\end{table}

Multiple generalisations of Theorem~\ref{Th:prototype} are proven with sharp estimates of $|Y|$ for various families of probability distributions. In this section, we derive the stochastic separation theorems for distributions with cluster structure that violate significantly the assumption (\ref{bounded}). For this purpose, in the  following subsections we introduce models of cluster structures and modify the notion of Fisher separability to separate clusters. The structure of separation functionals remains explicit with  a one-shot non-iterative learning but assimilates both information about the entire distribution and about the cluster being separated.

\subsection{Granular Models of Clusters}

The simplest model of a fine-grained distribution of data assumes that the data are grouped into dense clusters and each cluster is located inside a relatively small body (a granule) with random position. Under these conditions,  the distributions of data inside the small granules do not matter and may be put out of consideration. What is important is the geometric characteristics of the granules and their distribution. This is a simple one-level version of the granular data representation \cite{Zadeh1997, Pedrycz2008}. The possibility to replace points by compacts in neural network learning was considered by Kainen \cite{Kainen2004compacta}. He developed the idea that 'compacta can replace points'. In discussion, we will touch also a promising multilevel hierarchical granular representation.
 
Spherical granules allows a simple straightforward generalisation of Theorem~\ref{Th:prototype}. Consider spherical granules $G_z$ of radius $R$ with centres $\boldsymbol{z}\in \mathbb{B}_n$: 
$$G_z=\{\boldsymbol{z}'\,|\, \|\boldsymbol{z}'-\boldsymbol{z}\|\leq R\}.$$

  Let $G_x$ and $G_y$ be two such granules. Let us reformulate the Fisher separation condition with threshold $\alpha$  for granules:
\begin{equation}\label{SphereFisher1}
\alpha(\boldsymbol{x},\boldsymbol{x}')\geq  (\boldsymbol{x},\boldsymbol{y}') \mbox{  for all  } \boldsymbol{x}' \in G_x, \; \boldsymbol{y}' \in G_y .
\end{equation}
 
Elementary geometric reasoning gives that  the separability condition (\ref{SphereFisher1}) holds if $\boldsymbol{x}$ (the centre of $ G_x$) does not belong to the ball with radius $\frac{1}{2\alpha} \|\boldsymbol{y}\|+R(1+\frac{1}{\alpha})$ centred at $\frac{1}{2\alpha} \boldsymbol{y}$:
\begin{equation}\label{ExclCluster}
\boldsymbol{x} \notin  \left\{\boldsymbol{z} \ \left| \ \left\|\boldsymbol{z}-\frac{\boldsymbol{y}}{2\alpha }\right\|< \frac{\|\boldsymbol{y}\|}{2\alpha} +R\left(1+\frac{1}{\alpha}\right) \right.  \right\}.
\end{equation}

This is analogous to the ball of  excluded volume (\ref{excludedvolume}) for spherical granules. The difference from  (\ref{excludedvolume}) is that both $\boldsymbol{z}$ and $\boldsymbol{y}$ are inflated into balls of radius $R$.

Let $\mathbf{B}$ be the closure of the ball defined in  (\ref{excludedvolume}):
$$\mathbf{B}= \left\{\boldsymbol{z} \ \left| \ \left\|\boldsymbol{z}-\frac{\boldsymbol{y}}{2\alpha }\right\| \leq \frac{\|\boldsymbol{y}\|}{2\alpha} \right.  \right\}.$$

Condition (\ref{ExclCluster}) implies that the distance between $ \boldsymbol{x}$ and $\mathbf{B}$ is at least $R(1+\frac{1}{\alpha})$. In particular, $\|\boldsymbol{x}-\beta \boldsymbol{x}\| \geq R(1+\frac{1}{\alpha})$, where $\beta$ is the largest real number such that $\beta \boldsymbol{x} \in \mathbf{B}$. Then $\beta \boldsymbol{x}$ belongs to the boundary of $\mathbf{B}$, hence (\ref{eq:Fisher}) holds as an equality for $\beta  \boldsymbol{x}$:
$$
\alpha(\beta  \boldsymbol{x}, \beta  \boldsymbol{x}) = (\beta  \boldsymbol{x}, \boldsymbol{y}),
$$ 
or, equivalently, $\alpha\beta \| \boldsymbol{x}\|^2 = ( \boldsymbol{x}, \boldsymbol{y})$. Then
$$
\alpha(\boldsymbol{x},\boldsymbol{x}) = \alpha \|\boldsymbol{x}\| \cdot \|\boldsymbol{x}-\beta \boldsymbol{x}\| + \alpha \beta\|\boldsymbol{x}\|^2 \geq \alpha \|\boldsymbol{x}\| \cdot R\left(1+\frac{1}{\alpha}\right) + (\boldsymbol{x},\boldsymbol{y}) = (1+\alpha)R \|\boldsymbol{x}\|+(\boldsymbol{x},\boldsymbol{y}). 
$$ 

Thus, if $\boldsymbol{x}$ satisfies (\ref{ExclCluster}) then
\begin{equation}\label{eq:14corrected}
\alpha(\boldsymbol{x},\boldsymbol{x}) \geq (1+\alpha)R \|\boldsymbol{x}\|+(\boldsymbol{x},\boldsymbol{y}) \mbox{  that is  } \alpha((\boldsymbol{x},\boldsymbol{x}) -R \|\boldsymbol{x}\|) \geq  (\boldsymbol{x},\boldsymbol{y}) +R \|\boldsymbol{x}\|.
\end{equation}

Let  $\boldsymbol{x}' \in G_x, \; \boldsymbol{y}' \in G_y$.  The Cauchy–Schwarz inequality gives 
$|(\boldsymbol{x}'-\boldsymbol{x}, \boldsymbol{x})|\leq \|\boldsymbol{x}'-\boldsymbol{x}\| \| \boldsymbol{x}\|\leq R  \| \boldsymbol{x}\|$ and $|(\boldsymbol{y}'-\boldsymbol{y}, \boldsymbol{x})|\leq \|\boldsymbol{y}'-\boldsymbol{y}\| \| \boldsymbol{x}\|\leq R \| \boldsymbol{x}\|$. Therefore, 
$(\boldsymbol{x},\boldsymbol{x}')\geq (\boldsymbol{x},\boldsymbol{x})-R \| \boldsymbol{x}\|$ and
$(\boldsymbol{x},\boldsymbol{y})+R \| \boldsymbol{x}\| \geq (\boldsymbol{x},\boldsymbol{y}')$. Combination of two last inequalities with (\ref{eq:14corrected}) gives separability (\ref{SphereFisher1}).

If the point $\boldsymbol{y}$ belongs to the unit ball   $\mathbb{B}_n$ then the radius of the ball of excluded volume (\ref{ExclCluster}) does not exceed 
\begin{equation}\label{xi}
\xi=\frac{1}{2\alpha}+R\left(1+\frac{1}{\alpha}\right).
\end{equation}

 Further on, the assumption $\xi<1$ is used.

\begin{Theorem}\label{Th:SphericGranules} 
Consider a finite set of spherical granules $G_y$ with  radius $R$  and set of centres $Y$  in $\mathbb{B}_n$. Let $G_x$ be a granule  with  radius $R$ and a randomly chosen centre $\boldsymbol{x}$ from a distribution in the unit ball with the bounded probability density $\rho(\boldsymbol{x})$. Assume that $\rho(\boldsymbol{x})$ satisfies inequality (\ref{bounded}) and the upper estimate of the radius of excluded ball (\ref{xi}) $\xi<1$.  Let $1>r>\xi$ and 
\begin{equation}\label{ClusterBound}
|Y|<\delta\frac{1}{C} \left(\frac{r}{\xi}\right)^n .
\end{equation}

 Then the separability condition (\ref{SphereFisher1}) holds for    $G_x$ and all  $G_y$ ($\boldsymbol{y}\in Y$) with probability $p>1-\delta$.
\end{Theorem}
\begin{proof}
The separability condition (\ref{SphereFisher1}) holds for the granule  $G_x$ and all  $G_y$ ($\boldsymbol{y}\in Y$) if $\boldsymbol{x}$ does not belong to the excluded ball (\ref{ExclCluster})  for all $\boldsymbol{y}\in Y$. The volume of the excluded ball is  $V=\xi^n V_n(\mathbb{B}_n)$ for each $\boldsymbol{y}\in Y$.   The probability that point  $\boldsymbol{x}$ belongs to such a ball does not exceed $C\left(\frac{\xi}{r}\right)^n$ in accordance with the boundedness  condition (\ref{bounded}). Therefore, the probability that $\boldsymbol{x}$ belongs to the union of such balls does not exceed  $|Y|C\left(\frac{\xi}{r}\right)^n$. This probability is less than $\delta$ if $|Y|<\delta\frac{1}{C} \left(\frac{r}{\xi}\right)^n $.
\end{proof}

Table~\ref{Table3} shows how the number  $|Y|$ that guarantees separability (\ref{SphereFisher1}) of a random granule  $G_x$ from an  arbitrarily selected set of $|Y|$  granules with probability 0.99 grows with dimension for $\alpha = 0.9$, $r=0.9$, $C=1$ and $R=0.1$. 

\begin{table}[h]  
\begin{center}
\caption{\label{Table3} The upper bound on $|Y|$ that guarantees separation of granules $G_x$ and all  $G_y$ ($\boldsymbol{y}\in Y$)  (\ref{SphereFisher1}) with probability 0.99 according to Theorem~\ref{Th:SphericGranules}  for  $\alpha = 0.9$, $r=0.9$, $C=1$ and $R=0.1$ in various dimensions.}
\begin{tabular}{ r|llllll } 
 \hline
 {$n$} &  {25} & {50} & {100} & {150} & {200} \\
 \midrule
$|Y|\leq$&  $0.55$ & $30$ & $9.26 \times 10^4$ 
& $2.81\times 10^8$ & $8.58 \times 10^{11}$ \\
 \hline
\end{tabular}
\end{center}
\end{table}

The separability condition (\ref{SphereFisher1}) can be considered as Fisher separability (\ref{eq:Fisher})  with inflation points to granules. From this point of view, 
Theorem~\ref{Th:SphericGranules} is a version of  Theorem~\ref{Th:prototype} with inflated points. An inflated version of Corollary~\ref{Cor:1} also exists.

\begin{Corollary}\label{Cor:2Gran}
Let  $Y \subset \mathbb{B}_n$ be a random set $Y= \{\boldsymbol{y}_1, \ldots , \boldsymbol{y}_{|Y|}\}$. Assume that for each $j$ the density of conditional distribution  of vector $\boldsymbol{y}_j$ for any given positions of the other  $\boldsymbol{y}_k$ in $\mathbb{B}_n$  exists and  satisfies  inequality (\ref{bounded}). Consider a finite set of spherical granules $G_y$ with  radius $R$  and centres $\boldsymbol{y}\in Y$  in $\mathbb{B}_n$. For the radius of the excluded ball  (\ref{xi}) assume $\xi<r$, where $r<1$ is defined in (\ref{bounded}). Then, with probability 
$$p\geq 1- |Y|^2 C\left(\frac{\xi}{r}\right)^n$$
 for every two $\boldsymbol{x},\boldsymbol{y}\in Y$ ($\boldsymbol{x}\neq \boldsymbol{y}$) the separability condition (\ref{SphereFisher1}) holds.
Equivalently, it holds with probability $p>1-\delta$ ($\delta>0$) if
$$|Y|<\sqrt{\frac{\delta}{C}}\left(\frac{r}{\xi}\right)^{n/2}.$$

This upper border of $|Y|$ grows with $n$ in geometric progression.
\end{Corollary}

The idea of spherical granules implies that, in relation to the entire dataset,  the granules are more or less uniformly compressed in all directions and their diameter is relatively small (or, equivalently, the granules are inflated points, and this inflation is limited isotropically). Looking around, we can hypothesise quite  different properties: in some directions, the granules can have large variety, it can be as large of variety as the whole set, but the dispersion decays in the sequence of the granule's principal components while the entire set is assumed to be whitened. Large diameter of granules is  not an obstacle to the stochastic separation theorems. The following proposition gives a simple but instructive example. 
 
 \begin{Proposition}\label{Prop:Simplex}Let $1\geq \alpha >1/2$, $1>r>1/(2\alpha)$,  $1>\delta>0$. Consider an arbitrary set of $N$ intervals $I_j=[\boldsymbol{u_j}, \boldsymbol{v_j}] \in \mathbb{B}_n$ ($j=1, \ldots, N$). Let $\boldsymbol{x}$ be a randomly chosen point from a distribution in the unit ball with the bounded probability density $\rho(\boldsymbol{x})$. Assume that $\rho(\boldsymbol{x})$ satisfies inequality (\ref{bounded}) and $N<\frac{\delta}{2C} (2r\alpha)^n $. Then with probability $p>1-\delta$ point $\boldsymbol{x}$ is Fisher-separable from any $ \boldsymbol{y} \in \cup_j I_j$  with threshold $\alpha$ (\ref{eq:Fisher}).
 \end{Proposition}
 \begin{proof}
 For given $\boldsymbol{x}$ and $\alpha$, the  Fisher's separability inequality defines a half-space for  $\boldsymbol{y}$ (\ref{eq:Fisher}). An interval $I=[u,v]$ belongs to this half-space if and only if its ends, $u$ and $v$, belong to it, that is, $\boldsymbol{x} $ is $\alpha$-Fisher separable from $u$ and $v$. Therefore, we can apply Theorem~\ref{Th:prototype} to prove $\alpha$-Fisher separability of  $\boldsymbol{x} $  from the set $Y=\{u_j\}\cup \{v_j\}$, $|Y|=2N$.
 \end{proof}
 
The same statements are true for separation of a point from a set of simplexes of various dimension. For such estimates, only the number of vertices matters.
 
 Consider granules in the form of ellipsoids with decaying sequence of length of the principal axes. Let $d_1>d_2>....$ ($d_i>0$) be an infinite sequence of the upper bounds for semi-axes. Each ellipsoid granule in $\mathbb{R}^n$  has a centre,  $ \boldsymbol{z}$, an orthonormal basis of principal axes $E=\{\boldsymbol{e}_1, \boldsymbol{e}_2, \ldots , \boldsymbol{e}_n\}$, and a sequence of semi-axes, $A=\{a_1\geq a_2 \geq \ldots \geq a_n\} $ ($d_i\geq a_i>0$). This ellipsoid is given by the inequality:
 \begin{equation}\label{ellipsoid}
\mathbf{S}_{\boldsymbol{z}, E,A}=  \left\{\boldsymbol{z}' \left| \sum_{j=1}^n\frac{1}{a_j^2}(\boldsymbol{z}'-\boldsymbol{z},  \boldsymbol{e}_j)^2\leq 1\right\}\right. .
  \end{equation}
 
  Let the sequence  $d_1>d_2>....$ ($d_i>0$, $d_i\to 0$) be given.
 \begin{Theorem}\label{Th:EllGranules} 
Consider a set of $N$  elliptic granules (\ref{ellipsoid}) with centres  $\boldsymbol{z} \in \mathbb{B}_n$ and $a_i\leq d_i$.  Let $D$ be the union of all these granules.  Assume that $\boldsymbol{x}\in \mathbb{B}_n$ is a random point from a distribution in the unit ball with the bounded probability density $\rho(\boldsymbol{x})\leq \rho_{\max}$. Then for positive $ \varepsilon $, $\varsigma$
\begin{equation}\label{EllClusterBound}
{\bf P} ((\boldsymbol{x},\boldsymbol{z}') < \varepsilon  \mbox{  for all  }  \boldsymbol{z}'\in D, \;\; \& \;\; (\boldsymbol{x},\boldsymbol{x})>1-\varsigma)> 1 - N \rho_{\max}V_n(\mathbb{B}_n) a \exp(-bn),
\end{equation}
where $a$ and $b$ do not depend on the dimensionality.
\end{Theorem}
In proof of Theorem~\ref{Th:EllGranules} we construct explicit estimates of probability in (\ref{EllClusterBound}). This construction (Equation (\ref{Th3estimate}) below) is an important part of Theorem~\ref{Th:EllGranules}. It  is based on the following lemmas  about quasiorthogonality of random vectors.
\begin{Lemma}\label{Lem:QuasiOrth}
Let $\boldsymbol{e}\in  \mathbb{R}^n$ be any normalised vector, $\|  \boldsymbol{e}\|=1$.  Assume that $\boldsymbol{x}\in \mathbb{B}_n$ is a random point from a distribution in $\mathbb{B}_n$ with the bounded probability density   $\rho(\boldsymbol{x})\leq \rho_{\max}$.  Then, for any $\varepsilon>0$ the probability
\begin{equation}\label{ProbNonOrth}
{\bf P}((\boldsymbol{x},\boldsymbol{e})\geq \varepsilon)\leq \frac{1}{2} \rho_{\max}V_n(\mathbb{B}_n)(\sqrt{1-\varepsilon^2})^n.
\end{equation}
\end{Lemma}
\begin{proof}
The inequality $(\boldsymbol{x},\boldsymbol{e})\geq \varepsilon$ defines a spherical cap. This spherical cap can be estimated from above by the volume of a hemisphere of radius  $\sqrt{1-\varepsilon^2}$ (Figure~\ref{SphericalCap}). The volume $W$ of this hemisphere is 
$$W=\frac{1}{2}V_n(\mathbb{B}_n) (\sqrt{1-\varepsilon^2})^n$$

The probability that $\boldsymbol{x}$ belongs to this cap is bounded from above by the value $\rho_{max} W$, which gives the estimate (\ref{ProbNonOrth}). 
\end{proof} \clearpage

\begin{figure}[H]
\includegraphics[width=0.4\textwidth]{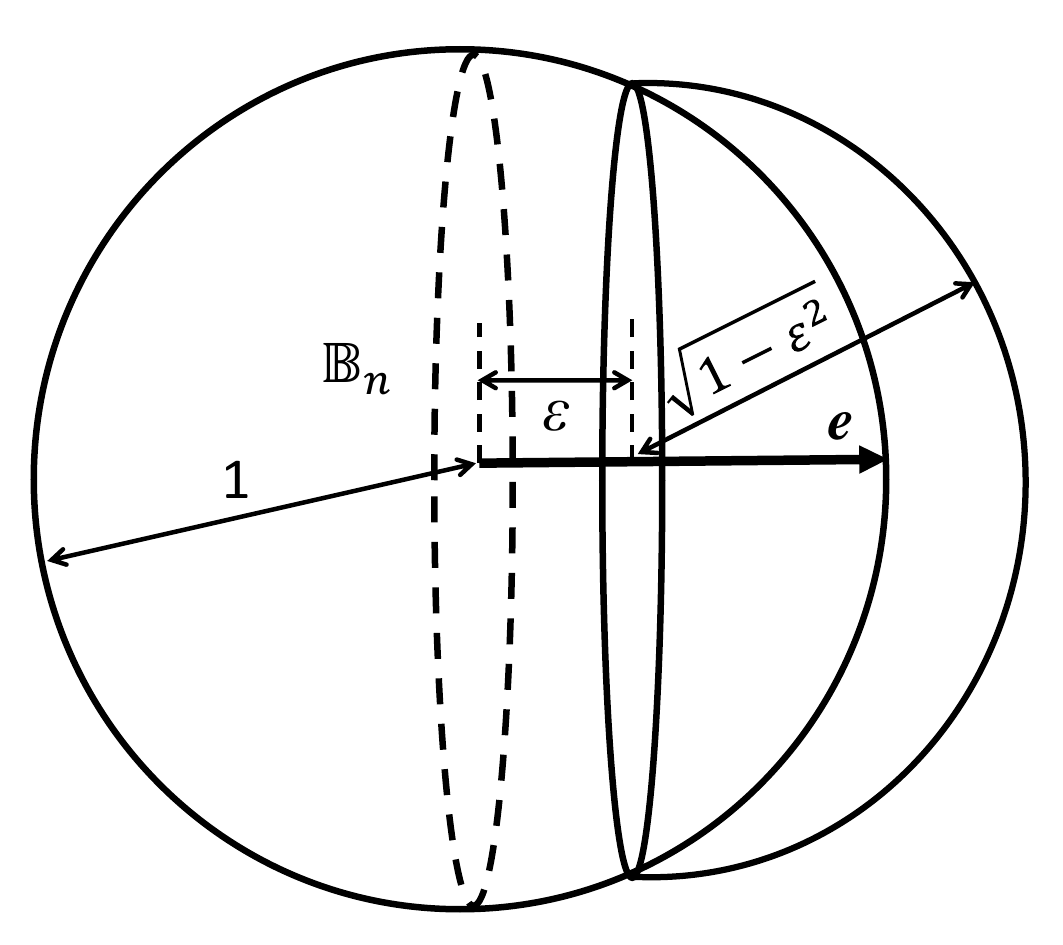}
\caption{\label{SphericalCap} Approximation of a spherical cap by a hemisphere. A spherical cap is portion of  $\mathbb{B}_n$ cut off by a plane on distance $\varepsilon$ from the centre. It is approximated from above by a hemisphere of radius $\sqrt{1-\varepsilon^2}$. The vector $\boldsymbol{x}$ should belong to this spherical cap to ensure the inequality $(\boldsymbol{x},\boldsymbol{e})\geq \varepsilon$.} 
\end{figure} 

\begin{Lemma}\label{Lem:QuasiOrthMany}
Let $\boldsymbol{e}_1, \ldots , \boldsymbol{e}_N \in \mathbb{R}^n$ be   normalised vectors,$\|  \boldsymbol{e}_i\|=1$.  Assume that $\boldsymbol{x}\in \mathbb{B}_n$ is a random point from a distribution in $\mathbb{B}_n$ with the bounded probability density   $\rho(\boldsymbol{x})\leq \rho_{\max}$.  Then, for any $\varepsilon>0$ the probability
\begin{equation}\label{NonOrthMany}
{\bf P}((\boldsymbol{x},\boldsymbol{e}_i)\leq \varepsilon \mbox{ for all } i=1, \ldots, N)\geq 1- \frac{1}{2}N\rho_{\max}V_n(\mathbb{B}_n)(\sqrt{1-\varepsilon^2})^n
\end{equation}
\end{Lemma}
\begin{proof}
Notice that
$${\bf P}((\boldsymbol{x},\boldsymbol{e}_i)\leq \varepsilon \mbox{ for all } i=1, \ldots, N)\geq 1-\sum_i {\bf P}((\boldsymbol{x},\boldsymbol{e}_i)\geq \varepsilon).$$

According to Lemma~\ref{Lem:QuasiOrth}, each term in the last sum is estimated from above by the expression $\frac{1}{2} \rho_{\max}V_n(\mathbb{B}_n)(\sqrt{1-\varepsilon^2})^n$   (\ref{ProbNonOrth}). 
\end{proof}

It is worth mentioning that the term $(\sqrt{1-\varepsilon^2})^n$ decays exponentially when $n$ increases. 

Let $\mathbf{S}_{\boldsymbol{z}, E,A}$ be an ellipsoid  (\ref{ellipsoid}). Decompose a vector  $\boldsymbol{x}\in \mathbb{R}^n$ in an orthonormal basis $E= \{\boldsymbol{e}_1, \ldots , \boldsymbol{e}_n\}$:  $\boldsymbol{x}=\sum_i ( \boldsymbol{x},  \boldsymbol{e}_i)\boldsymbol{e}_i=\|x\| \sum_i \boldsymbol{e}_i \cos \alpha_i$, where
 $ \cos \alpha_i=(\boldsymbol{x},\boldsymbol{e}_i)/\|\boldsymbol{x}\|$. Notice that $\sum_i  \cos^2 \alpha_i=1$ (the $n$-dimensional Pythagoras theorem). 

\begin{Lemma}\label{EllMaximization}
For a given  $\boldsymbol{x}\in \mathbb{R}^n$.  Maximisation of a linear functional  $(\boldsymbol{x},\boldsymbol{z}')$ on an ellipsoid (\ref{ellipsoid}) gives
\begin{equation}\label{EllMax}
\max_{\boldsymbol{z}'\in \mathbf{S}_{\boldsymbol{z}, E,A}} (\boldsymbol{x},\boldsymbol{z}')=(\boldsymbol{x},\boldsymbol{z})+\|\boldsymbol{x}\|\sqrt{\sum_i a_i^2 cos \alpha_i^2},
\end{equation}
 and the maximiser has the following coordinates in the principal axes:
\begin{equation}\label{EllArgmax} 
 z'_i= z_i +  \frac{a_i^2 \cos \alpha_i}{\sqrt{\sum_i a_i^2 \cos \alpha_i^2}},
 \end{equation} 
where  $z'_i=(\boldsymbol{z}', \boldsymbol{e}_i)$, and $z_i= (\boldsymbol{z}, \boldsymbol{e}_i)$ are coordinates of the vectors $\boldsymbol{z}'$, $\boldsymbol{z}$ in the basis $E$.
 \end{Lemma}
 
 \begin{proof}
 Introduce coordinates in the ellipsoid  $\mathbf{S}_{\boldsymbol{z}, E,A}$  (\ref{ellipsoid}): $\Delta_i=z'_i-z_i$. In these coordinates, the objective function is
 $$( \boldsymbol{x},\boldsymbol{z}')=( \boldsymbol{x},\boldsymbol{z})+\|\boldsymbol{x}\| \sum_i \Delta_i \cos \alpha_i.$$
 
For given $ \boldsymbol{x}$,  $\boldsymbol{z}$ we have to maximise $\sum_i \Delta_i \cos \alpha_i$ under the equality constraints:
 $$F(\Delta_1, \ldots, \Delta_n)= \frac{1}{2}\sum_i \frac{\Delta_i^2}{a_i^2}=\frac{1}{2},$$
 because the maximiser of a linear functional on a convex compact set belongs to the border of this compact.
 
 The method of Lagrange multipliers gives:
 $$\cos \alpha_i=\lambda \frac{\partial F}{\partial \Delta_i}=\lambda \frac{\Delta_i}{a_i^2}, \;\;
\Delta_i=\frac{1}{\lambda}a_i^2 \cos \alpha_i.$$ 

To find the Lagrange multiplier $\lambda$, we use the equality constrain again and get
$$\frac{1}{\lambda^2}\sum_i a_i^2 \cos^2 \alpha_i=1, \;\;
 \lambda=\pm  {\sqrt{\sum_i a_i^2 \cos^2 \alpha_i}},$$
where the `+'  sign corresponds to the maximum and the `--' sign corresponds  to the minimum of the objective function. Therefore, the required maximiser has the form (\ref{EllArgmax}) and the corresponding maximal value is given by (\ref{EllMax}).
 \end{proof}
  
 \begin{proof}[Proof of Theorem~\ref{Th:EllGranules} ]
The proof is organised as follows.  Select sufficiently small $R>0$ and find such $k$ that $d_{k+1}<R$. For each elliptic granule select the first $k$ vectors of its principal axes. There will be $N$ vectors of the first axes, $N$ vectors of the second axes, etc. Denote these families of vectors $\mathcal{E}_1$, $\mathcal{E}_2$, ..., $\mathcal{E}_k$: $\mathcal{E}_i$ is a set of vectors of the $i$th principal axis for granules. Let $\mathcal{E}_0$ be the set of the centres of granules. Select a small $\vartheta>0$. Use Lemma~\ref{Lem:QuasiOrthMany} and find the probability that for all $\boldsymbol{e}\in \mathcal{E}_i$ and for all $i=1, \ldots, k$ the following quasiorthogonality condition holds: 
$|(\boldsymbol{x}, \boldsymbol{e})| \leq   \frac{\vartheta}{\sqrt{k}  d_i  }$. Under this condition, evaluate the value of the separation functionals (\ref{EllMax}) in all granules as
\begin{equation}\label{estimateR2}
 (\boldsymbol{x},\boldsymbol{z}')\leq (\boldsymbol{x},\boldsymbol{z})+\|\boldsymbol{x}\|\sqrt{\sum_i a_i^2 cos \alpha_i^2}\leq  (\boldsymbol{x},\boldsymbol{z}) + \sqrt{\vartheta^2 +R^2},
 \end{equation}
where $\boldsymbol{z}$ is the centre of the granule. 
Indeed, 
$$ \|\boldsymbol{x}\|^2 \sum_i a_i^2 cos \alpha_i^2\leq  \sum_{i=1}^k d_i^2 (\boldsymbol{x},\boldsymbol{e}_i)^2 +\sum_{i=k+1}^n \|\boldsymbol{x}\|^2  R^2  cos \alpha_i^2.$$

The quasiorthogonality condition gives that  the first sum does not exceed $\vartheta$.
Recall that $\|\boldsymbol{x}\|\leq 1$ and $\sum_i cos \alpha_i^2=1$. Therefore, the second sum does not exceed $R^2$. This gives us the required estimate (\ref{estimateR2}).

The first term, $ (\boldsymbol{x},\boldsymbol{z})$ is also small with high probability. This quasiorthogonality of $\boldsymbol{x}$ and $N$ vectors of the centres of granules follows from Lemma~\ref{Lem:QuasiOrthMany}. It should be noted that the requirement of  qusiorthogonality of $\boldsymbol{x}$ to several  families of vectors ($N$ centres and $kN$ principal axes) increases the pre-exponential factor in the negative term in  (\ref{NonOrthMany}). This increase can be compensated by a slight increase in the dimensionality  because of the exponential factor there. 

Let us construct the explicit estimates for given $\varepsilon>0$, $\varsigma>0$. Take 
\begin{equation}\label{vartheta} 
 \vartheta=R=\varepsilon/(1+\sqrt{2}).
\end{equation}

 Under conditions of Theorem~\ref{Th:EllGranules} several explicit exponential estimates of probabilities hold:
\begin{enumerate}
\item Volume of a ball with radius $1-\varsigma$ is $V_n(\mathbb{B}_n) (1-\varsigma)^n$. therefore for probability of $\boldsymbol{x}$ belong to this ball, we have
$${\bf P} (  (\boldsymbol{x},\boldsymbol{x})\leq 1-\varsigma)\leq\rho_{\max}V_n(\mathbb{B}_n) (1-\varsigma)^n;$$ 
\item For every $\boldsymbol{z} \in \mathcal{E}_0$,
$${\bf P} (  (\boldsymbol{x},\boldsymbol{z})\geq \vartheta )\leq \rho_{\max}\frac{1}{2}V_n(\mathbb{B}_n)(\sqrt{1-\vartheta^2})^n;$$
\item For every  $\boldsymbol{e}\in \mathcal{E}_i$
$${\bf P} \left(  |(\boldsymbol{x},\boldsymbol{e})|\geq  \frac{\vartheta}{\sqrt{k}  d_i  }\right)\leq \rho_{\max}V_n(\mathbb{B}_n)\left(\sqrt{1-\left(\frac{\vartheta}{\sqrt{k}  d_i  }\right)^2}\right)^n.$$
\end{enumerate}


Thus, the probability 
 \begin{equation}\label{Th3estimate}
\begin{split}
&{\bf P} \left(  (\boldsymbol{x},\boldsymbol{x})\geq 1-\varsigma \; \& \; (\boldsymbol{x},\boldsymbol{z})\leq \vartheta \mbox{  for all  }\boldsymbol{z}\in \mathcal{E}_0 \; \&  \; |(\boldsymbol{x},\boldsymbol{e})|\leq  \frac{\vartheta}{\sqrt{k}  d_i  } \mbox{  for all  } \boldsymbol{e}\in \mathcal{E}_i, \,  i=1,\ldots, k \right) \\
&\geq 1-\rho_{\max}V_n(\mathbb{B}_n) \left[(1-\varsigma)^n +\frac{1}{2}N(\sqrt{1-\vartheta^2})^n+N\sum_{i=1}^k \left(\sqrt{1-\left(\frac{\vartheta}{\sqrt{k}  d_i  }\right)^2}\right)^n\right].
\end{split}
\end{equation}

If $(\boldsymbol{x},\boldsymbol{z})\leq \vartheta$  for all  $\boldsymbol{z}\in \mathcal{E}_0$  and $| (\boldsymbol{x},\boldsymbol{e})|\leq  \frac{\vartheta}{\sqrt{k}  d_i  }$  for all  $\boldsymbol{e}\in \mathcal{E}_i, \,  i=1,\ldots, k $ then, according to the choice of $\vartheta$ (\ref{vartheta}) and inequality (\ref{estimateR2}),  $  (\boldsymbol{x},\boldsymbol{z}')\leq  \varepsilon$ for all points from the granules $\boldsymbol{z}'\in D$.

Therefore, (\ref{Th3estimate}) proves Theorem~\ref{Th:EllGranules} with explicit estimate of the probability.

 If, in addition, $ (\boldsymbol{x},\boldsymbol{x})\geq 1-\varsigma$, $0<\alpha\leq 1$ and $\alpha(1-\varsigma)>\varepsilon$ then $$\alpha(\boldsymbol{x},\boldsymbol{x})> (\boldsymbol{x},\boldsymbol{z}') \mbox{  for all  } \boldsymbol{z}'\in D$$  for all points from the granules $\boldsymbol{z}'\in D$. This is the analogue of $\alpha$-Fisher separability of   point $\boldsymbol{x}$ from elliptic granules.  \end{proof}
 
Theorem~\ref{Th:EllGranules} describes stochastic separation of a random point in $n$-dimensional dataspace  from a set of $N$ elliptic granules. For given $N$  probability of $\alpha$-Fisher separability exponentially approaches 1 with dimensionality growth. Equivalently, for a given probability, the upper bound on the number of granules that guarantees such a separation with this probability grows exponentially with the dimension of the data. We require two properties of the probability distribution: compact support and the existence of a probability density bounded from above. The interplay between the dependence of the maximal density on the dimension (similarly to (\ref{bounded})) and the exponents in the probability estimates   (\ref{Th3estimate}) determines the estimate of the separation probability. 

In Theorem~\ref{Th:EllGranules} we analysed separation of a random point from a set of granules but it seems to be much more practical to consider separation of a random granule from a set of granules. For analysis of random granules a joint distribution of the position of the centre and the basis  of principal axes  is needed.  Existence of strong dependencies between the position of the centre and the directions of principal axes may in special cases  destroy the separability phenomenon. For example, if the first principal axis has length 1 or more and is parallel to the vector of the centre (i.e., $ \boldsymbol{e}_1 =\boldsymbol{x}/\|\boldsymbol{x}\|$) then this granule is not separated even from the origin. On the other hand, independence of these distributions guarantees stochastic separability, as  follows from Theorem~\ref{RandomGranule} below. Independence by itself is not needed. The essential condition is that for each orientation of the granule, the position of its centre remains rather uncertain.

\begin{Theorem}\label{RandomGranule}
Consider a set of $N$  elliptic granules (\ref{ellipsoid}) with centres  $\boldsymbol{z} \in \mathbb{B}_n$ and $a_i\leq d_i$.  Let $D$ be the union of all these granules.  Assume that $\boldsymbol{x}\in \mathbb{B}_n$ is a random point from a distribution in the unit ball with the bounded probability density $\rho(\boldsymbol{x})\leq \rho_{\max}$. Let  $\boldsymbol{x}$ be a centre of a random elliptic granule  $\mathbf{S}_x=\mathbf{S}_{\boldsymbol{x}, E_x,A_x}$ (\ref{ellipsoid}). Assume that for any  basis of principal axes $E$ and sequence of semi-axes $A=\{a_i\}$ ($a_i\leq d_i$) the conditional distribution 
of the centres of granules $\boldsymbol{x}$  given $E_x=E,\, A_x=A$
has a density in  $\mathbb{B}_n$ uniformly bounded from above:  
$$\rho(\boldsymbol{x} \ | \  E_x=E,\, A_x=A) \leq \rho_{\max}$$ 
and $\rho_{\max}$ does not depend on $ E_x,A_x$
Then for positive $ \varepsilon $, $\varsigma$
\begin{equation}\label{EllClusterBound2}
\begin{split}
{\bf P}& ((\boldsymbol{x} ,\boldsymbol{z}') \leq \varepsilon  \mbox{  for all  }  \boldsymbol{z}'\in D  \; \;\&\;\; (\boldsymbol{x},\boldsymbol{x}')\geq (\boldsymbol{x},\boldsymbol{x})-\varepsilon \mbox{  for all } \boldsymbol{x}'  \in  \mathbf{S}_x \;\; \& \;\;(x,x)\geq 1- \varsigma \\
&\geq  1 - N \rho_{\max}V_n(\mathbb{B}_n) a \exp(-bn),
\end{split}
\end{equation}
where $a$ and $b$ do not depend on the dimensionality.
\end{Theorem}
 
 In the proof of Theorem~\ref{RandomGranule} we estimate the probability (\ref{EllClusterBound2}) by a sum of decaying exponentials, which give explicit formulas for $a$ and $b$ as  was done for  Theorem~\ref{Th:EllGranules} in (\ref{Th3estimate}).
 \begin{proof}
We will prove (\ref{EllClusterBound2}) for an  elipsoid $ \mathbf{S}_x$ (\ref{ellipsoid}) with given (not random) basis $E$ and semiaxes $a_i\leq d_i$,  and with a random centre $ \boldsymbol{x}\in \mathbb{B}_n$ assuming that the distribution density of $ \boldsymbol{x}$ is  bounded from above by $\rho_{\max}$. 

Select sufficiently small $R>0$ and find such $k$ that $d_{k+1}<R$. For each granule, including $\mathbf{S}_x$ with the centre $\boldsymbol{x}$ select the first $k$ vectors of its principal axes. There will be $N+1$ vectors of the first axes, $N+1$ vectors of the second axes, etc. Denote these families of vectors $\mathcal{E}_1$, $\mathcal{E}_2$, ..., $\mathcal{E}_k$: $\mathcal{E}_i$ is a set of vectors of the $i$th principal axis for all granules, $\mathbf{S}_x$.  Let   $\mathcal{E}_0$ be the set of  of the centres of granules (excluding the centre $ \boldsymbol{x}$ of the granule $\mathbf{S}_x$. )

For a given $\vartheta>0$ the following estimate of probability holds (analogously to  (\ref{Th3estimate})).

\begin{equation}\label{Th4estimate}
\begin{split}
&{\bf P} \left(  (\boldsymbol{x},\boldsymbol{x})\geq 1-\vartheta \; \& \; (\boldsymbol{x},\boldsymbol{z})\leq \vartheta \mbox{  for all  }\boldsymbol{z}\in \mathcal{E}_0 \; \&  \; |(\boldsymbol{x},\boldsymbol{e})|\leq  \frac{\vartheta}{\sqrt{k}  d_i  } \mbox{  for all  } \boldsymbol{e}\in \mathcal{E}_i, \,  i=1,\ldots, k \right) \\
&\geq 1-\rho_{\max}V_n(\mathbb{B}_n) \left[(1-\vartheta)^n +\frac{1}{2}N(\sqrt{1-\vartheta^2})^n+(N+1)\sum_{i=1}^k \left(\sqrt{1-\left(\frac{\vartheta}{\sqrt{k}  d_i  }\right)^2}\right)^n\right].
\end{split}
\end{equation}  


If  $(\boldsymbol{x},\boldsymbol{x})\geq 1-\vartheta$ and $(\boldsymbol{x},\boldsymbol{z})\leq \vartheta$    for all   $\boldsymbol{z}\in \mathcal{E}_0$, and  $|(\boldsymbol{x},\boldsymbol{e})|\leq  \frac{\vartheta}{\sqrt{k}  d_i  }$  for all   $\boldsymbol{e}\in \mathcal{E}_i, \,  i=1,\ldots, k   $, then by (\ref{estimateR2})
$$(\boldsymbol{x},\boldsymbol{z}')\leq \vartheta+ \sqrt{\vartheta^2+R^2}\;\; \& \;\; (\boldsymbol{x},\boldsymbol{x}')\geq 1-\vartheta -\sqrt{\vartheta^2+R^2} \mbox{  for all }  \boldsymbol{z}'\in D, \boldsymbol{x}'\in \mathbf{S}_x .$$

Therefore, if we select  $R=\frac{\varepsilon}{1+ \sqrt{2}}$ and $\vartheta=\min\left\{\varsigma,\frac{\varepsilon}{1+ \sqrt{2}}\right\}	$, then the estimate (\ref{Th4estimate}) proves Theorem~\ref{RandomGranule}. Additionally, for this choice, $(\boldsymbol{x},\boldsymbol{x}')\geq 1-\varepsilon$ for all $\boldsymbol{x}'\in \mathbf{S}_x $. Therefore, if $\varepsilon<\frac{\alpha}{1+\alpha}$, then $\alpha (\boldsymbol{x},\boldsymbol{x}')> (\boldsymbol{x},\boldsymbol{z}')$ for all  $\boldsymbol{z}'\in D$ and $\boldsymbol{x}'\in \mathbf{S}_x$ with probability estimated in (\ref{Th4estimate}). This result can be considered as $\alpha$-Fisher separability of elliptic granules  in high dimensions with high probability. 
 \end{proof}
 
 Note that the the proof does not actually use that $d_i \to 0$. All that we use that $\limsup\limits_{i\to\infty}d_i < R$ for $R=\frac{\epsilon}{1+\sqrt{2}}$, where $\epsilon<\frac{\alpha}{1+\alpha}$. Hence the proof remains valid whenever $\limsup\limits_{i\to\infty}d_i < \frac{\alpha}{(1+\sqrt{2})(1+\alpha)}$.
  
It may be useful to formulate a version  of Theorem~\ref{RandomGranule} when ${\bf S}_{\boldsymbol x}$ is the granule of an arbitrary (non-random) shape but with a random centre as  a separate Proposition.
 
 \begin{Proposition}\label{prop:th4nonrandom}
Let $D$ be the union of $N$ elliptic granules (\ref{ellipsoid}) with centres in ${\mathbb B}_n$ with $a_i \leq d_i$.  
Let ${\bf S}_{{\boldsymbol z},E,A}$ be one more such granule. Let ${\boldsymbol x} \in {\mathbb B}_n$ be a random point from a distribution in the unit ball with the bounded probability
density $\rho(x) \leq \rho_{\text{max}}$. Let ${\bf S}_{\boldsymbol x} = {\bf S}_{{\boldsymbol z},E,A} + ({\boldsymbol x} - {\boldsymbol z})$ be the granule ${\bf S}_{{\boldsymbol z},E,A}$ shifted such that its centre becomes ${\boldsymbol x}$. Then Theorem~\ref{RandomGranule} is true for ${\bf S}_{\boldsymbol x}$.
\end{Proposition}
 The proof is the same as the proof of Theorem~\ref{RandomGranule}.

The estimates  (\ref{Th3estimate}) and (\ref{Th4estimate}) are far from being sharp. Detailed analysis for various classes of distributions may give better estimates as it was done for separation of finite sets~\cite{Grechuk2021}. This work needs to be done for separation of granules as well.

\subsection{Superstatistic Presentation of 'Granules'}

The alternative approach to the granular structure of the distributions are {\em soft clusters}. They can be studied in the frame of {\em superstatistical} approach with representation of data distribution by a random mixture of distributions of points in individual clusters. We start with the following remark.
Notice that Proposition \ref{prop:th4nonrandom} has the following easy corollary. 

\begin{Corollary}\label{cor:allrandom}
Let ${\bf S}_{\boldsymbol x}$ and $D$ be as in Proposition \ref{prop:th4nonrandom}. Let ${\boldsymbol x'}$ and ${\boldsymbol z'}$ be the points selected uniformly at random from ${\bf S}_{\boldsymbol x}$ and $D$, correspondingly. Then for positive $\epsilon, \zeta$
$$
{\bf P}(({\boldsymbol x}, {\boldsymbol z'}) \leq \epsilon \,\,\&\,\, ({\boldsymbol x}, {\boldsymbol x'}) \geq ({\boldsymbol x},{\boldsymbol x})-\epsilon \,\,\&\,\, ({\boldsymbol x}, {\boldsymbol x}) \geq 1 - \zeta) \geq  1 - N\rho_{\text{max}} V_n({\mathbb B}_n)a \exp(-bn),
$$
where the constants $a,b$ are the same as in Theorem~\ref{RandomGranule}.
\end{Corollary}
\begin{proof}
Let $f(n)= N\rho_{\text{max}} V_n({\mathbb B}_n)a \exp(-bn)$. 
Let $A \subset {\mathbb B}_n$ be the set of ${\boldsymbol x}$ such that (\ref{EllClusterBound2}) holds. Proposition \ref{prop:th4nonrandom} states that ${\bf P}({\boldsymbol x} \in A) \geq 1-f(n)$. Let $E$ be the event that $({\boldsymbol x}, {\boldsymbol z'}) \leq \epsilon \,\,\&\,\, ({\boldsymbol x}, {\boldsymbol x'}) \geq ({\boldsymbol x},{\boldsymbol x})-\epsilon \,\,\&\,\, ({\boldsymbol x}, {\boldsymbol x}) \geq 1 - \zeta$. By the law of total probability, 
\begin{equation*}
\begin{split}
{\bf P}(E)& = {\bf P}(E|{\boldsymbol x} \in A){\bf P}({\boldsymbol x} \in A) + {\bf P}(E|{\boldsymbol x} \not\in A){\bf P}({\boldsymbol x} \not\in A) \\
      & \geq {\bf P}(E|{\boldsymbol x} \in A){\bf P}({\boldsymbol x} \in A) = 1 \cdot {\bf P}({\boldsymbol x} \in A) \geq 1-f(n). 
\end{split}
\end{equation*} \end{proof}

Corollary \ref{cor:allrandom} is weaker than Proposition \ref{prop:th4nonrandom}. While Proposition \ref{prop:th4nonrandom} states that, with probability at least $1 - f(n)$, \emph{the whole} granule ${\bf S}_{\boldsymbol x}$ can be separated from \emph{all} points in $D$, \mbox{Corollary \ref{cor:allrandom}} allows for the possibility that there could be a small portions of ${\bf S}_{\boldsymbol x}$ and $D$ which are not separated from each other. As we will see below, this weakening allows us to prove the result in much greater generality, where the uniform distribution in granules is replaced by much more general log-concave distributions.

We say that density $\rho:{\mathbb R}^n \to [0,\infty)$ of random vector $\boldsymbol{x}$ (and the corresponding probability distribution) is \emph{log-concave}, if set $K=\{z\in{\mathbb R}^n \,|\, \rho(z)>0\}$ is convex and $g(z)=-\log(\rho(z))$ is a convex function on $K$. For example, the uniform distribution in any full-dimensional subset of ${\mathbb R}^n$ (and in particular uniform distribution in granules (\ref{ellipsoid})) has a log-concave density.

We say that $\rho$ is whitened, or \emph{isotropic}, if ${\mathbb E}[\boldsymbol{x}]=\boldsymbol{ 0}$, and
\begin{equation}\label{eq:isot}
{\mathbb E}[(\boldsymbol{x},\theta)^2)]=1\quad\quad \forall \theta \in \mathbb{S}^{n-1},
\end{equation}
where $\mathbb{S}^{n-1}$ is the unit sphere in ${\mathbb R}^n$. Equation \eqref{eq:isot} is equivalent to the statement that the variance-covariance matrix for the components of $\boldsymbol{x}$ is the identity matrix. This can be achieved by linear transformation, hence every log-concave random vector $\boldsymbol{x}$ can be represented as 
\begin{equation}\label{eq:Adef}
\boldsymbol{x}=\Sigma\boldsymbol{y}+\boldsymbol{x}_0,
\end{equation}
where $\boldsymbol{x}_0 = {\mathbb E}[\boldsymbol{x}]$, $\Sigma$ is (non-random) matrix and $\boldsymbol{y}$ is some isotropic log-concave random vector.

An example of standard normal distribution shows that the support of isotropic log-concave distribution may be the whole ${\mathbb R}^n$. However, such distributions are known to be concentrated in a ball of radius $\sqrt{n}(1+\delta)$ with high probability. 

Specifically, (\cite{Guedon}, Theorem 1.1) implies that for any $\delta\in(0,1)$  and any isotropic log-concave random vector in ${\mathbb R}^n$,
\begin{equation}\label{eq:tailbound1}
{\bf P}(\|\boldsymbol{x}\| \leq (1+\delta)\sqrt{n}) \geq 1-c\exp(-c'\delta^3 \sqrt{n})
\end{equation}
where $c,c'>0$ are some absolute constants. Note that we have $\sqrt{n}$ but not $n$ in the exponent, and this cannot be improved without requiring extra conditions on the distribution. We say that density $\rho:{\mathbb R}^n \to [0,\infty)$ is strongly log-concave with constant $\gamma>0$, or $\gamma$-SLC in short, if $g(z)=-\log(\rho(z))$ is strongly convex, that is, $g(z)-\frac{\gamma}{2}\|z\|$ is a convex function on $K$. (\cite{Guedon}, Theorem 1.1) also implies that 
\begin{equation}\label{eq:tailbound2}
{\bf P}(\|\boldsymbol{x}\| \leq (1+\delta)\sqrt{n}) \geq 1-c\exp(-c'\delta^4 n)
\end{equation}
for any $\delta\in(0,1)$, and any isotropic strongly log-concave random vector $\boldsymbol{x}$ in ${\mathbb R}^n$.

Fix some $\delta>0$ and infinite sequence $d=(d_1>d_2>\dots )$ with each $d_i>0$ and $d_i \to 0$. Let us call log-concave random vector $\boldsymbol{x}$ $(\delta,d)$-admissible if set $\Sigma \cdot B({\bf 0},(1+\delta)\sqrt{n})+\boldsymbol{x}_0$ is a subset of some ellipsoid ${\bf S}_{\boldsymbol{x}_0,E,A}$ (\ref{ellipsoid}), where $\Sigma$ and $\boldsymbol{x}_0$ are  defined in \eqref{eq:Adef} and $B({\bf 0},(1+\delta)\sqrt{n})$ is the ball with centre ${\bf 0}$ and radius $(1+\delta)\sqrt{n}$. Then \eqref{eq:tailbound1} and \eqref{eq:tailbound2} imply that $\boldsymbol{x} \in {\bf S}_{\boldsymbol{x}_0,E,A}$ with high probability. In combination with Proposition \ref{prop:th4nonrandom}, this implies the following results.

\begin{Proposition}\label{prop:logconc}
Let $\delta>0$ and infinite sequence $d=(d_1>d_2>\dots )$ with each $d_i>0$ and $d_i \to 0$ be fixed. Let ${\boldsymbol x} \in {\mathbb B}_n$ be a random point from a distribution in the unit ball with the bounded probability density $\rho(x) \leq \rho_{\text{max}}$. Let ${\boldsymbol x''}$ be a point selected from some $(\delta,d)$-admissible log-concave distribution, and let ${\boldsymbol x'}={\boldsymbol x''}-{\mathbb E}[{\boldsymbol x''}]+{\boldsymbol x}$. Let ${\boldsymbol z'}$ be the point  selected from a mixture of $N$ $(\delta,d)$-admissible log-concave distributions with centres in ${\mathbb B}_n$. Then for positive $\epsilon, \zeta$

\begin{equation*}
\begin{split}
{\bf P}(({\boldsymbol x}, {\boldsymbol z'}) & \leq \epsilon \,\,\&\,\, ({\boldsymbol x}, {\boldsymbol x'}) \geq ({\boldsymbol x},{\boldsymbol x})-\epsilon \,\,\&\,\, ({\boldsymbol x}, {\boldsymbol x}) \geq 1 - \zeta) \\ & \geq  1 - N\rho_{\text{max}} V_n({\mathbb B}_n)a \exp(-bn) - 2 c\exp(-c'\delta^3 \sqrt{n}),
\end{split}
\end{equation*}
for some constants $a,b,c,c'$ that do not depend on the dimensionality. 
\end{Proposition}
\begin{proof}
If follows from \eqref{eq:tailbound1} and $(\delta,d)$-admissibility of the distribution from which ${\boldsymbol x''}$ has been selected that  
$$
{\bf P}({\boldsymbol x'} \not\in {\bf S}_0) \leq c\exp(-c'\delta^3 \sqrt{n})
$$
for some ellipsoid ${\bf S}_0$ (\ref{ellipsoid}). Similarly, since ${\boldsymbol z'}$ is selected from a mixture of $N$ $(\delta,d)$-admissible log-concave distributions, we have 
$$
{\bf P}\left({\boldsymbol z'} \not\in \bigcup_{i=1}^N{\bf S}_i\right) \leq c\exp(-c'\delta^3 \sqrt{n})
$$
for some ellipsoids ${\bf S}_1, \dots, {\bf S}_N$ (\ref{ellipsoid}). 
Let $E$ be the event that $({\boldsymbol x}, {\boldsymbol z'}) \leq \epsilon \,\,\&\,\, ({\boldsymbol x}, {\boldsymbol x'}) \geq ({\boldsymbol x},{\boldsymbol x})-\epsilon \,\,\&\,\, ({\boldsymbol x}, {\boldsymbol x}) \geq 1 - \zeta$. If $E$ does not happen than either (i) ${\boldsymbol x'} \not\in {\bf S}_0$, or (ii) ${\boldsymbol z'} \not\in \bigcup_{i=1}^N{\bf S}_i$, or (iii) ${\boldsymbol x'} \in {\bf S}_0$ and ${\boldsymbol z'} \in \bigcup_{i=1}^N{\bf S}_i$, but $E$ still does not happen. The probabilities of (i) and (ii) are at most $c\exp(-c'\delta^3 \sqrt{n})$, while the probability of (iii) is at most $N\rho_{\text{max}} V_n({\mathbb B}_n)a \exp(-bn)$ by Proposition \ref{prop:th4nonrandom}. 
\end{proof}

Exactly the same proof in combination with \eqref{eq:tailbound2} implies the following version for strongly log-concave distributions.

\begin{Proposition}\label{prop:stronglog}
Let $\delta,\gamma>0$ and infinite sequence $d=(d_1>d_2>\dots )$ with each $d_i>0$ and $d_i \to 0$ be fixed. Let ${\boldsymbol x} \in {\mathbb B}_n$ be a random point from a distribution in the unit ball with the bounded probability density $\rho(x) \leq \rho_{\text{max}}$. Let ${\boldsymbol x''}$ be a point selected from some $(\delta,d)$-admissible $\gamma$-SLC distribution, and let ${\boldsymbol x'}={\boldsymbol x''}-{\mathbb E}[{\boldsymbol x''}]+{\boldsymbol x}$. Let ${\boldsymbol z'}$ be the point  selected from a mixture of $N$ $(\delta,d)$-admissible $\gamma$-SLC distributions with centres in ${\mathbb B}_n$. Then for positive $\epsilon, \zeta$
$$
{\bf P}(({\boldsymbol x}, {\boldsymbol z'}) \leq \epsilon \,\,\&\,\, ({\boldsymbol x}, {\boldsymbol x'}) \geq ({\boldsymbol x},{\boldsymbol x})-\epsilon \,\,\&\,\, ({\boldsymbol x}, {\boldsymbol x}) \geq 1 - \zeta) \geq  1 - N\rho_{\text{max}} V_n({\mathbb B}_n)a \exp(-bn) - 2 c\exp(-c'\delta^4 n),
$$  
for some constants $a,b,c,c'$ that do not depend on the dimensionality. 
\end{Proposition}

\subsection{The Superstatistic form of the Prototype Stochastic Separation Theorem}

Theorem~\ref{Th:prototype} evaluates the probability that a random point ${\boldsymbol x} \in {\mathbb B}_n$ with bounded probability  density is $\alpha$-Fisher separable from an exponentially large finite set $Y$ and demonstrates that under some natural conditions this probability tends to zero when dimension $n$ tends to $\infty$. This phenomenon has a simple explanation: for any ${\boldsymbol y}\in {\mathbb B}_n$  the set of such ${\boldsymbol x}\in{\mathbb B}_n$ that ${\boldsymbol x}$ is not $\alpha$-Fisher separable from  ${\boldsymbol y}$ is a ball with radius $\| {\boldsymbol y}\|/(2\alpha)<1$ and the fraction of this volume in ${\mathbb B}_n$ decays as 
$$\left(\frac{\| {\boldsymbol y}\|}{2\alpha}\right)^n.$$

These arguments can be generalised with some efforts for the situation when we consider an elliptic granule instead of a random point ${\boldsymbol x}$ and an {\em arbitrary } probability distribution instead of a finite set $Y$. Instead of the estimate of the probability of a point ${\boldsymbol x}$ falling into a the ball of excluded volume (\ref{excludedvolume}), we use the following proposition for separability of a random point ${\boldsymbol x'}$ of a  granule ${\bf S}_{\boldsymbol x}$ with a random centre ${\boldsymbol x }$ from an arbitrary point ${\boldsymbol z'} \in {\mathbb B}_n$.

\begin{Proposition}\label{prob:anyz}
Let ${\bf S}_{\boldsymbol x}$ be the granule defined in Proposition \ref{prop:th4nonrandom}. Let ${\boldsymbol x'}$ be the point selected uniformly at random from ${\bf S}_{\boldsymbol x}$. Let ${\boldsymbol z'} \in {\mathbb B}_n$ be an arbitrary (non-random) point. Then for positive $\epsilon, \zeta$
$$
{\bf P}(({\boldsymbol x}, {\boldsymbol z'}) \leq \epsilon \,\,\&\,\, ({\boldsymbol x}, {\boldsymbol x'}) \geq ({\boldsymbol x},{\boldsymbol x})-\epsilon \,\,\&\,\, ({\boldsymbol x}, {\boldsymbol x}) \geq 1 - \zeta) \geq  1 - \rho_{\text{max}} V_n({\mathbb B}_n)a \exp(-bn),
$$
where the constants $a,b$ do not depend on the dimensionality.
\end{Proposition} 
\begin{proof}
The fact that 
$$
{\bf P}(({\boldsymbol x}, {\boldsymbol x'}) \geq ({\boldsymbol x},{\boldsymbol x})-\epsilon \,\,\&\,\, ({\boldsymbol x}, {\boldsymbol x}) \geq 1 - \zeta) \geq  1 - \rho_{\text{max}} V_n({\mathbb B}_n)a \exp(-bn)
$$
is proved in Theorem~\ref{RandomGranule}, while the fact that
$$
{\bf P}(({\boldsymbol x}, {\boldsymbol z'}) \leq \epsilon) \geq  1 - \rho_{\text{max}} V_n({\mathbb B}_n)a \exp(-bn)
$$
follows from Lemma~\ref{Lem:QuasiOrth}.
\end{proof}

Propositions \ref{prop:logconc} and \ref{prop:stronglog} can be straightforwardly generalised in the same way

\begin{Proposition}\label{prop:logconc_anyz}
Let $\delta>0$ and infinite sequence $d=(d_1>d_2>\dots )$ with each $d_i>0$ and $d_i \to 0$ be fixed. Let ${\boldsymbol x} \in {\mathbb B}_n$ be a random point from a distribution in the unit ball with the bounded probability density $\rho(x) \leq \rho_{\text{max}}$. Let ${\boldsymbol x''}$ be a point selected from some $(\delta,d)$-admissible log-concave distribution, and let ${\boldsymbol x'}={\boldsymbol x''}-{\mathbb E}[{\boldsymbol x''}]+{\boldsymbol x}$. Let ${\boldsymbol z'} \in {\mathbb B}_n$ be an arbitrary (non-random) point. Then for positive $\epsilon, \zeta$

$$
{\bf P}(({\boldsymbol x}, {\boldsymbol z'}) \leq \epsilon \,\,\&\,\, ({\boldsymbol x}, {\boldsymbol x'}) \geq ({\boldsymbol x},{\boldsymbol x})-\epsilon \,\,\&\,\, ({\boldsymbol x}, {\boldsymbol x}) \geq 1 - \zeta) \geq  1 - \rho_{\text{max}} V_n({\mathbb B}_n)a \exp(-bn) - c\exp(-c'\delta^3 \sqrt{n}),
$$  
for some constants $a,b,c,c'$ that do not depend on the dimensionality. 
\end{Proposition}

\begin{Proposition}\label{prop:stronglog_anyz}
Let $\delta,\gamma>0$ and infinite sequence $d=(d_1>d_2>\dots )$ with each $d_i>0$ and $d_i \to 0$ be fixed. Let ${\boldsymbol x} \in {\mathbb B}_n$ be a random point from a distribution in the unit ball with the bounded probability density $\rho(x) \leq \rho_{\text{max}}$. Let ${\boldsymbol x''}$ be a point selected from some $(\delta,d)$-admissible $\gamma$-SLC distribution, and let ${\boldsymbol x'}={\boldsymbol x''}-{\mathbb E}[{\boldsymbol x''}]+{\boldsymbol x}$. Let ${\boldsymbol z'} \in {\mathbb B}_n$ be an arbitrary (non-random) point. Then for positive $\epsilon, \zeta$
$$
{\bf P}(({\boldsymbol x}, {\boldsymbol z'}) \leq \epsilon \,\,\&\,\, ({\boldsymbol x}, {\boldsymbol x'}) \geq ({\boldsymbol x},{\boldsymbol x})-\epsilon \,\,\&\,\, ({\boldsymbol x}, {\boldsymbol x}) \geq 1 - \zeta) \geq  1 - \rho_{\text{max}} V_n({\mathbb B}_n)a \exp(-bn) - c\exp(-c'\delta^4 n),
$$  
for some constants $a,b,c,c'$ that do not depend on the dimensionality. 
\end{Proposition}

We remark that because Propositions \ref{prob:anyz}--\ref{prop:stronglog_anyz} hold  for an arbitrary (non-random) point ${\boldsymbol z'} \in {\mathbb B}_n$, they also hold for point selected from \emph{any} probability distribution within ${\mathbb B}_n$, and in particular if point ${\boldsymbol z'}$ selected uniformly at random from \emph{any} set $D \subset {\mathbb B}_n$. 

 \subsection{Compact Embedding of Patterns and Hierarchical Universe \label{Sec:Compact}}

Stochastic separation theorems tell us that in large dimensions, randomly selected data points (or clusters of data) can be separated by simple and explicit functionals from an existing dataset with high probability, as long as the dataset is not too large (or the number of data clusters is not too large). The number of data points (or clusters) allowed in conditions of these theorems is bounded from above by an exponential function of dimension. Such theorems for data points (see, for example, Teorem~\ref{Th:prototype} and \cite{Grechuk2021}) or clusters (Theorems~\ref{Th:SphericGranules}--\ref{RandomGranule}) are valid for broad families of probability distributions. Explicit estimations of probability to violate the separability property were found.

There is a circumstance that can devalue this (and many other) probabilistic results in high dimension. We almost never know the probability of a multivariate data distribution beyond strong simplification assumptions. In the postclassical world, observations cannot really help because we never have enough data to restore the probability density (again, strong simplification like independence assumption or dimensionality reduction can help, but this is not a general multidimensional case). A radical point of view is possible, according to which there is no such thing as a general multivariate probability distribution, since it is unobservable. 

In the infinite-dimensional limit the situation can look simpler: instead of finite but small probabilities that  decrease  and tend  to zero with increasing dimension  (like in (\mbox{\ref{Th3estimate}) and (\ref{Th4estimate})}) some  statements  become generic and hold 'almost always'. Such limits for concentrations on spheres and their equators  were discussed by L\'{e}vy \cite{Levy1951} as an important part of the measure concentration effects. In physics, this limit corresponds to the so-called thermodynamic limit of statistical mechanics \cite{Khinchin1949,Thompson2015}. In the infinite-dimensional limit many statements about high or low probabilities transform into 0-1 laws: something happens almost always or almost newer. The original Kolmogorov 0-1 law states, roughly speaking, that an event that depends on an infinite collection of
independent random variables  but is independent of any finite subset of these variables has probability zero or one (for precise formulation we refer to the monograph \cite{Kolmogorov2018}).  The infinite-dimensional 0-1 asymptotic might bring more light and be more transparent than the probabilistic formulas.

From the infinite-dimensional point of view, the `elliptic granule' 
(\ref{ellipsoid}) with decaying sequence of diameters $d_1>d_2>....$ ($d_i>0$, $d_i\to 0$)  is a compact. The specific elliptic shape used in Theorem~\ref{Th:EllGranules} is not very important and many generalisations are possible for the granules with decaying sequence of diameters. The main idea, from this point of view,  is compact embedding of specific patterns into general population of data. This point of view was influenced by the hierarchy of Sobolev Embedding Theorems where the balls of embedded spaces appear to be compact in the image space.

The finite-dimensional hypothesis about granular structure of the datasets can be transformed into the infinite-dimensional view about compact embedding: the  patterns correspond to the compact subsets of the dataspace.   Moreover, this hypothesis can be extended to the hypothesis about hierarchical structure (Figure~\ref{HierarchicalUniverse}): the data that correspond to a pattern also have the intrernal  granular structure. To reveal this structure,  we can apply centralisation and whitening to a granule. After that, the granule will transform into a new unit ball,  the external set (the former `Universe') will typically become infinitely far (`invisible'), and the internal structure can be seeking in the form of collection of compact granules in new topology.}


\begin{figure}[H]
\centering
\includegraphics[width=0.95\textwidth]{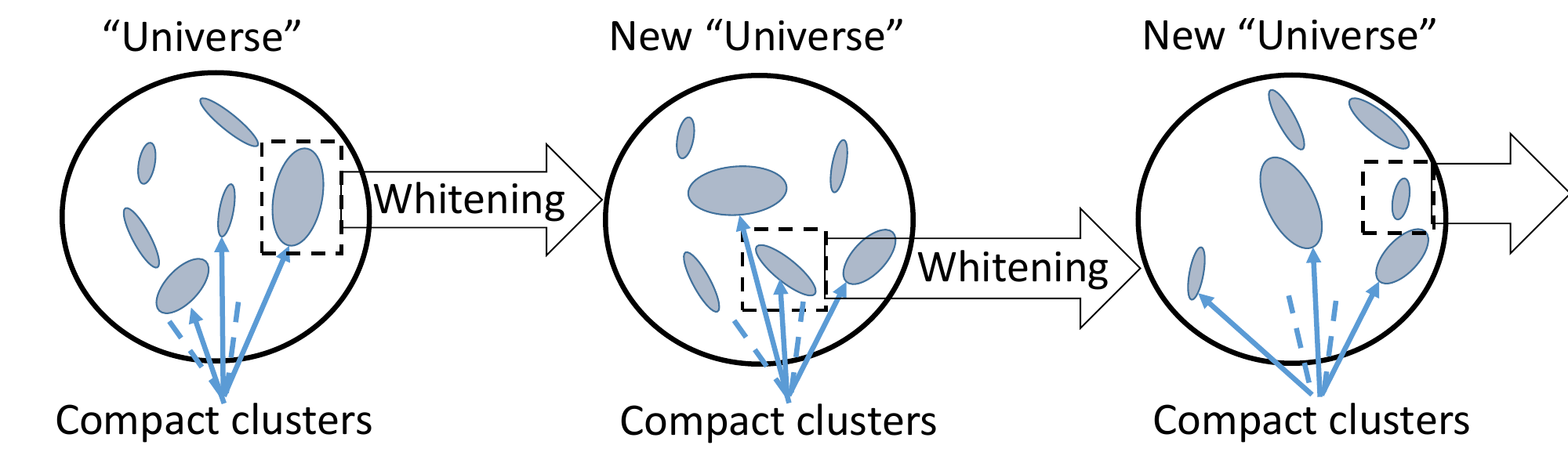}
\caption{\label{HierarchicalUniverse} Hierarchical Universe.
Each pattern is represented by a compact set embedded in the data universe. When we select this compact and apply whitening, it becomes a new universe and we see a set of compact patterns inside, etc.} 
\end{figure}

It should be stressed that this vision is not a theorem. It is proposed instead of typical dominance of smooth or even uniform distributions that populate theoretical studies in machine learning. On another hand, hierarchical structure was observed in various data analytics exercises: if there exists a natural semantic structure then we expect that data have the corresponding cluster structure. Moreover, various preprocessing operations make this structure more visible (see, for example, discussion of preprocessing in Appendix~\ref{PrePostclassical}). 

The compact embedding idea was recently explicitly used in data analysis (see, for example, \cite{LiuShaoLi2015,Vemulapalli2019,Bhattarai2019}).

The infinite-dimensional representation and compact embedding hypothesis brings light to the very popular phenomenon of vulnerability of AI decisions in high-dimension world.
According to recent research, such vulnerability seems to be a generic property of various formalisations of learning and attack processes  in high-dimensional systems~\mbox{\cite{TyukinEtAl2020,TyukinHigham2021,Colbrook2021}.}

 Let $Q$ be an infinite-dimensional Banach space. The patterns, representations of a pattern, or their images in an observer systems, etc. are modelled below by compact subsets of $Q$.

\begin{Theorem}[Theorem of high-dimensional vulnerability]\label{Th:vulnerable}Consider two compact sets, $K_{0,1} \subset Q$.
 For {\em almost every}  $\boldsymbol{y}\in Q$   there exists such continuous linear  functional  $\boldsymbol{l}$ on $Q$,  $\boldsymbol{l}\in Q^*$, that 
\begin{equation}\label{perturbation}
\boldsymbol{l}(\boldsymbol{x_1}-\boldsymbol{x_0})>0 \;\;\mbox{for all}\;\;\boldsymbol{x_0}\in K_0, \, \boldsymbol{x_1} \in (K_1+\boldsymbol{y}).
\end{equation}
\end{Theorem}
In particular, for every $\varepsilon >0$ there exist such  $\boldsymbol{y}\in Q$   and  continuous linear  functional  $\boldsymbol{l}$ on $Q$,  $\boldsymbol{l}\in Q^*$, that $\|\boldsymbol{y}\| < \varepsilon$ and  (\ref{perturbation}) holds.
If (\ref{perturbation}) holds, then $K_0 \cap (K_1+\boldsymbol{y}) =\emptyset$. The perturbation $\boldsymbol{y}$ takes $K_1$ out of the intersection with $K_0$. Moreover, linear separation of $K_0 $ and perturbed $K_1$ (i.e., $(K_1+\boldsymbol{y})$) is possible for almost always (\ref{perturbation}) (for almost any perturbation).

The definition of ``almost always'' is clarified in detail in Appendix~\ref{almost}. The set of exclusions, i.e., the perturbations that do not satisfy (\ref{perturbation}) in Theorem~\ref{Th:vulnerable}, is completely thin in the following sense, according to Definition~\ref{CompletelyThin}. A set $Y \subset Q$ is completely thin, if for any compact space $K$ the set of continuous maps $\Psi: K \rightarrow Q$ with non-empty intersection $ \Psi(K) \cap Y \neq \emptyset$ is set of first Bair category in the  Banach space $C(K,Q)$ of
continuous maps $K \rightarrow Q$ equipped by the maximum norm.

 \begin{proof}[Proof of Theorem~\ref{Th:vulnerable} ]Let $\overline{\rm co}(V)$ be a closed convex hull of a set $V\subset Q$. The following sets are convex compacts in $Q$: $\overline{\rm co}(K_0)$, $\overline{\rm co}(K_1)$,  and $\overline{\rm co}(K_0)-\overline{\rm co}(K_1)$. Let 
\begin{equation}\label{cocoy}
\boldsymbol{y} \notin (\overline{\rm co}(K_0)-\overline{\rm co}(K_1)).
\end{equation}

 Then the set $\overline{\rm co}(K_1)+\boldsymbol{y} - \overline{\rm co}(K_0)$ does not contain zero. It is a convex compact set. According to the Hahn–Banach separation theorem \cite{Rudin1991}, there exists a continuous linear separating functional  $\boldsymbol{l}\in Q^*$ that separates the convex compact  $\overline{\rm co}(K_1)+\boldsymbol{y} - \overline{\rm co}(K_0)$ from 0. The same functional separates its subset, $K_1+\boldsymbol{y}-K_0$ from zero, as required. 

The set of exclusions, $\overline{\rm co}(K_0)-\overline{\rm co}(K_1)$ (see (\ref{cocoy})) is a compact convex set in $Q$. According to Riesz's theorem, it is nowhere dense in $Q$ \cite{Rudin1991}. Moreover,  for any compact space $K$ the set of continuous maps $\Psi: K \rightarrow Q$ with non-empty intersection $ \Psi(K) \cap Y \neq \emptyset$ is a nowhere dense subset of  Banach space $C(K,Q)$ of
continuous maps $K \rightarrow Q$ equipped by the maximum norm.

Indeed, let  $ \Psi(K) \cap Y \neq \emptyset$. The set  $ \Psi(K)  $ is compact. Therefore, as it is proven, an arbitrary small perturbation $\boldsymbol{y}$ exists that takes $ \Psi(K) $ out of the intersection with $ Y$:  $(\Psi(K)+\boldsymbol{y})\cap Y=\emptyset$. The minimal value 
$$\min_{\boldsymbol{x_1}\in (\Psi(K)+\boldsymbol{y}), \;\boldsymbol{x_2}\in  Y}
\|\boldsymbol{x_1}-\boldsymbol{x_2}\|=\delta>0$$
exists and is positive because compactness $(\Psi(K)+\boldsymbol{y})$ and $Y$.

Therefore, $\Psi'(K)\cap Y=\emptyset$ for all $\Psi'$ from a ball of maps in $C(K,Q)$

$$\left\{\Psi' \, \left| \|\Psi'-(\Psi+\boldsymbol{y})\|<\frac{\delta}{2}\right.\right\}$$

This proofs that  the set of continuous maps $\Psi: K \rightarrow Q$ with non-empty intersection $ \Psi(K) \cap Y $ is a nowhere dense subset of $C(K,Q)$. Thus, the set of exclusions is completely thin.  
\end{proof}

The following Corollary is simple but it may seem counterintuitive: 
\begin{Corollary}
A compact set $K_0\subset Q$ can be  separated  from a countable set of compacts $K_i\subset Q$ by a single and arbitrary small perturbation $\boldsymbol{y}$ ($\boldsymbol{y}<\varepsilon$ for an arbitrary $\varepsilon>0$):
$$(K_0 +\boldsymbol{y})\cap K_i =\emptyset.$$

 Almost all perturbations $\boldsymbol{y}\in Q$ provide this separation and the set of exclusions is completely thin.
\end{Corollary}
\begin{proof}
First, refer to Theorem~\ref{Th:vulnerable} (for separability of $K_0$ from one $K_i$). Then mention that countable union of completely thin set of exclusions is completely thin, whereas the whole $Q$ is not (according to the Bair theorem, $Q$ is not a set of first category). 
\end{proof}

Separability theorems for compactly embedded patterns might explain why the vulnerability to adversarial  perturbations and stealth attacks is typical for high-dimensional AI systems based on data \cite{TyukinEtAl2020,TyukinHigham2021}. Two properties are important simultaneously: high dimensionality and compactness of patterns.

 \section{Multi-Correctors  of  AI Systems \label{Multicluster}}
 
 \subsection{Structure of Multi-Correctors}
 
\begin{figure}[H]
\includegraphics[width=0.55\textwidth]{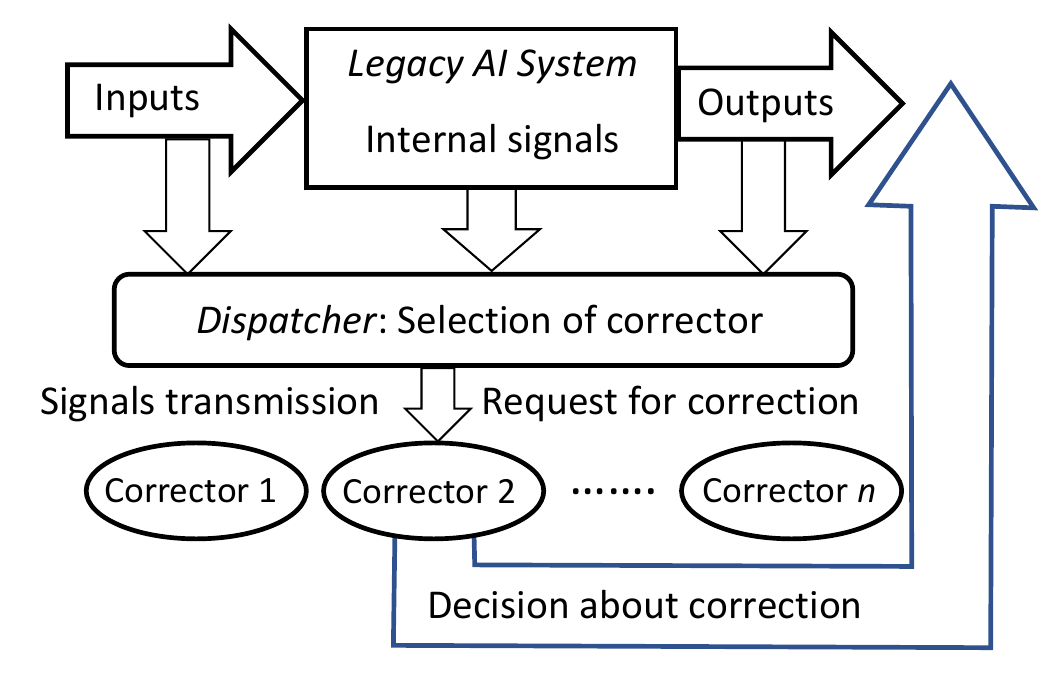}
\caption{Multi-corrector---a system of elementary correctors, controlled by the dispatcher, for reversible correction of legacy AI systems. The dispatcher receives signals from the AI system to be corrected (input signals of the AI system, internal signals generated in the decision-making process, and output signals) and selects from the elementary correctors the one that most corresponds to the situation and will process this situation to resolve the issue of correction. The decision rule, on the basis of which the dispatcher distributes situations between elementary correctors, is formed as a result of a cluster analysis of situations with diagnosed errors. Each elementary corrector processes situations from one cluster. When new errors are detected, the dispatcher modifies the definition of clusters. Cluster models are prepared and modified using the data stream online algorithms. \label{ComplexCorrector}}
\end{figure}

In this section, we present the construction of error correctors for multidimensional AI systems operating in a multidimensional world. It combines a set of elementary correctors (Figure~\ref{SingleCorrector}) and a dispatcher that distributes the tasks between them. The population of possible errors is presented as a collection of clusters. Each elementary corrector works with its own cluster of situations with a high risk of error. It includes a binary classifier that separates that cluster from the rest of situations. Dispatcher is based on an unsupervised classifier that performs cluster analysis of errors, selects the most appropriate cluster for  each operating situation, transmits the signals for analysis to the corresponding elementary corrector, and requests the correction decision from it (Figure~\ref{ComplexCorrector}).

In brief, operation of multi-correctors (Figure~\ref{ComplexCorrector}) can be described as follows:
\begin{itemize}
\item The correction system is organised as a set of elementary correctors, controlled by the dispatcher;
\item Each elementary corrector `owns' a certain class of errors and includes a binary classifier that separates situations with a high risk of these errors, which it owns, from other situations;
 \item For each elementary corrector, a modified rule is set for operating of the corrected AI system in a situation with a high risk of error diagnosed by the classifier of this corrector;
\item The input to the corrector is a complete vector of signals, consisting of the input, internal, and output signals of the corrected Artificial Intelligence system, (as well as, if available, any other available attributes of the situation);
\item The dispatcher distributes situations between elementary correctors;
\item The decision rule, based on which the dispatcher distributes situations between elementary correctors, is formed as a result of cluster analysis of situations with diagnosed errors;
\item Cluster  analysis of situations with diagnosed errors is performed using an  online  algorithm;
\item Each elementary corrector owns situations with errors from a single cluster;
\item After receiving a signal about the detection of new errors, the dispatcher modifies the definition of clusters according to the selected online algorithm and accordingly modifies the decision rule, on the basis of which situations are distributed between elementary correctors;
\item After receiving a signal about detection of new errors, the dispatcher chooses an elementary corrector, which must process the situation, and the classifier of this corrector learns according to a non-iterative explicit rule.
\end{itemize}

Flowcharts of these operations are presented in Appendix~\ref{Sec:Flowcharts}. 
Multi-correctors  satisfy  the following requirements: 
\begin{enumerate}
\item Simplicity of construction; 
\item Correction should not damage the existing skills of the system; 
\item Speed (fast non-iterative learning); 
\item Correction of new errors without destroying previous corrections.
\end{enumerate}

For implementation of this structure, the construction of classifiers for elementary correctors and the online algorithms for clustering should be specified. For elementary correctors many choices are possible, for example: 
\begin{itemize}
\item Fisher's linear discriminant is simple, robust, and is proven to be applicable in high-dimensional data analysis \cite{Gorbetal2018,Grechuk2021};
\item Kernel versions of non-iterative linear discriminants extend the area of application
of the proposed systems, their separability properties were quantified and tested \cite{TyukinKernel2019};
\item Decision trees of mentioned elementary discriminants with bounded depth. These algorithms require small (bounded) number  of iterations. 
 \end{itemize}

The population of clustering algorithms is huge \cite{XuWunsch2008}. The first choice for testing of multi-correctors \cite{TyukinCluster2021} was partitioning around centroids by $k$ means algorithm. The closest candidates for future development are multi-centroid algorithms that present clusters by  networks if centroids  (see, for example, \cite{TaoCW2002}. This approach to clustering meets the idea of compact embedding, when the network of centres corresponds to the $\varepsilon $-net approximating the compact.
 
\subsection{Multi-correctors in Clustered Universe: A Case Study \label{CaseStudy}}

\subsubsection{Datasets}

In what follows our use-cases will evolve around a standard problem of supervised multi-class classification. In order to be specific and to ensure reproducibility of our observations and results, we will work with a well-known and widely available CIFAR-10 dataset  \cite{krizhevsky2009learning, CIFAR10}.  The CIFAR-10 dataset is a collection of $32\times 32$ colour images that are split across $10$ classes: 
\[
\mbox{`airplane', `automobile', `bird', `cat', `deer', `dog', `frog', `horse', `ship', `truck'}
\]
with `airplane' being a label of Class $1$, and `truck' being a label of Class $10$. The original CIFAR-10 dataset is further split into two subsets: a {\it training} set containing $5000$ images per class (total number of images in the training set is 50,000), and a {\it testing} set with $1000$ images per class (total number of images in the testing set is 10,000).

\subsubsection{Tasks and Approach}

We focus on two fundamental tasks: for a given {\it legacy classifier}:
\begin{itemize}
\item (Task 1) devise an algorithm to {\it learn a new class} without catastrophic forgetting and retraining,  and;
\item (Task 2) develop an algorithm to {\it predict} classification errors in the legacy classifier.
\end{itemize}

Let us now specify these tasks in more detail. 

As a legacy classifier we have used a deep convolutional neural network whose structure is shown in Table \ref{table:network_structure}.  The network's training set comprised 45,000 images corresponding to Class $1$--$9$ ($5000$ images per class), and the test set comprised $9000$ images from the CIFAR-10  testing  set ($1000$ images per class). No data augmentation was invoked as a part of the training process. The network by stochastic gradient descent with the momentum parameter was set to $0.9$ and mini-batches were of size $128$.  Overall, we trained the network over $70$ epochs executed in $7$ training episodes of $10$-epoch training, and the learning rate was equal to $0.1/(1+0.001 k)$, where $k$ is the index of a training instance (a mini-batch) within a training episode. 

The network's accuracy, expressed as the percentage of correct classifications, was $0.84$ and $0.73$ on the training and testing sets, respectively (rounded to the second decimal point). The network was trained in MATLAB R2021a. 
Each $10$-epoch training episode took approximately $1.5$ h to complete on an HP Zbook 15 G3 laptop with a Core i7-6820HQ CPU, $16$ Gb of RAM, and Nvidia Quadro 1000M GPU.

\begin{table}
  \caption{Architecture of the {\it legacy classifier}}
  \label{table:network_structure}
  \centering
    \def\arraystretch{1.2}
  \small
  \begin{tabular}{|c|c|c|}
    \hline
   Layer number     &  Type     & Size  \\
    \hline
    1  &  Input  & $32\times32\times 3$    \\
    \hline
    2     & Conv2d & $4\times4\times 64$    \\
    3     & ReLU    &   \\
    4     & Batch normalization       &    \\
    5     & Dropout $0.25$     &   \\
    \hline
    6    & Conv2d & $2\times2\times 64$    \\
    7     & ReLU    &   \\
    8     & Batch normalization       &    \\
    9     & Dropout $0.25$      &   \\
    \hline
    10    & Conv2d & $3\times3\times 32$    \\
    11    & ReLU    &   \\
    12     & Batch normalization       &    \\
    13    & Dropout $0.25$      &   \\
    \hline
    14    & Conv2d & $3\times3\times 32$    \\
    15    & ReLU    &   \\
    16     & Batch normalization       &    \\
    17     & Maxpool        & pool size $2\times 2$, stride $2\times 2$  \\
    18    & Dropout $0.25$      &   \\ 
   \hline     
    19     & Fully connected     & 128 \\
    20     & ReLU     &  \\
    21    & Dropout $0.25$      &   \\ 
    \hline
    22     & Fully connected     & 128 \\
    23     & ReLU     &  \\
    24    & Dropout $0.25$      &   \\ 
\hline
    25     & Fully connected     & 9 \\
    26     & Softmax     & 9 \\
     \hline
  \end{tabular}
\end{table}

{\it Task 1 (learning a new class)}. Our first task was to equip the trained network with a capability to learn a new class without expensive retraining. In order to achieve this aim we adopted an approach and algorithms presented in \cite{GorTyukPhil2018,TyukinCluster2021}. According to this approach, for every input image $\boldsymbol{u}$ we generated its latent representation $\boldsymbol{x}$ of which the composition is shown in Table~\ref{table:feature_vector}.  In our experiments we kept all dropout layers active after training. This was implemented by using ``forward'' method instead of ``predict'' when accessing feature vectors of relevant layers in the trained network. The procedure enabled us to simulate an environment in which  AI correctors operate on data that are subjected to random perturbations.

 This process constituted our legacy AI system.

\begin{table}
  \caption{Latent representation of an image}
  \label{table:feature_vector}
  \centering
    \def\arraystretch{1.2}
  \small
  \begin{tabular}{|c|c|c|c|c|}
    \hline
   Attributes      &  $\boldsymbol{x}_1,\dots,\boldsymbol{x}_9$       & $\boldsymbol{x}_{10},\dots,\boldsymbol{x}_{137}$   & $\boldsymbol{x}_{138},\dots,\boldsymbol{x}_{265}$ & $\boldsymbol{x}_{266},\dots,\boldsymbol{x}_{393}$\\
    \hline
   Layers         & 26 (Softmax)  & 19 (Fully connected) & 22 (Fully connected)  &  23 (ReLU)\\
     \hline
  \end{tabular}
\end{table}

Using these latent representations of images, we formed two sets: $\mathcal{X}$ and $\mathcal{Y}$. The set $\mathcal{X}$ contained latent representations of the new class (Class $10$---`trucks') from the CIFAR-10 training set ($5000$ images), and the set $\mathcal{Y}$ contained latent representations of all other images in CIFAR-10 training set (45,000 images). These sets have then been used to construct a multi-corrector in accordance with the following algorithm presented in \cite{TyukinCluster2021}.

\begin{algorithm}
\caption{(Few-shot AI corrector \cite{TyukinCluster2021}: 1NN version. Training). Input: sets $\mathcal{X}$, $\mathcal{Y}$; the number of clusters, $k$; threshold, $\theta$ (or thresholds $\theta_1,\dots,\theta_k$).}
\label{alg:fast}
	\begin{enumerate}[leftmargin=0.7 cm]
		\item Determining the centroid  $\bar{\boldsymbol{x}}$ of the $\mathcal{X}$. Generate two sets, $\mathcal{X}_c$, the centralised\linebreak set $\mathcal{X}$, and $\mathcal{Y}^{\ast}$, the set obtained from $\mathcal{Y}$ by subtracting $\bar{\boldsymbol{x}}$ from each of its\linebreak  elements.
		\item  Construct Principal Components for the centralised set $\mathcal{X}_c$.
		\item  Using Kaiser, broken stick, conditioning rule, or otherwise, select $m\leq n$\linebreak  Principal Components, $h_1,\dots,h_m$, corresponding to the first largest\linebreak  eivenvalues $\lambda_1\geq \cdots \geq \lambda_m >0$ of the covariance matrix of the set $\mathcal{X}_c$,\linebreak  and project the centralized set $\mathcal{X}_c$ as well as $\mathcal{Y}^{\ast}$ onto these vectors.\linebreak  The operation returns sets $\mathcal{X}_r$ and $\mathcal{Y}_{r}^{\ast}$, respectively:
	\[
		\begin{aligned}
		\mathcal{X}_r&=\{\boldsymbol{x} | \boldsymbol{x}= H \boldsymbol{z}, \ \boldsymbol{z}\in\mathcal{X}_c\}\\
		\mathcal{Y}_r^\ast&=\{\boldsymbol{y} | \boldsymbol{y}=H \boldsymbol{z}, \ \boldsymbol{z}\in\mathcal{Y}^\ast\}, \ H=\left(\begin{array}{c} h_1^T \\ \vdots \\ h_m^T\end{array}\right).
		\end{aligned}
		\]
		\item Construct matrix $W$
	\[
	W=\mathrm{diag} \left(\frac{1}{\sqrt{\lambda_1}},\dots,\frac{1}{\sqrt{\lambda_m}}\right)
	\]
corresponding to the whitening transformation for the set $\mathcal{X}_r$. Apply the\linebreak  whitening transformation to sets $\mathcal{X}_r$ and $\mathcal{Y}_{r}^\ast$. This returns sets $\mathcal{X}_w$ and $\mathcal{Y}_{w}^{\ast}$:
	\[
	\begin{aligned}
		\mathcal{X}_w&=\{\boldsymbol{x} | \boldsymbol{x}= W \boldsymbol{z}, \ \boldsymbol{z} \in\mathcal{X}_r\}\\
		\mathcal{Y}_w^\ast&=\{\boldsymbol{y} | \boldsymbol{y}=W \boldsymbol{z}, \ \boldsymbol{z} \in\mathcal{Y}_r^\ast\}.
		\end{aligned}
	\]
		\item Cluster the set $\mathcal{Y}_{w}^\ast$ into $k$ clusters $\mathcal{Y}_{w,1}^\ast,\dots,\mathcal{Y}_{w,k}^\ast$ (using e.g. the k-means\linebreak  algorithm or otherwise). Let $\bar{\boldsymbol{y}}_1,\dots,\bar{\boldsymbol{y}}_k$ be their corresponding centroids.
		\item  For each pair  $(\mathcal{X}_w,\mathcal{Y}_{w,i}^\ast)$, $i=1,\dots,k$, construct (normalised)  Fisher\linebreak  discriminants  $\boldsymbol{w}_1,\dots,\boldsymbol{w}_k$:
		\[
		\boldsymbol{w}_i= \frac{(\mathrm{Cov}(\mathcal{X}_w)+\mathrm{Cov}(\mathcal{Y}_{w,i}^\ast))^{-1} \bar{\boldsymbol{y}}_i }{\|(\mathrm{Cov}(\mathcal{X}_w)+\mathrm{Cov}(\mathcal{Y}_{w,i}^\ast))^{-1} \bar{\boldsymbol{y}}_i \|}.
		\]
An element $\boldsymbol{z}$ is associated with the set $\mathcal{Y}_{w,i}^\ast$ if $(\boldsymbol{w}_i,\boldsymbol{z})>\theta$ and with the set\linebreak   $\mathcal{X}_w$ if $(\boldsymbol{w}_i,\boldsymbol{z})\leq\theta$. 

If multiple thresholds are given then an element $\boldsymbol{z}$ is associated with the\linebreak  set $\mathcal{Y}_{w,i}^\ast$ if $(\boldsymbol{w}_i,\boldsymbol{z})>\theta_i$ and with the set  $\mathcal{X}_w$ if $(\boldsymbol{w}_i,\boldsymbol{z})\leq\theta_i$. 
	\end{enumerate}
Output: vectors $\boldsymbol{w}_i$, $\bar{\boldsymbol{x}}$, $i=1,\dots,k$, matrices $H$ and $W$.
\end{algorithm}

Integration logic of the multi-corrector into the final system was as follows \cite{TyukinCluster2021}:

\begin{Remark}\normalfont Since the set $\mathcal{Y}$ corresponds to data samples from previously learned classes, a positive response in the multi-corrector (condition $(\boldsymbol{w}_{\ell},\boldsymbol{x}_w)>\theta$ holds) 'flags' that this data point is to be associated with classes that have already been learned (Classes $1$--$9$). Absence of a positive response indicates that the data point is to be associated with the new class (Class $10$).
\end{Remark}

{\it Task 2 (predicting errors of a trained legacy classifier)}. In addition to learning a new class without retraining, we considered the problem of predicting correct performance of a trained legacy classifier. In this setting, the set $\mathcal{X}$ of vectors corresponding to {\it incorrect} classifications on CIFAR-10 {\it training} set, and the set $\mathcal{Y}$ contained latent representations of images  form CIFAR-10 training set that have been correctly classified. Similar to the previous task, predictor of the classifier's error was constructed in accordance with Algorithms \ref{alg:fast} and \ref{alg:fast:test}.

\begin{algorithm}
\caption{(Few-shot AI corrector \cite{TyukinCluster2021}: 1NN version. Deployment). Input: a data vector $\boldsymbol{x}$, the set's $\mathcal{X}$ centroid vector $\bar{\boldsymbol{x}}$, matrices $H$, $W$,  the number of clusters, $k$, cluster centroids $\bar{\boldsymbol{y}}_1,\dots,\bar{\boldsymbol{y}}_k$,  threshold, $\theta$ (or thresholds $\theta_1, \dots, \theta_k$), discriminant vectors, $\boldsymbol{w}_i$, $i=1,\dots,k$.}
\label{alg:fast:test} 

	\begin{enumerate}
		\item  Compute
		\[
		\boldsymbol{x}_w=W H (\boldsymbol{x}-\bar{\boldsymbol{x}})
		\]
		\item  Determine
		\[
		\ell=\arg \min_{i} \|\boldsymbol{x}_w-\bar{\boldsymbol{y}}_i\|.
		\]
		\item   Associate the vector  $\boldsymbol{x}$ with the set $\mathcal{Y}$ if $(\boldsymbol{w}_{\ell},\boldsymbol{x}_w)>\theta$ and with the set $\mathcal{X}$\linebreak  otherwise. If multiple thresholds are given then associate the vector  $\boldsymbol{x}$\linebreak  with the set $\mathcal{Y}$ if $(\boldsymbol{w}_{\ell},\boldsymbol{x}_w)>\theta_\ell$ and with the set $\mathcal{X}$ otherwise.
	\end{enumerate}
	
Output: a label attributed to the vector $\boldsymbol{x}$.
\end{algorithm}

{\it Testing protocols.}  Performance of the algorithms was assessed on CIFAR-10 {\it testing} set. For Task 1, we tested how well our new system---the legacy network shown in \mbox{Table~\ref{table:network_structure}} combined with the multi-corrector constructed by Algorithms \ref{alg:fast} and \ref{alg:fast:test}---performs on images from CIFAR-10 {\it testing} set. For Task 2, we assessed how well the multi-corrector, trained on CIFAR-10 {\it training} set, predicts errors of the legacy network for images of $9$ classes (Class $1$---$9$) taken from CIFAR-10 {\it testing} set.

\subsubsection{Results}

\begin{figure}[h]
\centering
\includegraphics[width=0.9\textwidth]{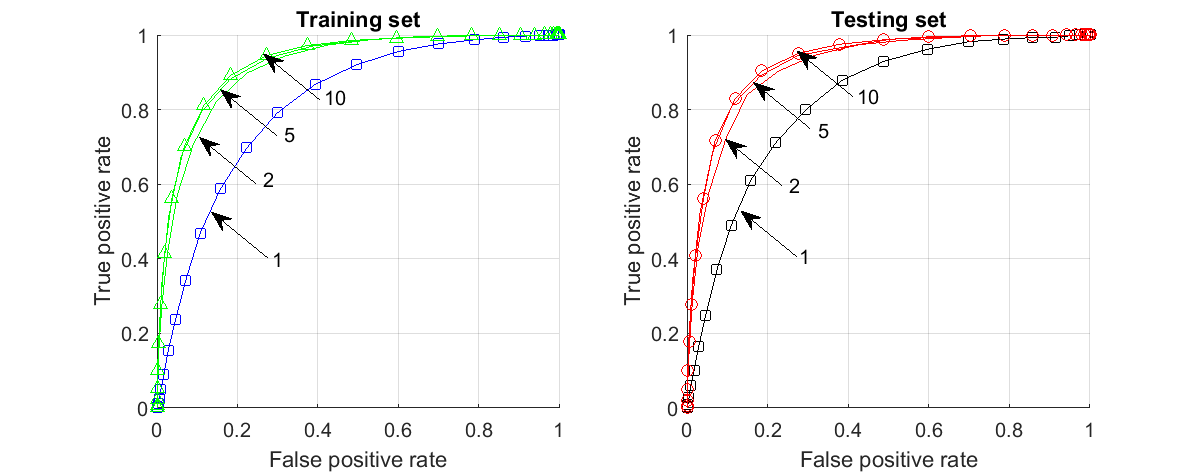}
\caption{Clustered universe in learning a new class. Arrows and numbers show the number of clusters in the multi-corrector for which that specific ROC curve was constructed. {  The squares (blue on the left and black on the right) correspond to an elementary corrector with one cluster, other lines (green on the right and red on the left) correspond to the multi-correctors with 2,5, and 10 clusters.}}\label{fig:new_class}
\end{figure}

{\it Task 1 (learning a new class)}. Performance of the multi-corrector in the task of learning a new class is illustrated in Figure \ref{fig:new_class}. In these experiments, we projected onto the first $20$ principal components. The rationale for choosing these $20$ principal components was that for these components the ratio of the largest eigenvalue to the eigenvalue that is associated with the principal component is always smaller than $10$. The figure shows ROC curves in which true positives are images from the new class and identified as a new class, and False positives are defined as images from already learned classes (Classes $1$---$9$) but identified as a new class (Class $10$) by the combined system. As we can see from Figure \ref{fig:new_class}, performance of the system saturates at about $10$ clusters which indicates a peculiar granular structure of the data universe in this example: clusters are apparently not equal in terms of their impact on the overall performance, and the benefit of using more clusters decays rapidly as the number of clusters grows.


\begin{figure}
\centering
\includegraphics[width=0.9\textwidth]{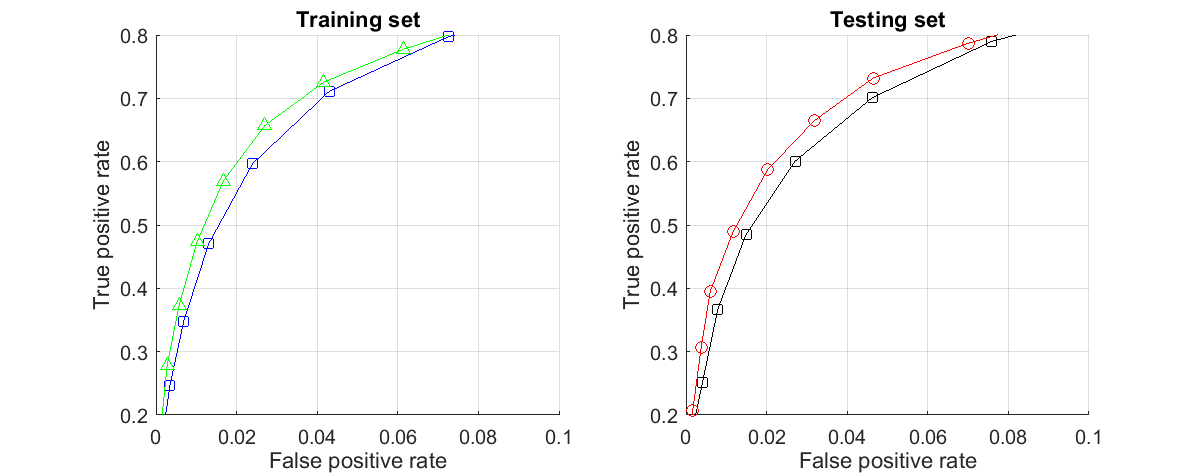}
\includegraphics[width=0.9\textwidth]{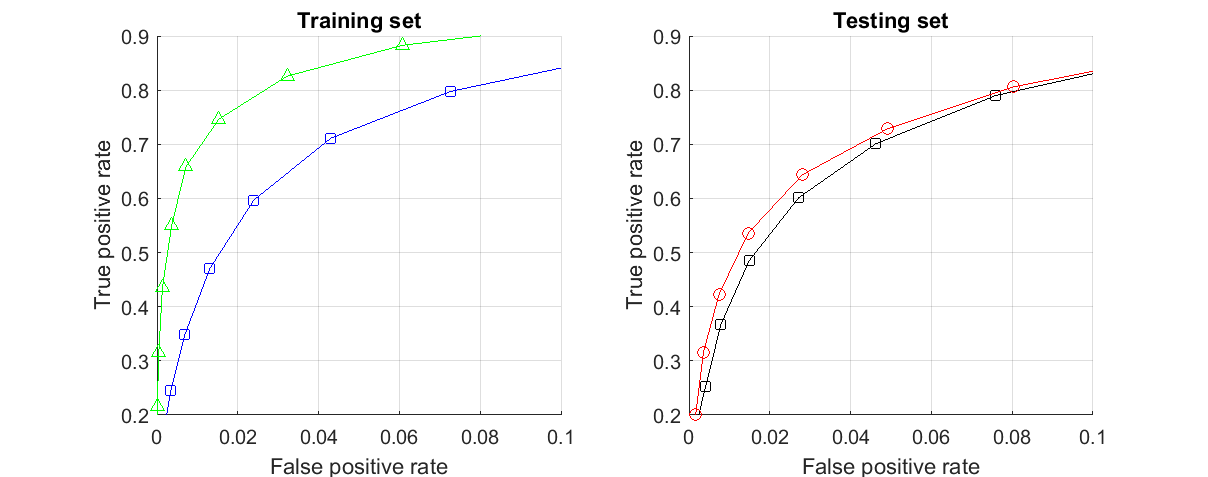}
\includegraphics[width=0.9\textwidth]{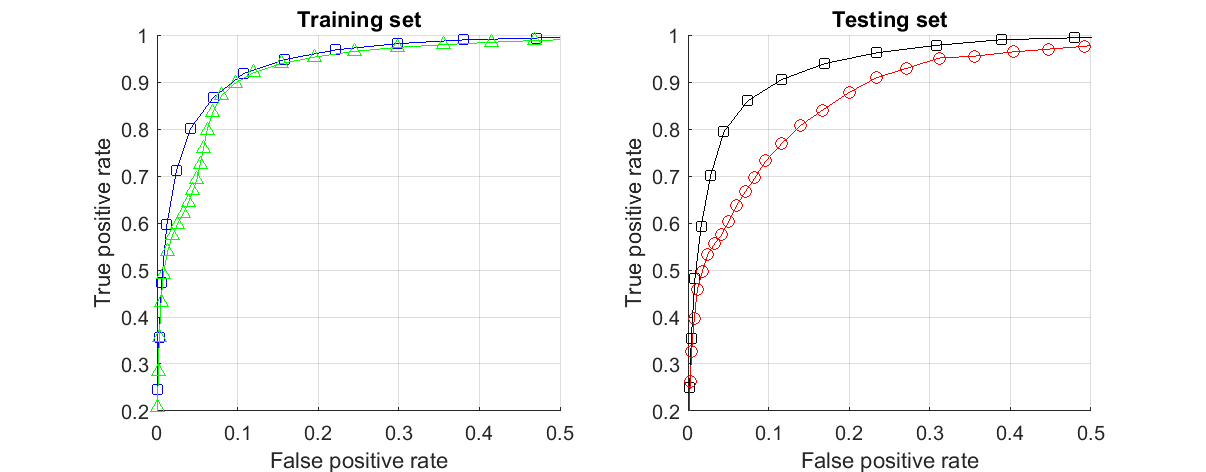}
\caption{Clustered universe in learning a new class - the impact of dimension of the ambient space. Curves marked with squares { (blue on the left and black on the right) correspond to corrector with a single cluster, curves marked by green triangles on the left and and red  circles on the right } correspond to correctors with multiple clusters.  Top panel: the application of Algorithm \ref{alg:fast} and \ref{alg:fast:test} to the same data but with retained first $100$ principal components instead of the first $20$ components (see Figure \ref{fig:new_class}). Middle panel: projecting onto the first $100$ principal components and using $300$ clusters. Bottom panel: projecting onto the first $300$ principal components and using $300$ clusters.}\label{fig:new_class_dimensions}
\end{figure}

We note that the system performance and generalisation depends on both ambient dimension (the number of principal components used) and the number of clusters. This phenomenon is illustrated in in Figure \ref{fig:new_class_dimensions}. When the number of dimensions increases (top row in Figure \ref{fig:new_class_dimensions}), the gap between a single-cluster corrector and a multi-cluster corrector narrows. Yet, as can be observed from this experiment, the system generalises well. 

When the number of clusters increases from $10$ to $300$, the system overfits. This is not surprising as given the size of our training set (50,000 images to learn from) splitting the data into $300$ clusters implies that each $100$-dimensional discriminant in \mbox{Algorithm \ref{alg:fast}} is constructed, on average, from mere $170$ samples. The lack of data to learn from and 'diffusion' and shattering of clusters in high dimension could be contributors to the instability. Nevertheless, as the right plot shows, the system still generalises at the level that is similar to the $10$-cluster scenario. 

When the ambient dimension increases further we observe a dramatic performance collapse for the multi-corrector constructed by Algorithms \ref{alg:fast} and \ref{alg:fast:test}. Now $300$-dimensional vectors are built from on average $170$ points. The procedure is inherently unstable and in this sense  such results are expected in this limit.

\begin{figure}[H]
\centering
\includegraphics[width=0.9\textwidth]{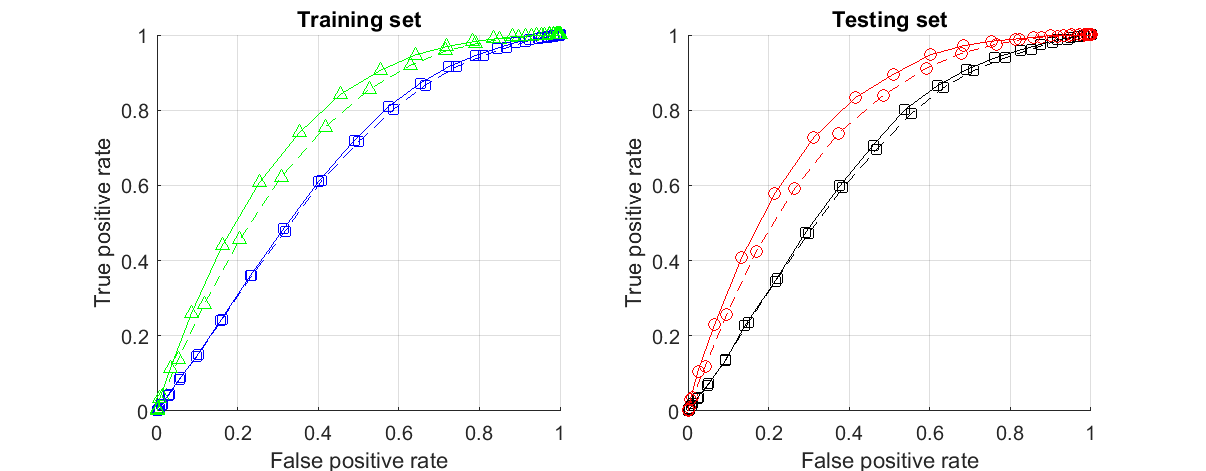}
\caption{{ Prediction of errors. Solid curves marked by green triangles (on the left)  and red circles (on the right) correspond to  $10$-cluster multi-corrector. Solid curves marked by squares (blue on the left and black on the right) are produced by a single-cluster elementary corrector.} Dashed lines with the same marks show performance of the same system but constructed on data sets in the reduced feature space formed by attributes  $1-137$ (see Table \ref{table:feature_vector}).}\label{fig:prediction_errors}
\end{figure}

{\it Task 2 (predicting errors)}. A very similar picture occurs in the task of predicting errors of legacy classifiers. For our specific case, performance of  $10$-cluster multi-corrector with projection onto $20$ principal components in shown in Figure \ref{fig:prediction_errors}. In this task, true positives are  errors of the original classifier which have been correctly identified as errors by the corrector. False positives are data correctly classified by the original deep neural network but which nevertheless have been labelled as errors by the corrector. According to Figure \ref{fig:prediction_errors}, the multi-corrector model generalises well and delivers circa $70\%$ specificity and sensitivity on the test set. 

Another interesting phenomenon illustrated by  Figure \ref{fig:prediction_errors} is the apparent importance of how the information from the legacy AI model is aggregated into correcting cascades. Dashed lines in Figure \ref{fig:prediction_errors} show what happens if latent representations are formed by signals taken from layers $26$ and $19$ only. In this case the impact of clustering becomes less pronounced, suggesting the importance of feature selection for optimal performance.

{\it Computational efficiency.} Computational costs of constructing multi-correctors is remarkably small. For example, learning a new class with a $10$-cluster multi-corrector and $20$~principal components took $1.32$ seconds on the same hardware used to train the original legacy classifier. When the number of clusters and dimension increases to $300$ and $300$, respectively, the amount of time needed to construct the multi-corrector was $37.7$ s. These figures show that not only clustered universes and multi-correctors are feasible in applications but they are also extremely efficient computationally. We do not wish to suggest that they are a replacement of deeper retraining. Yet, as we see from these experiments, they can be particularly efficient in the tasks of incremental learning---learning an additional class in a multi-class problem---if implemented appropriately.

\subsubsection{Dimensionality and Multi-Corrector Performance}

 The CIFAR-10 { training} set contains 5000 images per class, and   the testing set contains 1000 image per class.  The total number of data samples is 60,000. Dimension of the input space is 3072. Dimension of the space of latent representation is 393. The shortened feature space with coordinates $x_{1}-x_{137}$ is also used. Three versions of PCA dimensionality reduction were tested, with 20, 100, and 300 principal components. We can see that the number of samples significantly exceeds all the dimensions (60,000 versus 20, 100,  137, 300, and 393). The question arises: is this classical or already postclassical zone of data dimensionality (see Figure~\ref{Fig:Post})? 

 Compare the number of samples to the critical size $|Y|$ of the dataset $Y$ that allows one to separate a random point $\boldsymbol{x}$ from the set $Y$ by Fisher's discriminant (Definition~\ref{def:Fisher0}) with threshold $\alpha=0.8$ and probability $p=1-\delta=0.99$. Theorem~\ref{Th:prototype} gives this estimate. If $\boldsymbol{x}$ is uniformly distributed in a ball then, according to Theorem~\ref{Th:prototype}, we produce the following table.

  Table~\ref{Table6} ensures us that for dimensions 100, 137,  300, and 393 the CIFAR-10
dataset is very deeply in the postclassical area. The only question appeared for dimension  20. Theorem~\ref{Th:prototype} gives that for this dimension, the postclassical area ends at $|Y|>121$. Nevertheless, the multi-correctors work well in this dimension.  The reason for this efficiency could be the fine-grained cluster structure of the dataset.  Theorem~\ref{Th:prototype} is true for {\em any } dataset $Y$ without any hypothesis about data distribution. It estimates the number of points $|Y|$. On the contrary, according to Theorems~\ref{Th:SphericGranules} and  \ref{Th:EllGranules}, for a fine-grained structure the number of granules should be counted and not the number of points.

\begin{table}[h]  
\begin{center}
\caption{\label{Table6} The upper bound on $|Y|$ that guarantees separation of a random point $\boldsymbol{x}$, uniformly distributed in a ball,  from set $Y$  by Fisher's discriminant with probability 0.99 for  $\alpha = 0.8$,  in various dimensions.}
\begin{tabular}{ rlllll } 
 \hline
 $n$ &  20 & 100 & 137 & 300 & 393  \\
$|Y|\leq$& $1.21\times 10^2$ & $2.58\times 10^{18}$ & $9.21 \times 10^{25}$ & $1.72 \times 10^{59}$ & $1.65\times 10^{78} $\\
 \hline
\end{tabular}
\end{center}
\end{table}

Stochastic separation theorems are needed to evaluate the areas of applicability of machine learning algorithms in the multidimensional world of postclassical data. They also provide ideas for developing appropriate algorithms. The first stochastic separation theorems  led to elementary correctors (Figure~\ref{SingleCorrector})  \cite{GorbanTyuRom2016,GorbanTyukinNN2017}.  The theorems for data with fine-grained distributions are related to the multi-corrector algorithm. Of course, the detailed structure of multi-correctors may vary, and in this work we considered the first and basic version.

\section{Conclusions \label{Conclusion}}

In this work, we used the modified Donoho's definition of postclassical data (\mbox{Section~\ref{Postclassical}}). The postclassical data are defined by relations between the intrinsic dimensionality of the data $\dim(DataSet)$ and the logarithm of the number of data samples (\ref{logarithmicPost}),  $\dim(DataSet) \gg \log N$. In the postclassical area (Figure~\ref{Fig:Post}), the classical statistical learning approaches may become useless and the phenomena of curse and blessing of dimensionality become important. Among these phenomena are quasiorthogonality \cite{Kurkova1993, KainenKurkova2020, GorbTyuProSof2016}, systematically used in our work, and stochastic separation theorems \cite{GorbanTyukinNN2017,Grechuk2021}.

Distributions of data in real life tasks can be far from any regular distribution.  One of the typical phenomena is rich cluster structure. Multi-clustering and recently described hubness phenomena are important  in high-dimensional data analysis and it is impossible to analyse the real life datasets without accounting of them \cite{Cai2010, Mani2019, Ma2020, Albergante2020}. We used the granular distributions as models for multi-clustered data. Three models of clusters are proposed: spherical clusters, elliptic clusters, and superstatistical model, where clusters are represented by the peaks of distribution density and the whole distribution is a random mixture of these such peaks.

Hypothesis of compactness of granules has different forms for these cluster models. For spherical clusters, compactness is considered as a relatively small diameter of the granules comparing to the data standard deviation. This approach is close to the Duin measurement of compactness \cite{Duin1999}. For the elliptic granules, the diameter can be large, but the sequence of the main diameters should decay. This idea is borrowed from functional analysis, the theory of Kolmogorov $n$-width \cite{Kolmogoroff1936, Tikhomirov1960, Dung2013}
 in its simplest form. 

In Section~\ref{StochFineGrain}, we formulated and proved stochastic separation theorems for fine-grained distributions. Instead of separation of random points we considered separation of clusters. The multi-clustered datasets  demonstrate the curse and blessing of dimensionality effects for smaller dimensions than the classical distributions with the same number of data points because these effects depend on the number of clusters and their compactness characteristics, see Theorem~\ref{Th:SphericGranules} for spheric granules, Theorems~\ref{Th:EllGranules} and  \ref{RandomGranule} for elliptic granules, and Propositions \ref{prop:logconc}--\ref{prob:anyz} for granules modelled by the distribution peaks of different shapes.

The probability of a multivariate real-life data distribution is usually unknown and we never have enough data to restore the probability density for postclassical data.
Therefore, in Section~\ref{Sec:Compact} we developed the infinite-dimensional approach that does not use the unobservable probability distributions. For measure concentration on spheres and equators, infinite-dimensional limit was considered by L\'{e}vy  in his functional analysis book \cite{Levy1951}. Instead of spheric or elliptic granules, just compact subsets are considered and Theorem~\ref{Th:vulnerable} about separability in families of compact sets explains  why the vulnerability to adversarial  perturbations and stealth attacks is typical for high-dimensional AI systems based on data \cite{TyukinEtAl2020,TyukinHigham2021}. Two properties are important simultaneously: high dimensionality and compactness of patterns.
 
Multi-corrector, a special ideal device for correction of AI errors in the worlds of high-dimensional multi-clustered data, is developed and tested (Section~\ref{Multicluster}). It includes a family of elementary correctors   managed by a dispatcher (Figure~\ref{ComplexCorrector}). The dispatcher distributes situations between elementary correctors using a classification model created in the course of cluster analysis of diagnosed errors. Each elementary corrector deals with its own cluster. Multi-correctors are tested on the CIFAR-10 database solving two tasks: (i) learn a new class (without catastrophic forgetting and retraining) and (ii) predict classification errors. Testing was organised for a different number of principal components involved and for a different number of clusters. The tests demonstrates that the multi-corrector model generalises well with appropriate specificity and sensitivity on the test set. The details are presented in Figures \ref{fig:new_class}--\ref{fig:prediction_errors}.

Several directions of future work have become open. The main challenge is to develop a technology for creating reliable and self-correcting augmented AI ecosystems in which each AI is dressed-up with a cloud of correctors. These correctors increase the reliability of AI by removing errors and at the same time serve as  a special storage device---a memory of detected errors for further interiorisation. The correctors also enable knowledge transfer between AIs and can be used to protect their “host” AI from various attacks by repairing the effects of malicious actions. In addition, they may  model  attacks on AIs \cite{TyukinEtAl2020,TyukinHigham2021}, opening new ways to assess the efficiency of defence measures and protocols employed by AI owners. There are also many special technical questions that require further attention and work. These include the analysis of reducibility of multidimensional data and the development of precise criteria, enabling one to decide if a given dataset is a postclassical dataset, to which our current work applies, or if it is the classical one, to which conventional statistical learning approaches may still be applicable.

 \section{Discussion \label{Discussion}}

The preprocessing in the postclassical data world ({Figure}~\ref{PostClassDim} and {Appendix}~\ref{PrePostclassical}) 
is a challenging task because no classical statistical methods are applicable when the sample size is much smaller than data dimensionality (the Donoho area (Section~\ref{Postclassical}, (\ref{DonohoPost})   \cite{Donoho2000}). The correlation transformation (Appendix~\ref{correlation transformation}) moves data out of the Donoho area yet, certain specific non-classical effects still persist when the sample size remains much smaller than the exponential of the data dimensionality (\ref{logarithmicPost}). Dimensionality reduction methods should combine two sets of goals: sensible grouping and extraction of relevant features. For these purposes, combining supervised and unsupervised learning techniques is necessary. Data labels from supervised approaches add sense and context to the subsequent analysis of unlabelled data. The simple geometric methods like supervised PCA, semisupervised PCA (Appendix~\ref{PCAy}), and Domain Adaptation PCA (DAPCA)  (Appendix~\ref{PCAy}) may serve as prototypes of more complex and less controllable approaches. They can  also be used to simplify large deep learning systems \cite{GorbanMirkesTukin2020}.

Data in postlclassical world are rarefied. At the same time, values of regular functionals on data are concentrated near their median values \cite{GianMilman2000,Vershynin2018}. Combinations of these properties produce the `blessing of dimensionality' \cite{Donoho2000,Kainen1997,Levy1951}. The most important manifestation of these effects for applied data analysis beyond the central limit theorem are quasiorthogonality \cite{Kurkova1993,KainenKurkova2020,GorbTyuProSof2016} and stochastic separation theorems \cite{GorbanTyukinNN2017,Grechuk2021}. These results give the theoretical backgrounds for creation of intellectual devices of a new type: correctors of AI systems.
In this paper, we presented a  new family of stochastic separation theorems for fine-grained data distributions with different geometry of clusters (Section~\ref{StochFineGrain}). These results enable development of multi-correctors for multidimensional AI with a granular distribution of errors. On real data, such correctors showed better performance  than simple correctors. 

Various versions of multi-correctors that provide fast and reversible correction of AI errors should be supplemented by an additional special operation of interiorisation  of corrections. Accumulation of many corrections will, step by step, spend the blessing of dimensionality resource: after implementing  elementary corrections, the probability of success for  new correctors may decrease. This can be considered as accumulation of technical debt.  In psychology, interiorisation is the process of making  skills, attitudes, thoughts, and knowledge an integrated part of one's own being. For large legacy AI systems, interiorisation of corrections means the supervising retraining of the system. Here a complex ``legacy system+multi-corrector'' acts as a supervisor and labels the data, while the system itself learns by assimilating the fast flow of generated data.

The construction of correctors with their subsequent interiorisation can be considered as a tool for solving the problem of model degradation and concept drift. An increase in the error rate is a signal of degradation of the model and a systematic decrease in performance~\cite{Tsymbal2004}. The nature of data changes in time, due to the evolution of the system under analysis. Coping with this phenomena required combination of supervised, semi-supervised, and even unsupervised learning.  Semi-supervised and unsupervised methods help to self-assess model degradation in preprocessing mode in real time and modify the classification model and features before actual errors occur \cite{Cerquitelli2019}. 
Error correctors provide reversible modification of AI systems without iterative retraining and can assimilate significant concept drift.

We refuse the classical hypothesis of the regularity of the data distribution and assume that the data can have a rich fine-grained structure with many clusters and corresponding peaks in the probability density. 
In this work, we generalise this framework and ideas to a much richer class of distributions. We introduce a new model of data---a possibly {\it infinite-dimensional} data universe with hierarchical structure in which each data cluster has a granular internal structure, etc. 
The idealised concept of granular  Hierarchical Universe (Figure~\ref{HierarchicalUniverse})  is intended to replace the ideal picture of a smooth unimodal distribution popular in statistical science.

The infinite-dimensional version of theorems about separation of compact clusters and families of such clusters demonstrates the importance of the hypothesis about compact embedding of data clusters (Section~\ref{Sec:Compact}). The hypothesis of images compactness appeared in data analysis and machine learning several times in many different forms. Perhaps,  it was first introduced by E.M. Braverman \cite{Braverman1967}. This was a guess about the data structure in the real world. It is now widely accepted that real data are rarely i.i.d samples from a regular distribution. Getting the right guess about the distribution of data is essential to the success of machine learning.

According to a modern deep learning textbook, ``the goal of machine learning research is not to seek a universal learning algorithm or the absolute best learning algorithm. Instead, our goal is to understand what kinds of distributions are relevant to the 'real world' that an AI agent experiences and what kinds of machine learning algorithms perform well on
data drawn from the kinds of data generating distributions we care about'' (\cite{Goodfellow2016}, Section~5.5.2]).

\vspace{6pt} 

\authorcontributions{{Conceptualisation}  and methodology, A.N.G. and I.Y.T.; writing---original draft preparation, A.N.G., I.Y.T., and B.G.;~writing---review and editing, all authors; software and validation, I.Y.T., E.M.M., and S.V.S. All authors have read and agreed to the published version of the manuscript.}

\funding{I.Y.T.  was funded by  {UKRI}
(Alan Turing AI Acceleration Fellowship EP/V025295/1). A.N.G., E.M.M., S.V.S., and I.Y.T. were founded by  the {Ministry of Science and Higher Education of the Russian Federation}
(Project No. 075-15-2020-808).}

\conflictsofinterest{The authors declare no conflict of interest. The funders had no role in the design of the study; in the collection, analyses, or interpretation of data; in the writing of the manuscript, or in the decision to publish the results.} 

\abbreviations{Abbreviations}{The following abbreviations are used in this manuscript:\\

\noindent 
\begin{tabular}{ll}
AI & Artificial Intelligence\\
i.i.d. & independent identically distributed\\
ML & Machine Learning\\
PCA & Principal Component Analysis\\
TCA & Transfer Component Analysis\\
DAPCA & Domain Adaptation PCA\\
\end{tabular}}

\appendixtitles{yes} 

\appendix

\section{Elementary Preprocessing of Postclassical Data \label{PrePostclassical}}
 
\subsection{Measure Examples by Examples and Reduce the Number of Attributes to $\dim(DataSet) $ \label{correlation transformation}}

Assume that the number of data points is less than the number of attributes (\ref{DonohoPost}). In this situation, we can decrease the dimension of space by many simple transformations. It is possible to apply PCA and delete all the components with vanishing eigenvalues. This could be a non-optimal approach if originally $d$ is very large. It is also possible to restrict the analysis by the space generated by the data vectors. Let the data sample be a set of $N$ vectors  $\boldsymbol{x}_i$ in ${\mathbb R}^d$. One way to reduce the description is the following {\em correlation transformation} that maps the dataspace into cross-correlation space:
\begin{enumerate}
\item Centralize data (subtract the mean);
\item Delete coordinates with vanishing variance; ({\em Caution}: signals  with small variance may be important, whereas signals with large variance may be irrelevant for the target task! This standard operation can help but can also impair the results.)
\item Standardise data (normalise to unit standard deviations in coordinates), or use another normalisation, if this is more appropriate; ({\em Caution}: transformation to the dimensionless variables is necessary but selection of the scale (standard deviation) affects the relative importance of the signals and can impair the results.)
\item Normalise the data vectors to unit length: $\boldsymbol{x}_i \mapsto \boldsymbol{x}_i/\|\boldsymbol{x}_i\| $ ({\em Caution}: this simple normalisation is  convenient but deletes one attribute, the length. If this attribute is expected to be important than  it could be reasonable to use the mean value of  $\|\boldsymbol{x}_i\|$ that gives normalisation to the unit average length.) 
\item Introduce coordinates in the subspace spanned by the dataset, ${\rm Span}\{\boldsymbol{x}_i\}$ using projections on $\boldsymbol{x}_i$.
\item Each new data point $\boldsymbol{y}$ will be represented by a $N$-dimensional vector of inner products with coordinates $(\boldsymbol{y},\boldsymbol{x}_i)$.
\end{enumerate}

 After this transformation, the data matrix becomes the Gram matrix  $(\boldsymbol{x}_i,\boldsymbol{x}_j)$. For the centralised and normalised data, these inner products can be considered as correlation coefficients.  For such datasets, the number of attributes coincides with the number of data points. The next step may be PCA or another method of dimensionality reduction. The simple and routine formalisation operations can significantly affect the results of data analysis and choosing the right option cannot be done a priori.
 
However, if the dataset is truly multidimensional, then the correlation transformation can return a data matrix with strong diagonal dominance. Centralised random vectors will be almost orthogonal due to the phenomenon of quasi-orthogonality \cite{Kurkova1993, GorbTyuProSof2016}. This effect can make the   application of PCA after the correlation transformation less efficient.
 
There is a different approach to dealing with relatively small samples in multidimensional data spaces. In the Donoho area (see (\ref{DonohoPost}) and Figure~\ref{Fig:Post}a) we can try to produce a probabilistic generative model and then use it for generating additional data. 

The zeroth approximation is the na\"{i}ve Bayes model. This means assuming that the attributes are independent. The probability distribution is the product of distributions of attributes values. In dimension $d$, we need to fit the $ d $ one-dimensional densities, which is much easier than reconstructing the $ d $-dimensional density in the entire data space. The  na\"{i}ve Bayes  approximation can be augmented  by accounting strong pair correlations, etc. The resulting approximation may be represented in the form of a Bayesian network \cite{Cobb2007,Chen2012}. 

There are many methods for generating the probability distribution from data, based on the maximum likelihood estimation married with the network representation of the distribution, like deep latent Gaussian models \cite{Rezende2014}. 

The physical interpretation of the log-likelihood as energy (or free energy) gave rise to many popular heuristic approaches like the Boltzmann machine or restricted Boltzmann machine \cite{Hinton2012} that create approximation of the energy. 

Extensive experience was accumulated in  the use of various generative models of probability distribution. They can be used to leave the Donoho area by augmentation of the dataset with additional samples generated by the model. The statistical status of such augmentation is not always clear because selection of the best model is an intractable problem and we never have enough data and time to solve it. In large dimension,  the models are tested on a standard task: accurate imputations of missing data for the samples never seen before. These tests should check if the majority of correlations captured by the model are significant (and not spurious) and may be used to evaluate the False Discovery Rate (FDR). 

A good heuristic should provide a reasonable balance between the risk of missing significant correlations and the risk of including spurious correlations.
This is a typical multiple testing problem and in the postclassical data world we cannot be always sure that we solved this problem properly. The standard correcting for multiplicity (see, for example,~\cite{Noble2009}) may result in too many false negative errors (missed correlations). However, without such corrections, any findings should be seen as hypothesis generating and not as definitive results \cite{Streiner2011}. This difficulty can be considered as the fundamental incompleteness of the postclassical datasets. 

\subsection{Unsupervised, Supervised, and Semisupervised PCA \label{PCAy}}

PCA remains the standard and very popular tool for dimensionality reduction and unsupervised data preprocessing. It was introduced by K. Pearson in 1900 as a tool for data approximation by straight lines and planes of best fit. Of course, minimisation of the mean square distance from the data point to its projection on a plane (i.e., mean square error of the approximation) is equivalent to maximisation of the variance of projections  (because Pythagorean theorem). This second formulation became the main definition of PCA in   textbooks \cite{Joliffe2011}. The third definition of PCA, which we will use below, is more convenient for developing various generalisations. \cite{GorbanZin2010}. 

 Let a data sample   $\boldsymbol{x}_i \in {\mathbb R}^d$ $(i=1,\ldots, N)$ be given and centralised,  and let $\Pi$ be a projector of $ {\mathbb R}^d$ on a $q$-dimensional plane. The problem is to find the  $q$-dimensional  plane that maximises the scattering of the data projections
\begin{equation}\label{Scat}
\frac{1}{2} \sum_{i,j=1}^n\|\Pi (\boldsymbol{x}_i-\boldsymbol{x}_j)\|^2.
\end{equation}

For projection on a straight line (1D subspace) with the normalised basis vector $\boldsymbol e$ the scattering (\ref{Scat}) is 
\begin{equation}\label{Scat1}
\frac{1}{2} \sum_{i,j=1}^N (\boldsymbol{x}_i-\boldsymbol{x}_j,\boldsymbol e)^2=N\sum_{i=1}^N (\boldsymbol{x}_i,\boldsymbol e)^2 =N(N-1) (\boldsymbol e, Q \boldsymbol e)
\end{equation}
where the coefficients of the quadratic form $ (\boldsymbol e, Q \boldsymbol e)$ are the sample covariance  coefficients $q_{lm}=\frac{1}{N-1}\sum_{i} x_{il}x_{im}$,  and $x_{il}$ $(l=1,\ldots, d)$ are coordinates of the data vector $\boldsymbol{x}_i$. 

If $\{\boldsymbol{e}_1,\ldots , \boldsymbol{e}_q\}$ is an orthonormal basis of the $q$-dimensional plane in  data space, then the maximum scattering of   data projections (\ref{Scat}) is achieved, when $\boldsymbol{e}_1,\ldots , \boldsymbol{e}_q$ are   eigenvectors of  $Q$ that correspond to the $q$ largest eigenvalues  of $Q$ (taking into account  possible multiplicity) $\lambda_1\geq \lambda_2\geq \ldots \geq \lambda_q$. This is the standard PCA exactly. A deep problem with using PCA in data analysis is that the major components are not necessarily the most important or even relevant for the target task. Users rarely need to simply explain a certain fraction of variance. Instead, they  need to solve a classification, prediction, or other meaningful task.
Discarding certain major principal components is a common practice in many applications. First principal components are frequently considered to be associated with technical artifacts in the analysis of omics datasets in bioinformatics, and their removal might improve the downstream analyses \cite{Sompairac2019,Hicks2017}. Even more than 10 first principal components have to be removed sometimes, in order to increase the signal/noise ratio~\cite{Krumm2012}. 

The component ranking can be made more meaningful if we change the form (\ref{Scat}) and include additional information about the target problem in the principal component definition. The form (\ref{Scat}) allows many useful generalisations. Introduce weight $W_{ij}$ for each pair: 
\begin{equation}\label{ScatW}
H=\frac{1}{2} \sum_{i,j=1}^n W_{ij}\|\Pi (\boldsymbol{x}_i-\boldsymbol{x}_j)\|^2.
\end{equation}

The weight   $W_{ij}$ may be positive for some pairs (repulsion) or negative for some other pairs (attraction). The weight matrix is symmetric, $W_{ij}=W_{ji}$. 
Again, the problem of $H$ maximisation leads to a diagonalisation of a symmetric matrix. Consider projection on a 1D subspace with the normalised basis vector $ \boldsymbol{e}$ and define a new quadratic form with coefficients $ q^W_{lm}$:
\begin{equation}\label{Scat1W}
H=\sum_{lm}\left[\sum_i\left(\sum_r W_{ir}\right) x_{il}x_{im} -\sum_{ij}W_{ij}x_{il}x_{jm}\right] e_l e_m = \sum_{lm} q^W_{lm}  e_l e_m.
\end{equation}

Maximum of $H$ (\ref{ScatW}) on $q$-dimensional planes is achieved when this plane is spanned by $q$ eigenvectors of the matrix $Q^W=(q^W_{lm})$ (\ref{Scat1W}) that correspond to $q$ largest eigenvalues   of $Q^W$ (taking into account  possible multiplicity) $\lambda_1\geq \lambda_2\geq \ldots \geq \lambda_q$.

To prove this statement we can mention that the functional $H$ for a $q$-dimensional plane (\ref{ScatW}) is the sum of the functionals (\ref{Scat1W}) calculated for vectors from any orthonormal basis of this plane. Let this basis be $ \{\boldsymbol{e}_1, \ldots , \boldsymbol{e}_q\}$. Decompose each $\boldsymbol{e}_i$ in the orthonormal basis of $Q^W$ eigenvectors  and follow the classical proof for PCA.

There are several methods for the weights assignment:
\begin{itemize}
\item {\em Classical PCA}, $W_{ij}\equiv 1$;
\item {\em Supervised PCA for classification} tasks \cite{Koren2004,MirkesSupPCA2016}. The dataset is split into several classes, $K_v$ $(v=1,2, \ldots ,r)$. Follow the strategy 'attract similar and repulse dissimilar'. If  $\boldsymbol{x}_i$ and $\boldsymbol{x}_j$ belong to the same class, then $W_{ij}=-\alpha<0$ (attraction). If  $\boldsymbol{x}_i$ and $\boldsymbol{x}_j$ belong to different classes, then  $W_{ij}= 1$ (repulsion). This preprocessing can substitute several layers of feature extraction deep learning network \cite{GorbanMirkesTukin2020}. 
\item {\em Supervised PCA for any supervising task}. The dataset for supervising tasks is augmented by labels (the desired outputs). There is proximity (or distance, if possible) between these desired outputs. The weight $W_{ij}$ is defined as a function of this proximity. The closer the desired outputs are, the smaller the weights should be. 
They can change sign (from  classical repulsion, $W_{ij}>0$ to  attraction,  $W_{ij}<0$) or simply change the strength of repulsion.
\item {\em Semi-supervised PCA} was defined for a mixture of labelled and unlabelled data \cite{Song2008}. The data are labelled for classification task. For the labelled data, weights are defined as above for supervised PCA. Inside the set of  unlabelled data the classical PCA repulsion is used.
\end{itemize}

All these  modifications of PCA are formally very close. They are defined by a maximisation of the functional (\ref{ScatW}) for different distributions of weights. This maximisation is transformed into the spectral problem of a symmetric matrix $Q^W$ (see (\ref{Scat1W}) or its simple modification (\ref{transPCA})). The dimensionality reduction is achieved by projection of data onto linear span of $q$ eigenvectors of $Q^W$ that correspond to the largest eigenvalues. 

How many components to retain is a nontrivial question even for the classic PCA~\cite{Cangelosi2007}. The methods based on the evaluation of the fraction of variance unexplained or, what is the same, the relative mean square error of the data approximation by the projection, are popular but we should have in mind that this projection should not only approximate the data but also  be a filter that selects  meaningful features. Therefore, the selection of components to keep depends on the problem we aim to solve and heuristic approaches with several trials of different numbers of components may be more useful than an unambiguous formal criterion. Special attention is needed to the cases when some eigenvalues of $Q^W$ become negative. Let $\lambda_1 \geq \lambda_2 \geq \ldots \geq \lambda_r >0$ but for other eigenvalues $0 \geq \lambda_{r+1}\geq \ldots$. In this case, a further increase in the dimension of the approximating plane above $r$ does not lead to an increase in $H$ but definitely increases the quality of data approximation. The standard practice is not to use eigenvectors that correspond to non-positive eigenvalues~\cite{GorbanMirkesTukin2020}.
 
\subsection{DAPCA---Domain Adaptation PCA \label{DAPCA}}

The classical hypothesis of machine learning is existence of the probability distribution and the same (even unknown) distribution for the training and test sets. 
The problem of domain adaptation arises when the training set differs from the data that the system should work with under operational conditions. Such situations are typical. The problem is that the new data have no known labels. We have to utilise a known labelled training set (from the ``source domain') and a new unlabelled training set (from the 'target domain'). The idea is to modify the data and to make the non-labelled data   as close to the labelled one as possible. This transformation should erase the difference between the data distributions in two sets and, at the same time, do not destroy the possibility to solve  effectively the machine learning problem for the labelled set. 

The key question in domain learning is definition of the objective functional:  how to measure the difference in distributions between the source domain sample and the target domain sample. The clue to  the answer gives the idea \cite{Ben-David2010}: 
\begin{itemize}
\item Select a family of classifiers in data space;
\item Choose the best classifier from this family for separation the source domain samples from the target domain samples;
\item The error of this classifier is an objective function for maximisation (large classification error means that the samples are indistinguishable by the selected family of classifiers).
\end{itemize}

Ideally, there are two systems: a classifier that distinguishes the feature vector as either a source or target and a feature generator that learns a combination of tasks: to mimic the discriminator and to ensure the successful learning in the source domain. There are many attempts to implement this idea \cite{Sun2015,Saito2018}. In particular, an effective neural network realisation trains a deep neural network system to accurately classify source samples but decreases the ability of the associated  classifier that uses the same feature set to detect whether each example belongs to the source or target domains \cite{Ganin2016}. The scattering objective function (\ref{ScatW}) can combine these two targets for learning of feature generation: success in the learning in the source domain and indistinguishability of the source and target datasets.

{\em Transfer Component Analysis (TCA)} was proposed to specify attraction between the clouds of projections of labelled and unlabelled data \cite{Matasci2011}. The distance between the source and target samples was defined as the distance between the projections of their mean points.  Attraction between the mean points of the labelled and unlabelled data was postulated. Let $\boldsymbol{\mu}_{\rm L}$ and $\boldsymbol{\mu}_{\rm U}$ be these mean points. Their attraction means that a new term should be added to $Q^W$ (\ref{Scat1W}):
\begin{equation}\label{transPCA}
q^W_{lm}=\sum_i\left(\sum_r W_{ir}\right) x_{il}x_{im} -\sum_{ij}W_{ij}x_{il}x_{jm} -\beta ({\mu}_{{\rm L} l}-{\mu}_{{\rm U} l})({\mu}_{{\rm L}m}-{\mu}_{{\rm U} m}),
\end{equation}
 where weights $W_{ir}$ are assigned by  the same rules as in semisupervised PCA, and  $\beta>0$ is the attraction coefficient between the mean points of the labelled and unlabelled data samples. 

{\em Domain Adaptation PCA (DAPCA)} also takes advantage of this idea of task mix within a weighted PCA framework (\ref{ScatW}). The classifier used is the classical kNN ($k$ nearest neighbours). Let the source dataset (input vectors) be $\mathbf{X}$, the target dataset be  $\mathbf{Y}$, $\mathbf{X}$ is split into different classes: $\mathbf{X}=K_1\cup \ldots \cup K_r$. Enumerate points in $\mathbf{Y} \cup \mathbf{X}$ The weights are:
\begin{itemize}
\item If  $\boldsymbol{x}_i, \boldsymbol{x}_j \in K_v$ then $W_{ij}=-\alpha<0$ (the source samples from one class, attraction);
 \item If  $\boldsymbol{x}_i\in K_u$ $\boldsymbol{x}_j \in K_v$ ($u\neq v$) then $W_{ij}=1$ (the source samples from different classes, repulsion);
\item  $\boldsymbol{x}_i, \boldsymbol{x}_j \in \mathbf{Y}$ then $W_{ij}=\beta>0$ (the target samples, repulsion);
\item For each target sample  $\boldsymbol{x}_i \in \mathbf{Y}$ find $k$ closest source samples in $\mathbf{X}$. Denote this set $E_i$. For each  $\boldsymbol{x}_j \in  E_i$, $W_{ij}=-\gamma<0$ (the weight for connections of a target sample and the $k$ closest source samples, attraction).
\end{itemize}

The weights in this method depend  on three non-negative numbers, $\alpha$, $\beta$, and $\gamma$ and on the number of nearest neighbours, $k$.  Of course, the values of the constants can vary for different samples and classes, if there is sufficient reason for such a generalisation.	

kNN classification can be affected by irrelevant features that create difference between the source and target domains and should be erased in the feature selection procedure. This difficulty can be resolved by the {\em iterative DAPCA}. Use the basic algorithm as the first iteration. It gives the $q$-dimensional plane of major components (the eigenvectors $Q^W$)  with the orthogonal projector in it $\Pi_1$. Find for each target sample $k$ nearest neighbours from the source samples in the projection on this plane (use for definition of $k$ nearest neighbours  the seminorm $\|\Pi_1(\boldsymbol{x})-\Pi_1(\boldsymbol{y}\|$). Assign new $W_{ij}$ using these nearest neighbours. Find new projector $\Pi_2$ and new nearest neighbours. Iterate. The iterations converge in a finite number of steps, because the functional $H$ (\ref{Scat1W}) increases at each step (as in the $k$-means and similar splitting algorithms). Even if the convergence (in high dimensions) is too long, then the early stop can produce a useful feature set. The  iterative DAPCA helps also to resolve the classical distance concentration difficulty: in essentially  large dimensional distributions the kNN search may be affected by the distance concentration phenomena: most of the distances are close to the median value \cite{Pestov2013}. Even use of fractional norms or quasinorms do not save the situation \cite{Mirkes2020}, but dimensionality reduction with deleting the irrelevant features may help.

If the target domain is empty then TPA, DAPSA, and iterative DAPCA  degenerate to the semi-supervised PCA in the source domain. If there is no source domain then they turn into classical PCA in the target domain. 

The described procedures of supervised PCA, semi-supervised PCA, TCA, DAPCA, or iterative DAPCA prepare a relevant feature space. The distribution of data in this space is expectedly far from a regular unimodal distribution. It is assumed that in this space the samples will form dense clumps with a lower data density between them.

\section{ 'Almost Always' in Infinite-Dimensional Spaces \label{almost}} 

As it was mentioned in Section~\ref{Sec:Compact}, in the infinite-dimensional limit many statements about high or low probabilities transform into  0-1 laws: something happens almost always or almost newer. Such limits for concentrations on spheres and their equators  were discussed by L\'{e}vy \cite{Levy1951} as an important part of the measure concentration effects. In physics, this limit corresponds to the so-called thermodynamic limit of statistical mechanics~\cite{Khinchin1949,Thompson2015}.  The original Kolmogorov 0-1 law states, roughly speaking, that an event that depends on an infinite collection of
independent random variables  but is independent of any finite subset of these variables, has probability zero or one (for precise formulation we refer to the monograph \cite{Kolmogorov2018}).  The infinite-dimensional 0-1 asymptotic might bring more light and be more transparent than the probabilistic formulas. 

This may be surprising, but the problem is what 'almost always' means.
Formally, various definitions of genericity are constructed as
follows. All systems (or cases, or situations, and so on) under
consideration are somehow parameterised---by sets of vectors,
functions, matrices, etc. Thus, the 'space of systems' $Q$ can be
described. Then the {\it  'meagre (or thin) sets'} are introduced into $Q$,
i.e., the sets, which we shall later neglect. The union of a finite
or countable number of meager  sets, as well as the intersection of
any number of them should be meager set again, while the whole $Q$ is
not thin. There are two traditional ways to determine thinness.

\begin{enumerate}
\item The sets of {\it measure zero} are negligible.  
\item The sets of {\em Baire first category} are negligible.
 \end{enumerate}

The first definition requires existence of a special measure such that all relevant distributions are expected to be absolute continuous with respect to it. In Theorem~\ref{Th:prototype}, for example,  we assumed that the probability distribution (yet unknown) has density and is absolutely continuous with respect to Lebesgue measure. Moreover, we used a version of the 'Smeared (or Smoothed) Absolute Continuity' (SmAC) condition (\ref{bounded}) \cite{GorbanGrechukTykin2018,Gorbetal2018}, which means that the  sets of relatively small volume cannot have high probability, whereas absolute continuity means that sets of zero volume have probability zero. Unfortunately, in the infinite-dimensional  spaces we usually do not have such a sensible measure. It is very easy to understand if we look on the volumes of balls in Hilbert space with orthonormal basis $\{\boldsymbol{e}_i\}$. If the measure of a ball is function of its radius and the measure of a ball of radius $R$ is finite, then the balls of radius   $R/4$ have zero measure (because infinitely many such balls with the centres at points $R\boldsymbol{e}_i/2$  can be packed in the ball of  radius   $R/4$), and, therefore, the ball of radius $R$ has zero measure because it can be covered by a countable set of balls of radius $R/4$. Hence, all balls have either zero or infinite measure.

The second definition is widely accepted when we deal with the functional parameters.
The construction begins with nowhere dense sets. The set $Y$ is
nowhere dense in $Q$, if in any non-empty open set $V\subset Q$ (for
example, in a ball) there exists a non-empty open subset $W\subset V$
(for example, a ball), which does not intersect with $Y$: $W\cap Y
=\emptyset$. Roughly speaking, Y is 'full of holes'---in any
neighbourhood of any point of the set $Y$ there is an open hole.
Countable union of nowhere dense sets is called the set of first
category. The second usual way is to define thin sets as the {\it
sets of first category}. A   residual set  (a 'thick' set)  is
the complement of a set of the first category. If a set is not  meagre it is said to be of the second category. The Baire classification is nontrivial in the so-called  Baire spaces, where every intersection of a countable collection of open dense sets is also dense. Complete metric spaces and, in particular, Banach spaces are Baire spaces. Therefore, for Banach spaces of functions, the common definition of negligible set is 'set of first Baire category'.  Such famous results as transversality theorem in differential topology \cite{Golub1974} or Pugh closing lemma \cite{Pugh1967} and Kupka-Smale theorem  \cite{Palis1982} in differential dynamics.

Despite these great  successes, it is also widely recognised that the Bair category approach to generic properties requires at least great care. Here are some examples of correct but useless statements about
'generic' properties of function: almost every continuous
function is not differentiable; almost every $C^1$-function is not
convex. Their meaning for applications is most probably this: the
genericity used above for continuous functions or for
$C^1$-function is irrelevant to the subject.

Contradictions between the measure-based  and  category-based  definitions of negligible sets are well known even in dimension one: even the real line  $R$ can be divided into two sets, one of
which has zero measure, the other is of first category \cite{Oxtoby2013}. Genericity in the sense of measure and genericity in the sense of category differ significantly in the applications where both concepts can be used. 

The conflict between the two main views on genericity and negligibility  stimulated efforts to invent new and stronger approaches. The formal requirements to new definitions are: 
\begin{itemize}
\item A union of countable family of thin sets should be thin. 
\item Any subset of a thin set should be thin. 
\item The whole space is not thin. 
\end{itemize}

Of course, if we take care not to throw the baby out with the bath water then in $\mathbb{R}^n$, where both classical definition are applicable, we expect that thin sets should be of first category and have zero measure. It was not clear a priori whether such a theory is possible with proof nontrivial and important generic properties. It turned out that it is possible. To substantiate the effectiveness of evolutionary optimisation, a theory of completely negligible sets in Banach spaces was developed.   \cite{Gorban1984,Gorban2007}.

Let $Q$ be a real Banach space. Consider compact subsets in $Q$ parameterised by points of a compact space $K$. It can be presented as a Banach space $C(K,Q)$ of
continuous maps $K \rightarrow Q$ in the maximum norm. 
 
\begin{Definition}\label{CompletelyThin}A set $Y \subset Q$ is completely thin, if for any
compact space $K$ the set of continuous maps $\Psi: K \rightarrow Q$ with
non-empty intersection $ \Psi(K) \cap Y \neq \emptyset$ is 
set of first Bair category.
\end{Definition}

The union of a finite or countable number of completely thin sets is completely thin. Any subset of a completely thin point is completely thin, while the whole $Q$ is not. 
A set $Y$ in the Banach space $Q$ is completely thin,   if for any
compact set $K$ in $Q$ and arbitrary positive $\varepsilon>0$ there
exists a vector $q \in Q$, such that $\|q\|<\varepsilon$ and $K+q$
does not intersect $Y$: $(K+q) \cap Y =\emptyset$.
All compact sets in infinite-dimensional Banach spaces and closed linear subspaces with
infinite codimension  are completely thin. 

Only empty set is completely thin in a finite-dimensional space  $\mathbb{R}^n$. 

Examples below demonstrate that almost all continuous functions have very natural properties: the set of zeros is nowhere dense, and the (global) maximiser is unique. Below the wording 'almost always' means: the set of exclusions is
completely thin. 

\begin{Proposition}[\cite{Gorban1984,Gorban2007}]\label{2.2}Let $X$ have no isolated
points. Then
\begin{itemize}
\item Almost always a function $f \in C(X)$ has nowhere dense
set of zeros $\{x \in X \, | \, f(x)=0 \}$ (the set of exclusions is
completely thin in $C(X)$).  
\item Almost always a   function $f \in C(X)$ has only one point of global maximum.
\end{itemize}
 \end{Proposition}
 
 The following proposition is a tool for proof that some  typical properties of functions hold almost always for all functions from a generic compact set.
 
 \begin{Proposition}[\cite{Gorban1984,Gorban2007}]\label{2.1}If a set $Y$ in the Banach space $Q$ is completely thin, then for any compact metric space $K$ the set of continuous maps $\Psi: K
\rightarrow Q$ with non-empty intersection $ \Psi(K) \cap Y
\neq \emptyset$ is completely thin in the Banach space $C(K,Q)$. \ \
 \end{Proposition}
 
\begin{Proposition}[\cite{Gorban1984,Gorban2007}] \label{2.3}Let $X$ have no isolated points. Then
for any compact space $K$ and almost every continuous map $\Psi: K
\rightarrow C(X)$ all functions $f \in \Psi(K)$ have nowhere
dense sets of zeros (the set of exclusions is completely thin in
$C(K,C(X))$).  
\end{Proposition}

In other words, in almost every compact family of continuous functions all the functions have nowhere dense sets of zeros.
 
Qualitatively, the concept of a completely thin set was introduced as a tool for identifying typical properties of infinite-dimensional objects, the violation of which is unlikely (`improbable')  in any reasonable sense. 

\section{Flowchart of Multi-Corrector Operation \label{Sec:Flowcharts}}

\begin{figure}[H]
 \includegraphics[width=0.82\textwidth]{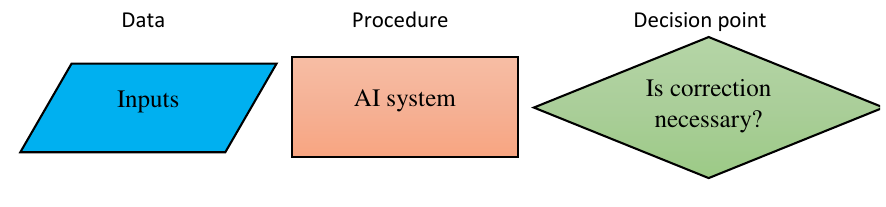}
\caption{  {\em Notations used in the flowcharts.} 
 All flowcharts 
 use a unified set of blocks: blocks in the form of parallelograms display data, rectangular blocks display procedures, and blocks in the form of rhombuses display the branching points of processes (algorithms) or decision points. The arrows reflect the transfer of data and control. \label{Fig:A1}} 
\end{figure}

In   Section~\ref{Multicluster}, we introduced multi-corrector of AI systems. The basic scheme of this device is presented in  Figure~\ref{ComplexCorrector}. It includes several elementary correctors (see Figure~\ref{SingleCorrector}) and a dispatcher. A cluster of errors is owned by each elementary corrector. An elementary corrector evaluates the risk of errors from its own cluster for an arbitrary operation situation and takes the decision to correct or not to correct the legacy AI decision for this situation. For any situation, the dispatcher selects the most appropriate elementary corrector to make a decision about correction. To find a suitable corrector, it uses a cluster error model. When new errors are found, the cluster model changes. More detailed presentation of multi-corrector operation is given by the following flowcharts. The notations are described in Figure~\ref{Fig:A1}.

\begin{figure}[H]
 \includegraphics[width=0.88\textwidth]{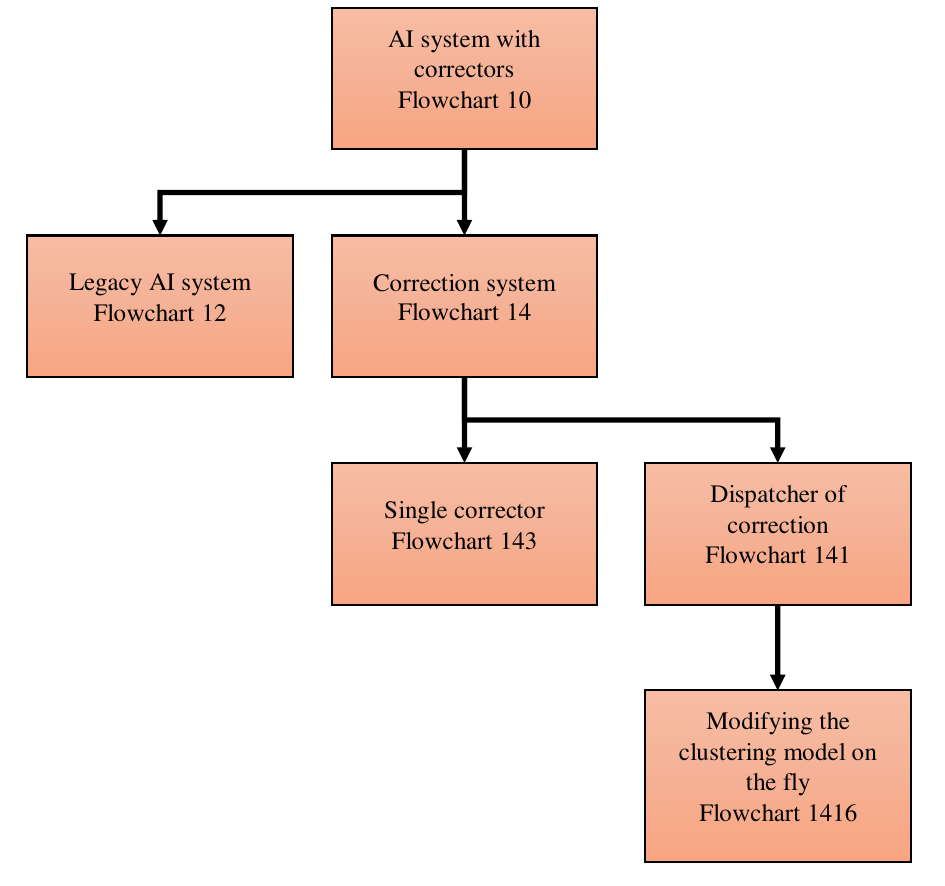}
\caption{ {\em The tree of flowcharts}: 10---Operation of the modified AI system (Figure~\ref{Fig:A3}); 12---Operation of the legacy AI system  (Figure~\ref{Fig:A4}),  14--- Operation of the correction system  (Figure~\ref{Fig:A5}); 143---Single corrector operation  (Figure~\ref{Fig:A6});  141---The work of the dispatcher  (Figure~\ref{Fig:A7}); 1415--- Online modification of the cluster model  (Figure~\ref{Fig:A8}).  \label{Fig:tree}} 
\end{figure}

Flowcharts and blocks  are numbered. The  flowchart number is mentioned at the top of the drawings.  If a block is present in different flowcharts, then it carries the number assigned to it in the top-level flowchart.  The relations between different flowcharts are presented in Figure~\ref{Fig:tree}.

  \begin{figure}[H]
 \includegraphics[width=0.54\textwidth]{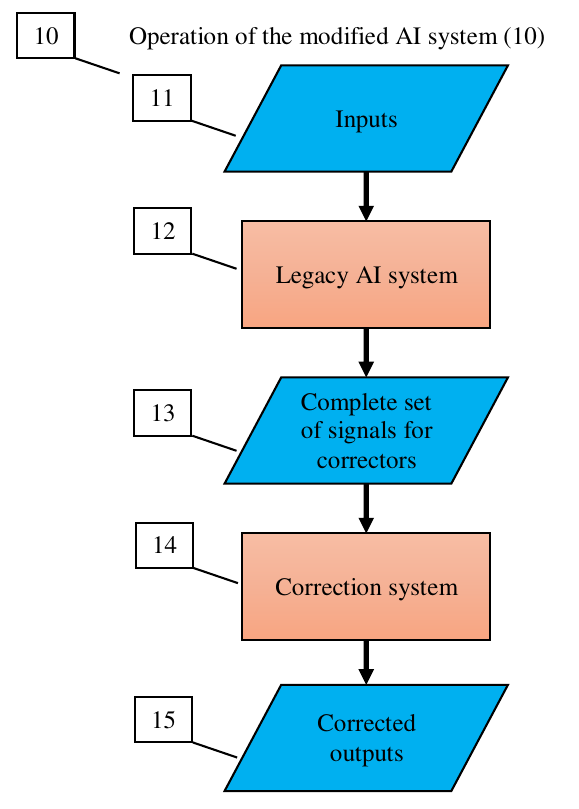}
\caption{ {\em Operation of the modified AI system (10).}
 Input signals (11) are fed to the input of the AI system (12), which at the output gives out the complete vector of the signal (13) that can be used for correction.
The complete signal vector (13) is fed to the input of the correction system (14). The correction system (14) calculates the correction of the output signals (15). \label{Fig:A3}} 
\end{figure}

  \begin{figure}[H]
  \includegraphics[width=0.9\textwidth]{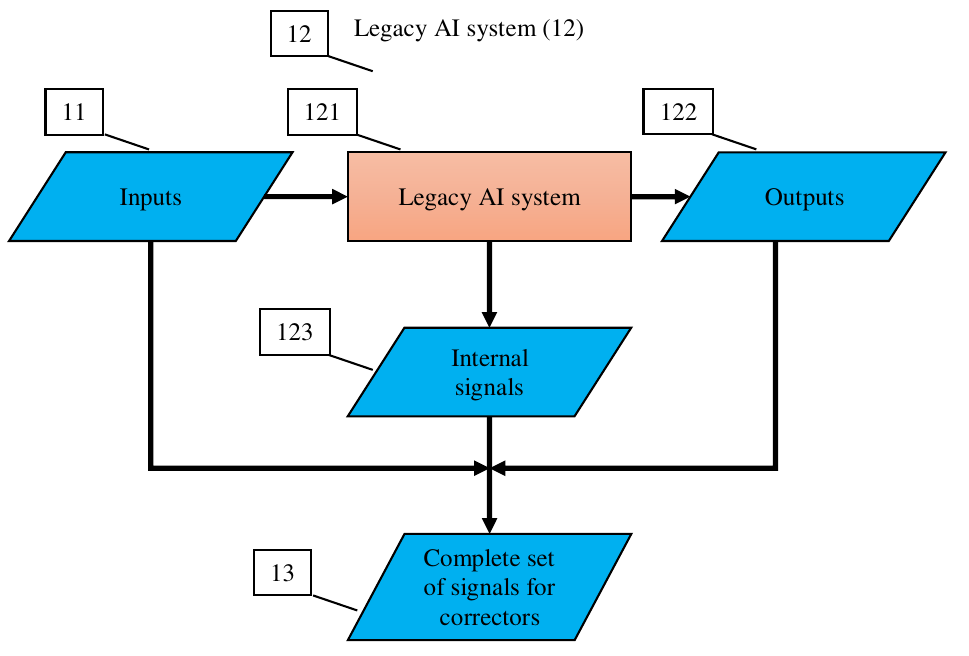}
\caption{ {\em Operation of the legacy AI system (12).}
Input signals (11) are fed to the input of the AI system. The AI system generates vectors of internal signals (123) and output signals (122). Input signals (11), internal signals (123), and output signals (122) form the complete signal vector (13). \label{Fig:A4}} 
\end{figure}

  \begin{figure}[H]
 \includegraphics[width= 0.5\textwidth]{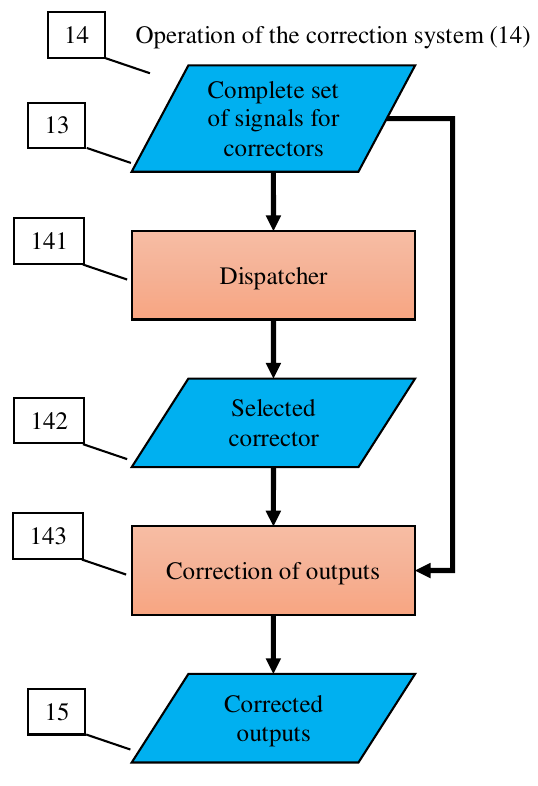}
\caption{ {\em  Operation of the correction system (14).}
The complete vector of signals (13) is fed to the dispatcher input (141). The dispatcher (141) selects from the correctors the one that most closely matches the situation (142). The selected corrector (142) and the complete signal vector (13) are used to correct the signals (13). The computed corrected outputs (15) are returned. \label{Fig:A5}} 
\end{figure}

 \begin{figure}[H]
\includegraphics[width=0.85\textwidth]{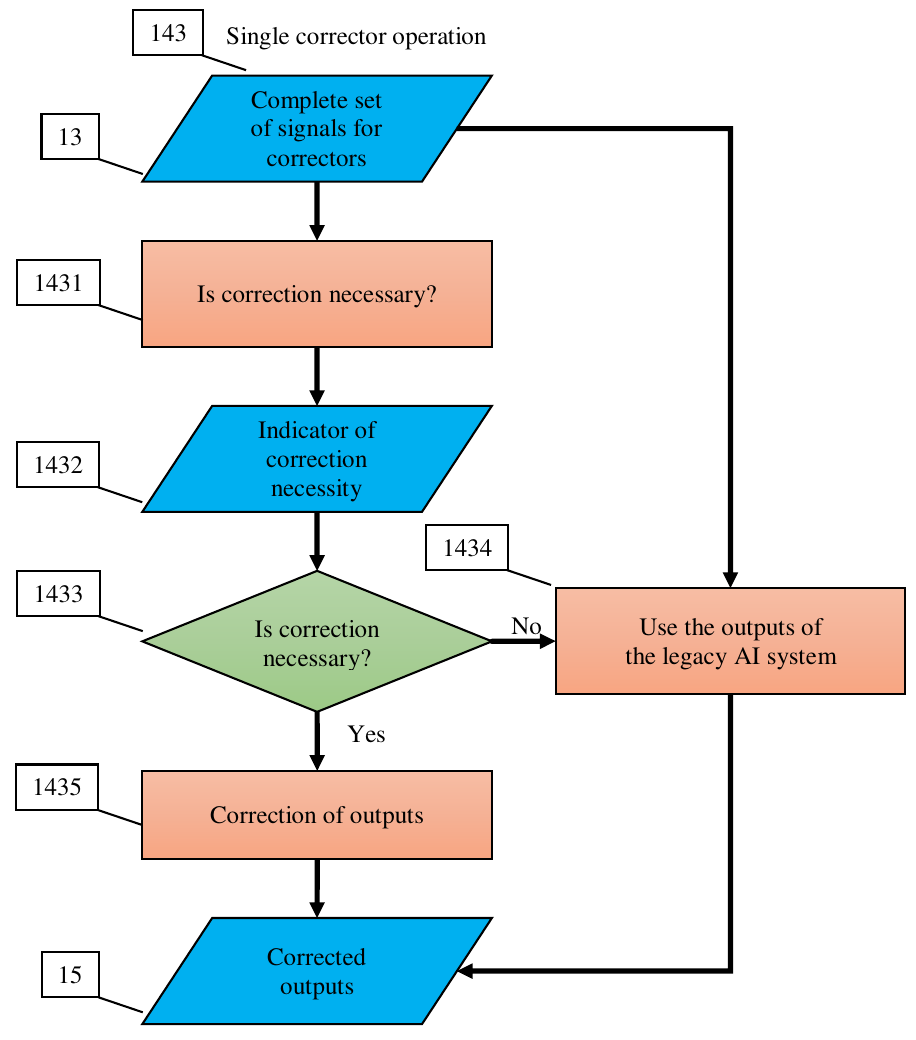}
\caption{ {\em Single corrector operation (143).}
The complete  vector of signals (13) is used to decide whether a correction is needed (1431). If it is necessary, then correction (1435) is performed, and the resulting vector of output signals (15) is sent to the output. If there is no need for correction, then the vector of output signals is extracted (1434) from the complete vector of signals (13), and the resulting vector of output signals (15) is transmitted to the output. \label{Fig:A6}} 
\end{figure} 

 \begin{figure}[H]
\includegraphics[width= 0.86\textwidth]{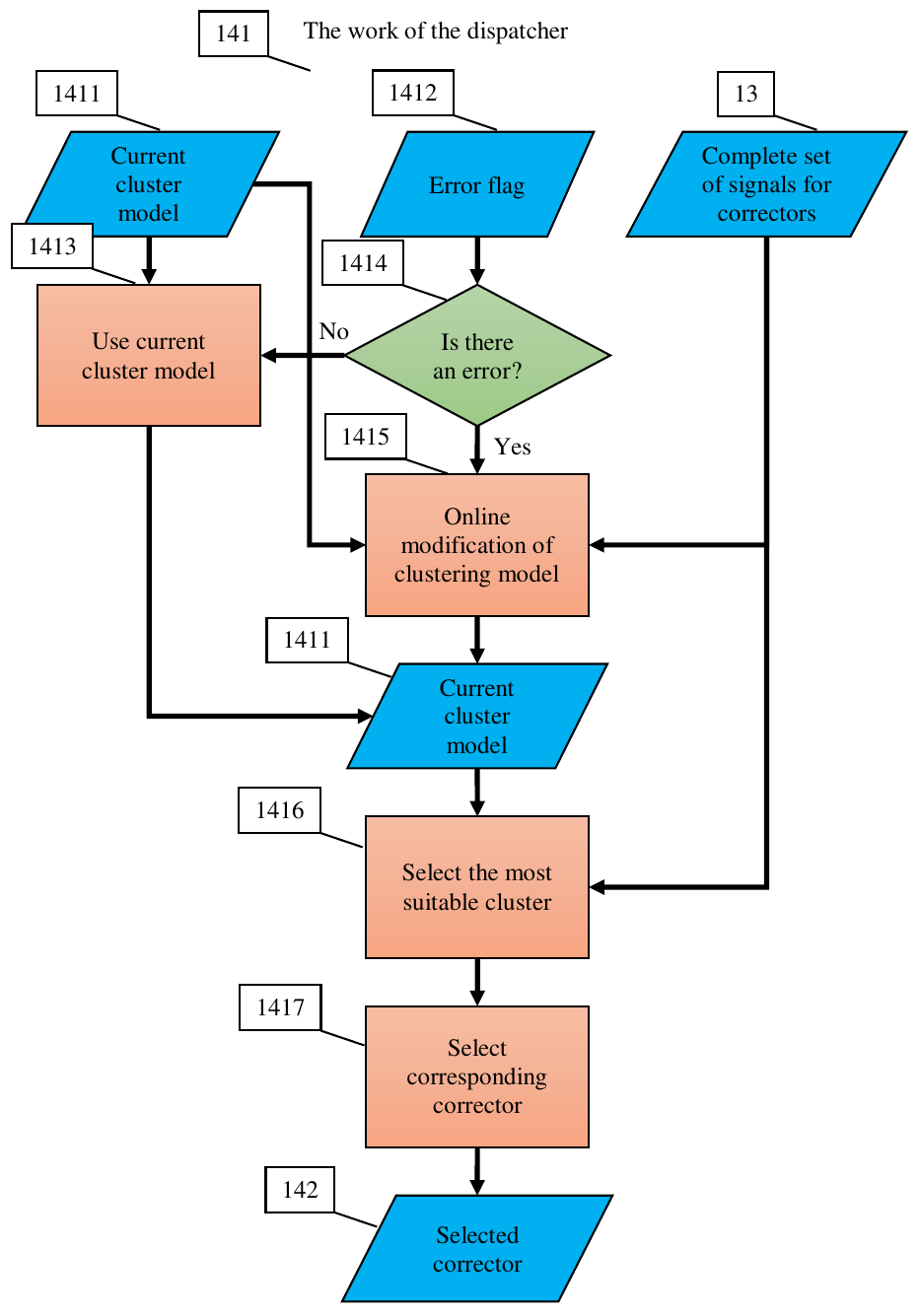}
\caption{{\em The work of the dispatcher (141).}
If the error flag (1412) is detected (1414), then the current cluster model (1411) and the complete signal vector (13) are used to modify the cluster model (1415) online. The modified cluster model becomes the current one (1411). If the error flag (1412) is not detected (1414), then the current cluster model (1411) is selected (1413) for use (1411). Based on the cluster model (1411) and the complete  signal vector (13), the most suitable cluster (1416) is selected. Then, the corrector (142) corresponding to this cluster is selected (1417). \label{Fig:A7}}
\end{figure} 

 \begin{figure}[H]
\includegraphics[width= 0.85\textwidth]{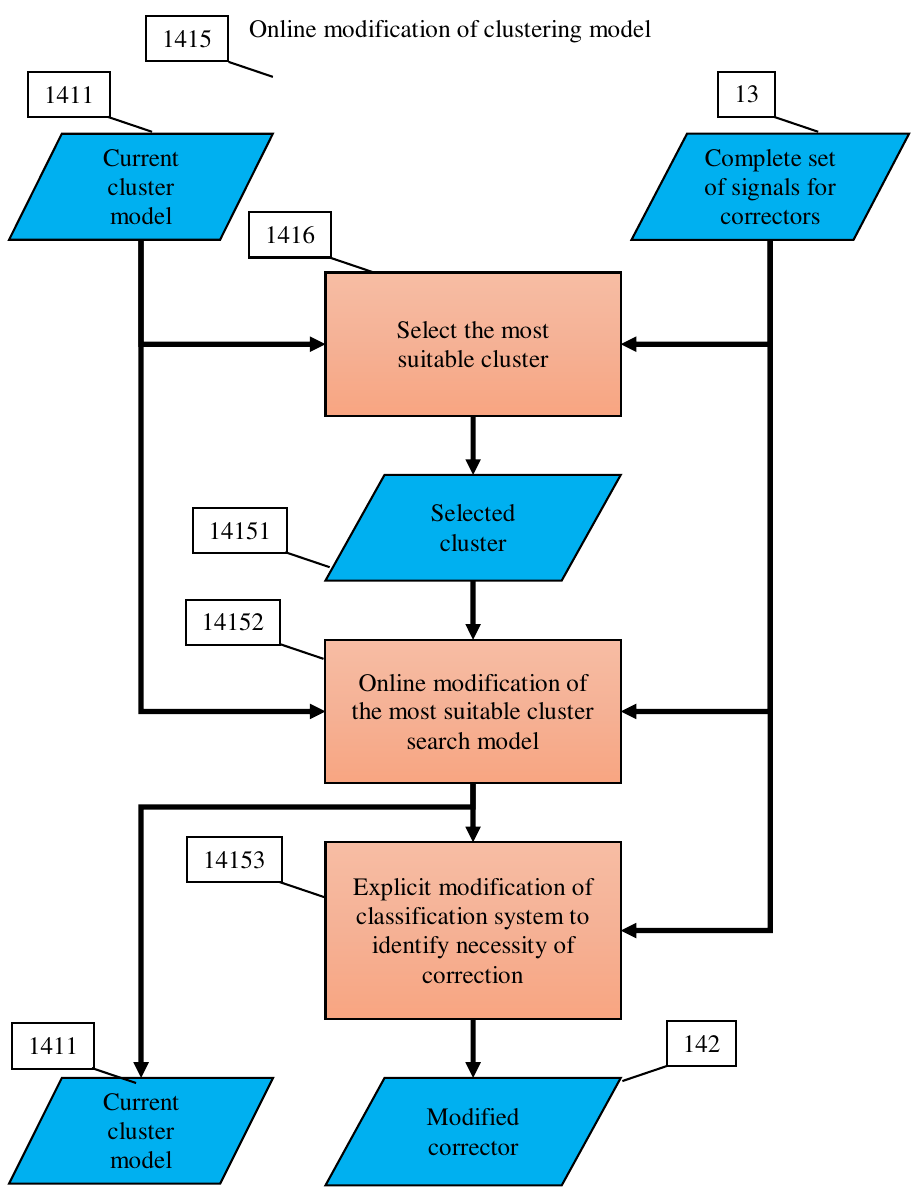}z
\caption{ {\em Online modification of the cluster model (1415).} 
Based on the current cluster model (1411) and the complete  signal vector (13), the most suitable cluster (14151) is selected (1416). Online modification of the rule for determining the most suitable cluster (14152) is performed. After setting up the new cluster model (1411),  the classifier for this corrector to make a decision about the need for correction is explicitly modified (14153). The modified corrector (142) together with the new cluster model (1411) forms an updated version of the correction system (14). \label{Fig:A8}} 
\end{figure}


\reftitle{References}




 \end{document}